\documentclass[twocolumn]{autart}
\usepackage{xcolor}
\usepackage{tikz,lipsum,lmodern}
\usepackage[most]{tcolorbox}
\usepackage{algorithm}
\usepackage{epstopdf}
\usepackage{algorithmic}
\usepackage{caption}
\usepackage{subcaption}

\makeatletter
\let\classAND\AND
\let\AND\relax
\usepackage{algorithmic}

\let\AND\classAND
\AtBeginEnvironment{algorithmic}{\let\AND\algoAND}

\newenvironment{proof}{\textit{Proof:}}{\hfill$\square$}
\usepackage{lineno}
\usepackage{float}
\usepackage{lipsum}
\usepackage{fancyhdr}
\usepackage{url}
\usepackage{bbold}
\usepackage{amsfonts}
\usepackage{amsmath,amssymb}
\usepackage{blindtext}
\usepackage{diagbox}
\usepackage{booktabs}
\usepackage{graphicx}
\usepackage{listings}
\usepackage{wrapfig}
\usepackage{mathrsfs}
\usepackage{array}
\usepackage{times}
\usepackage{url}
\usepackage{upgreek}
\usepackage{makecell}
\usepackage{bbm}
\usepackage{epstopdf}
\usepackage{color}
\usepackage{flushend}
\usepackage{multirow}
\usepackage{rotating}
\usepackage{standalone}
\usepackage{enumitem}
\usepackage{flushend}
\usepackage{multirow}
\usepackage{rotating}
\usepackage{standalone}
\usepackage{enumitem}
\usepackage{stackengine}
\usepackage{algorithmic}
\newcommand{\mc}{\mathcal}

\newcommand{\mr}{\mathrm}
\newcommand{\mb}{\mathbb}
\newcommand{\m}[1]{{\bf{#1}}}

\renewcommand{\mc}[1]{\ensuremath{\mathcal{#1}}}

\newcommand\norm[1]{\lVert#1\rVert}

\newcommand{\x}{\textbf{x}}
\newcommand{\y}{\textbf{y}}
\newcommand{\xx}{\mathrm{x}}
\newcommand{\yy}{\mathrm{y}}
\newcommand{\zz}{\mathrm{z}}

\newcommand{\W}{\textbf{W}}
\newcommand{\p}{\textbf{p}}
\newcommand{\g}{\mathbf{g}}
\newcommand{\M}{\textbf{M}}

\newcommand{\mB}{\mathbf{B}}
\newcommand{\F}{\mathbf{F}}
\newcommand{\uu}{\textbf{u}}
\newcommand{\bbD}{\mathbf{D}}
\newcommand{\bbI}{\mathbf{I}}

\newcommand{\bbH}{\mathbf{H}}

\newcommand{\bbE}{\mathbf E}
\newcommand{\hbH}{\mathbf H}

\newcommand{\degree}{{\mathrm{deg}}}
\usepackage{lineno}

\DeclareMathOperator*{\argmin}{arg\,min}

\begin{document}
\begin{frontmatter}
\title{A Penalty-Based Method for Communication-Efficient Decentralized Bilevel Programming}

\thanks[footnoteinfo]{
Corresponding author}      

\author[Iran]{Parvin Nazari}\ead{ p$\_$nazari@aut.ac.ir},    
\author[USA1]{Ahmad Mousavi}\ead{mousavi@american.edu},               
\author[USA2]{Davoud Ataee Tarzanagh\thanksref{footnoteinfo}}\ead{tarzanaq@upenn.edu},  
\author[USA3]{George Michailidis}\ead{gmichail@ucla.edu}  

\address[Iran]{Amirkabir University of Technology, Iran}
\address[USA1]{American University, USA}  
\address[USA2]{University of Pennsylvania, USA}
\address[USA3]{University of California, Los Angeles, USA}


\begin{abstract}
Bilevel programming has recently received attention in the literature due to its wide range of applications, including reinforcement learning and hyper-parameter optimization. However, it is widely assumed that the underlying bilevel optimization problem is solved either by a \textit{single} machine or, in the case of multiple machines connected in a star-shaped network, i.e., in a federated learning setting. The latter approach suffers from a high communication cost on the central node (e.g., parameter server). Hence, there is an interest in developing methods that solve bilevel optimization problems in a communication-efficient, \textit{decentralized} manner. To that end, this paper introduces a \textit{penalty function}-based decentralized algorithm with theoretical guarantees for this class of optimization problems. Specifically, a distributed alternating gradient-type algorithm for solving consensus bilevel programming over a decentralized network is developed. A key feature of the proposed algorithm is the estimation of the hyper-gradient of the penalty function through decentralized computation of \textit{matrix-vector} products and a few \textit{vector communications}.
 The estimation is integrated into an alternating algorithm for solving the penalized reformulation of the bilevel optimization problem. Under appropriate step sizes and penalty parameters, our theoretical framework ensures non-asymptotic convergence to the optimal solution of the original problem under various convexity conditions. Our theoretical result highlights improvements in the iteration complexity of decentralized bilevel optimization, all while making efficient use of vector communication. Empirical results demonstrate that the proposed method performs well in real-world settings.
\end{abstract}
\end{frontmatter}
\section{Introduction} \label{sec:introduction}
\begin{figure*}[t]
\centering
\vspace{-.2cm}
\begin{minipage}{0.5\textwidth}
 \centering
\includegraphics[width=0.6\textwidth]{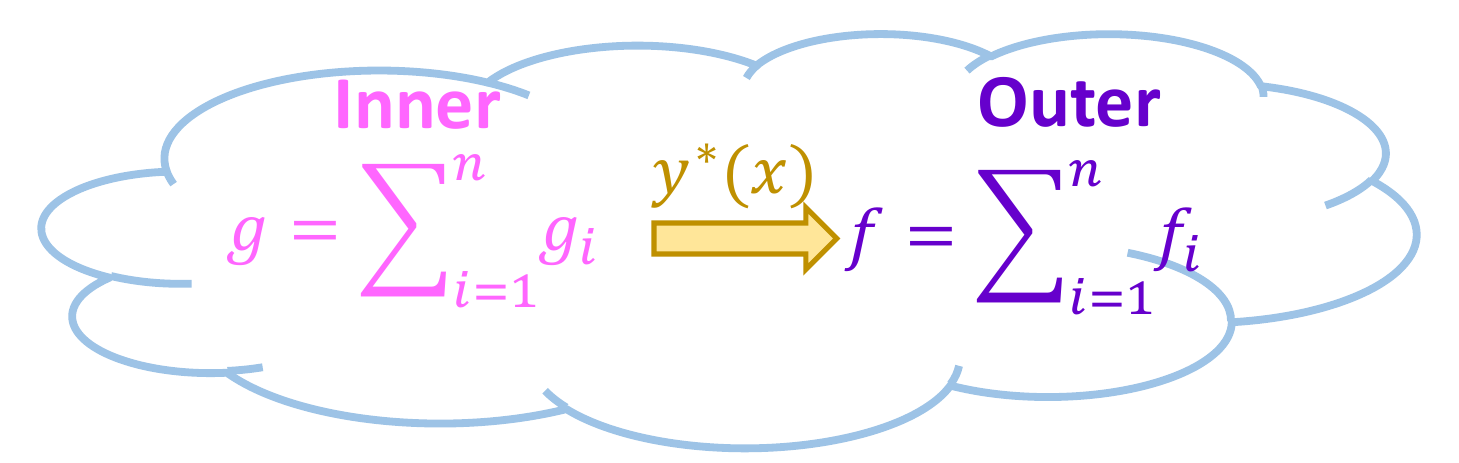}
\end{minipage}%
\hfill
\begin{minipage}{0.5\textwidth}
 \centering
\includegraphics[width=0.7\textwidth]{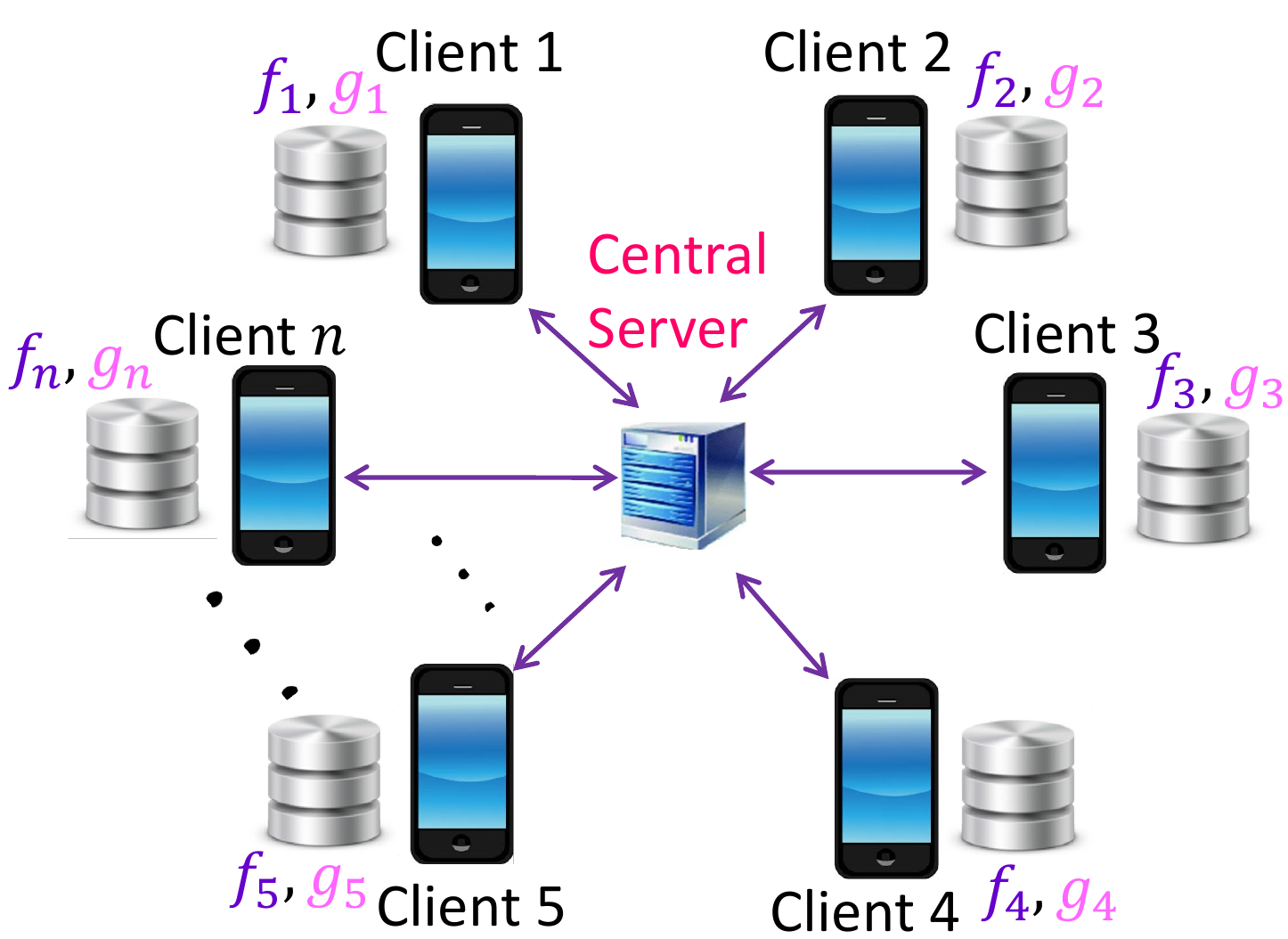}
\subcaption{Star-Shaped Network }\label{fig:1a}
\end{minipage}%
\begin{minipage}{0.5\textwidth}
 \centering
\includegraphics[width=0.7\textwidth]{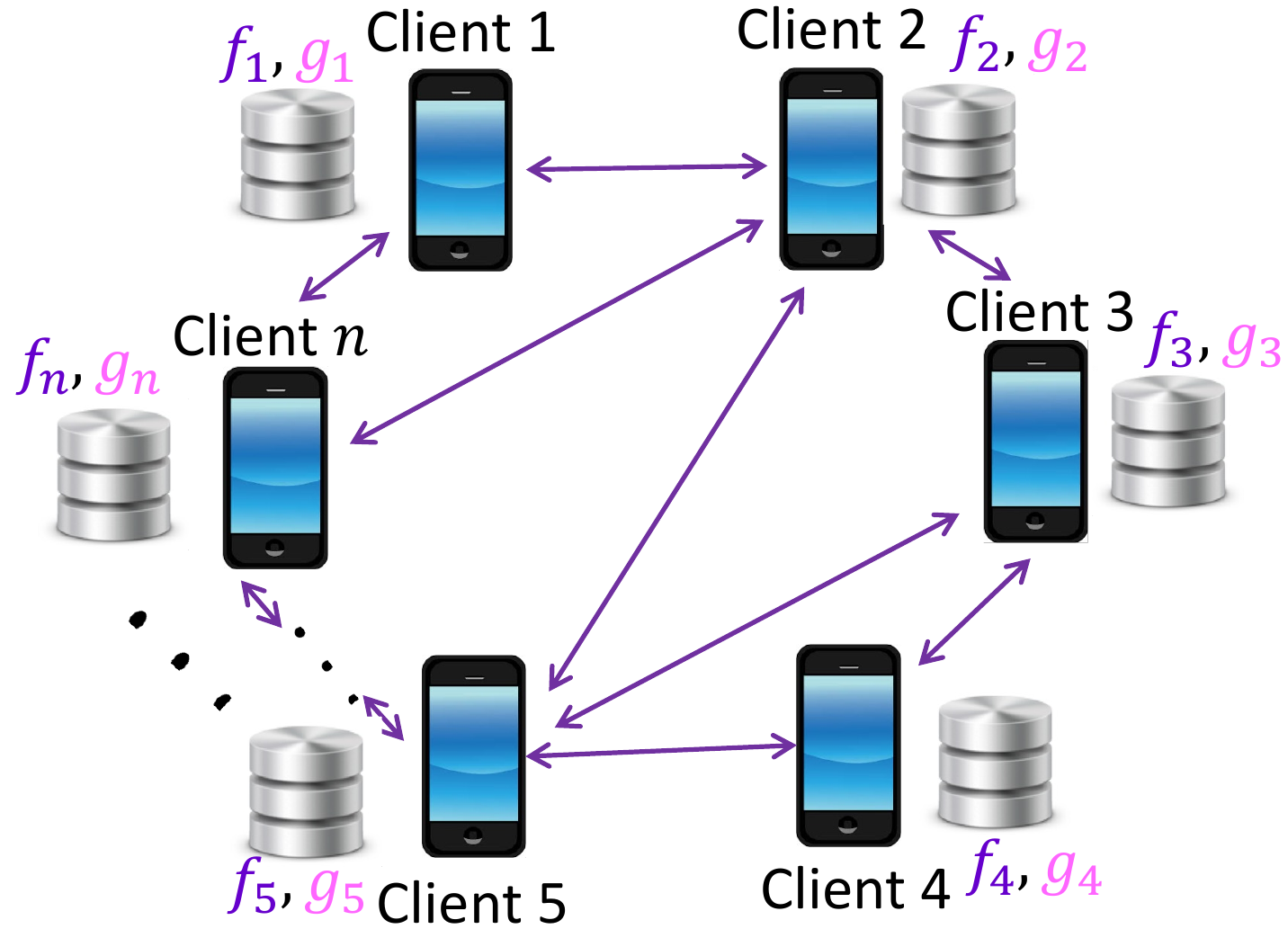}
\subcaption{Decentralized Network}\label{fig:1b}
\end{minipage}%
\caption{Distributed Bilevel Optimization.} \label{fig:1}
\end{figure*}

Bilevel programming has found applications in various fields, including economics \cite{dempe1998implicit,stackelberg1934marktform,von1952theory}, transportation \cite{migdalas1995bilevel}, management \cite{bostian2015incorporating,whittaker2017spatial}, meta-learning~\cite{bertinetto2018meta}, hyperparameter optimization~\cite{feurer2019hyperparameter}, neural network architecture search~\cite{liu2018darts}, data hypercleaning~\cite{shaban2019truncated}, and reinforcement learning~\cite{wu2020finite}. Some early works \cite{aiyoshi1984solution,edmunds1991algorithms} transformed the bilevel problem into a single-level optimization problem by replacing the lower-level problem (Eq.~\eqref{eqn:main:dblo}) with its optimality conditions. Recently, gradient-based approaches designed for the original bilevel problem have gained popularity due to their simplicity and effectiveness \cite{chen2021closing,ghadimi2018approximation,liu2021investigating,Ji2020ProvablyFA}. In an attempt to move away from traditional bilevel optimization, \cite{tarzanagh2022fednest} proposed a federated bilevel optimization problem, where the inner and outer objectives are distributed over a \textit{star-shaped} network (Eq.~\eqref{eqn:main:dblo} and Figure~\ref{fig:1}(a)). This approach employs \textit{local updates} between two communication rounds, and clients (nodes) only need to compute matrix-vector products and exchange vectors. Although federated bilevel optimization can achieve non-federated iteration complexity using only vector operations, it may suffer from high communication costs on the parameter (central) server.

Decentralized bilevel optimization focuses on solving bilevel problems within a decentralized framework. This approach offers additional advantages such as accelerated convergence, safeguarding data privacy, and resilience to limited network bandwidth compared to centralized settings or single-agent training \cite{lian2017can}; see Figure \ref{fig:1}. 
In addition to their communication efficiency, decentralized bilevel algorithms exhibit increased robustness against node and link failures, maintaining convergence toward the desired solution as long as the network remains connected.
In contrast, centralized bilevel algorithms inevitably fail when the central server crashes. An instance of this, decentralized meta-learning, naturally emerges in scenarios like medical data analysis, particularly in safeguarding patient privacy. Relevant studies include examples by \cite{zhang2019metapred}, \cite{altae2017low}, and \cite{kayaalp2022dif}.
Indeed, decentralized bilevel optimization serves as a backbone for numerous decentralized learning paradigms across peer-to-peer networks. These paradigms include multi-agent adaptations of meta-learning \cite{liu2021boml,rajeswaran2019meta}, hyperparameter optimization \cite{mackay2019self,okuno2021p}, solving area under curve problems \cite{liu2019stochastic,qi2021stochastic}, and reinforcement learning \cite{hong2020two,zhang2020bi}. Recent work has addressed the problem of solving finite sum bilevel programming (Eq. \eqref{eqn:main:dblo}) over a \textit{decentralized} network architecture \cite{chen2022decentralized,yang2022decentralized}. However, existing methods tend to be complicated and impractical for large-scale bilevel applications. Specifically, the proposed algorithms \cite{chen2022decentralized,yang2022decentralized} involve expensive Hessian computations within nodes and/or communication of matrices between nodes; refer to Section~\ref{sec:Related work} for further discussion).

This paper addresses these challenges and presents algorithms and associated theory for \textit{fast and communication-efficient} decentralized bilevel optimization. The main contributions are:
\begin{enumerate}[label=$\bullet$, wide, labelwidth=!,itemindent=!, labelindent=1pt]
     \item \textbf{Lightweight Computation and Communication via Penalized Optimization:} Development of a Decentralized Alternating Gradient Method (DAGM) for solving bilevel problems with lightweight decentralized communication and computation. DAGM approximates the original consensus problem using a penalized reformulation and estimates its hyper-gradient (the gradient of the outer function) through a Decentralized Inverse Hessian-Gradient-Product (DIHGP) using \textit{local matrix-vector products} and \textit{decentralized communication of vectors} . The DIHGP approach employs a few terms in the Neumann series to provide a stable inverse Hessian approximation without explicitly instantiating any matrices, utilizing efficient vector-Jacobian products similar to \cite{lorraine2020optimizing}, but in a decentralized setting.
    \item  \textbf{Iteration Complexity and Acceleration:}
From a theoretical perspective, we establish convergence rates and communication complexity bounds achievable by DAGM for smooth strongly convex, convex, and non-convex bilevel problems. Remarkably, the iteration complexity of DAGM achieves a linear acceleration (an $n^{-1}$ factor in the complexity bound) even with vector communication, in comparison with extensive matrix computation/communication results in \cite{chen2022decentralized,yang2022decentralized}; see Table~\ref{tab:1}.
\item \textbf{Experimental Evaluation:} From a practical perspective, we assess DAGM's performance in addressing large-scale problems and pinpoint the essential elements contributing to the resilience and scalability of DIHGP. As far as we know, this marks the initial empirical and theoretical exploration of Neumann series-based DIHGP, featuring vector costs, within the realm of bilevel optimization.
 \end{enumerate}
 \textbf{Organization:} Section \ref{sec:Related work} discusses prior work on the topic. Section \ref{sec:Preliminaries and Notations} presents the problem formulation and assumptions. Section \ref{sec:A Decentralized Alternating Method} introduces the DAGM algorithm, and Section \ref{sec:convergence} establishes its theoretical results. Section \ref{sec:numerical results} presents simulation results demonstrating the improved convergence speed of DAGM. Finally, Section \ref{sec:conclusion} provides concluding remarks.

 \noindent\textbf{Notation.} For a smooth function $h(\xx,\yy):\mb{R}^{d_1}\times\mb{R}^{d_2}\rightarrow\mb{R}$ in which $\yy=\yy(\xx):\mb{R}^{d_1}\rightarrow\mb{R}^{d_2}$, we denote $\nabla{h}\in\mb{R}^{d_1}$ the gradient of $h$ as a function of $\xx$ and  $\nabla_{\xx}{h}$, $\nabla_{\yy} h$ the partial derivatives of $h$ with respect to $\xx$ and $\yy$, respectively. $\nabla_{\m{x}\m{y}}^2 h$ and $\nabla_{\yy}^2 h$ denote the Jacobian matrix and the Hessian matrix of $h$, respectively. A mapping $h$ is $L$-Lipschitz continuous if and only if (iff) for some $L\in \mathbb{R}_{+}$, $\|h(\mathrm{x})-h(\mathrm{y}) \|\leq L \|\mathrm{x}-\mathrm{y} \|$, $\forall \mathrm{x},\mathrm{y}\in \mathbb{R}^d$. Also, it is said to be $\mu$-strongly convex iff for some $\mu\in \mathbb{R}_{+}$, $h(\xx)\geq h(\yy)+\nabla h(\yy)^{\top}(\xx-\yy)+\frac{\mu}{2}\|\xx-\yy\|^2$,  $\forall (\xx,\yy) \in  \mb{R}^{d_1}\times\mb{R}^{d_2}$.  Additional notation is defined when required, but for reference purposes the reader can also find it summarized in Table~\ref{table:summary:notation}.
\section{Related Work }\label{sec:Related work}

\textbf{Bilevel Optimization.} Initially introduced by \cite{stackelberg1952theory} in game theory, bilevel optimization involves a hierarchy of decision-making actions, reflecting a Stackelberg game. Recent advancements \cite{finn2017model,franceschi2017forward,ghadimi2018approximation,grazzi2020iteration,pedregosa2016hyperparameter,Ji2020ProvablyFA} include iterative techniques such as gradient-based updates and implicit differentiation (AID) to solve bilevel problems. Works such as  \cite{finn2017model,franceschi2017forward} use gradient descent with iterative differentiation (ITD), while \cite{domke2012generic,ghadimi2018approximation} utilize AID, solving a linear system for the hypergradient. The convergence and efficiency of these methods have been studied \cite{franceschi2018bilevel,ghadimi2018approximation,shaban2019truncated}, with \cite{franceschi2018bilevel,Ji2020ProvablyFA} showing asymptotic convergence for ITD-based algorithms and \cite{ghadimi2018approximation,Ji2020ProvablyFA} providing finite-time convergence for AID-based approaches.
 The efficacy of penalty approaches in bilevel optimization was first studied in \cite{ye1997exact,white1993penalty}. Subsequent studies by \cite{liu2021value} and \cite{mehra2021penalty} introduced methods using log-barrier and gradient norm techniques, demonstrating asymptotic convergence.  \cite{lu2023first} developed a penalty method for bilevel problems with a convex lower level, achieving convergence to a weak KKT point but not examining further relationships. More recently, \cite{shen2023penalty} established the finite-time convergence for constrained bilevel problems without lower-level strong convexity.

\noindent \textbf{Bilevel Optimization over Star-Shaped Networks.} Optimization methods of this type were first developed in \cite{gao2022convergence,li2022local} and \cite{tarzanagh2022fednest} in \textit{homogeneous} and general \textit{heterogeneous} federated settings, respectively; see Figure~\ref{fig:1a}. Recent work has enhanced the complexity and communication efficiency of these methods through the use of momentum, variance reduction, and hypergradient estimation \cite{huang2022fast,huang2023achieving,li2024communication,yang2024simfbo}. Although they enable decentralized computation, as data are localized on each node, they may suffer from high communication costs involving the central (parameter) server.

\textbf{Bilevel Optimization over Decentralized Networks.} This type of distributed optimization was first proposed in \cite{lu2022decentralized,lu2022stochastic}; see Figure~\ref{fig:1b}. These methods inherit the computational and communication properties of \textit{single-level} decentralized gradient descent \cite{lan2017communication,nazari2019adaptive,nazari2019dadam,nedic2009distributed,shi2015extra,yuan2016convergence,zeng2018nonconvex}. Specifically, \cite{lu2022decentralized,lu2022stochastic} considered a variant of Problem \eqref{eqn:main:dblo}, where only the lower-level objective involves distribution over many nodes. \cite{lu2022stochastic} proposed a stochastic linearized augmented Lagrangian method. Our work is closely related to \cite{chen2022decentralized,yang2022decentralized}, where the authors studied the general bilevel Problem \eqref{eqn:main:dblo} and showed that their proposed algorithms enjoy $\mathcal{O}(\epsilon^{-2})$ sample complexity. However, these algorithms involve expensive Hessian computations within nodes and/or communication of matrices between agents. Indeed, \cite{chen2022decentralized} proposes a JHIP oracle for estimating the Jacobian-Hessian-inverse product, while \cite{yang2022decentralized} introduces a subroutine for estimating the Hessian-inverse using the Neumann series approach. However, both approaches necessitate computing the entire Jacobian or Hessian matrices, which becomes complex in both
computation and communication and highly time-consuming when $d_2$ is large; refer to Table~\ref{tab:complexity_comparison}.

Gossip-type methods \cite{yang2022decentralized} also require bidirectional communication of matrices between the agents; each iteration involves one ``gossip" round where all $m$ edges communicate bidirectionally, thus necessitating waiting for the slowest communication (out of the $2m$ ones) to be completed.
The VRDBO method in \cite{gao2022stochastic} utilized momentum-based techniques for decentralized bilevel optimization, yielding superior complexity results of $\mathcal{O}(\epsilon^{-1.5})$.
The Prometheus algorithm in \cite{liu2022prometheus} achieves $\mathcal{O}(\epsilon^{-1})$ sample complexity for constrained decentralized bilevel optimization, which is a
near-optimal sample complexity and outperforms existing decentralized bilevel algorithms. Following these methods, a variety of distributed bilevel algorithms, offering both theoretical guarantees and empirical effectiveness have been developed; see
\cite{chen2023decentralized,dong2023single,ji2023network,jiao2022asynchronous,kong2024decentralized,liu2022interact,qiu2023diamond,yousefian2021bilevel,zhang2023communication}.

However, existing decentralized bilevel optimization methods still suffer from the additional challenges posed by the complexity of nested problems and hyper-gradient computation and communication. To address these challenges, this paper introduces a penalized reformulation of the bilevel Problem \eqref{eqn:main:dblo}. This enables the application of a standard alternating gradient-type optimization approach using the Neumann series that exclusively requires matrix-vector products and constrained vector communications. The Neumann series framework offers a decentralized algorithm that exhibits enhanced speed and scalability \cite{lorraine2020optimizing,tarzanagh2022fednest}, and it is capable of accommodating millions of parameters. A distinguishing aspect of this work, unlike \cite{chen2022decentralized,yang2022decentralized}, lies in achieving linear acceleration (an $n^{-1}$ improvement in complexity) even when involving vector computation and improved communication as provided in Table \ref{tab:complexity_comparison}. 

\begin{table*}[t]
 \centering
\begin{tabular}{|c|cc|c c|}
  \hline
Algorithm & & Stationary Measure  & Complexity &\\
\hline
 \multicolumn{4}{c}{\textbf{Strongly convex}}
 \\
\hline
  BA  \cite{ghadimi2018approximation} & &  & $\mathcal{O}\left(\log^2 \epsilon^{-1}\right)$&
\\
  AccBio \cite{ji2021lower}& & $f(\xx_{K},\yy^*(\xx_{K}))-{f}^*$ &  $\mathcal{O}\left(\log^2 \epsilon^{-1}\right)$&
\\
  AmIGO \cite{arbel2021amortized}& &  &  $\mathcal{O}\left(\log \epsilon^{-1}\right)$&
 \\
\hline
 \textbf{DAGM} (Decentralized)  &  & $\frac{1}{n}1^{\top}\mathbf{f}(\bar\x_{K},\y^*(\bar\x_{K}))-\frac{1}{n}1^{\top}\mathbf{f}^*$ & $\mathcal{O}\left( \log (n^{-1}\epsilon^{-1}/(1-\sigma))\right)$&
\\ \hline
 \multicolumn{5}{c}{\textbf{Convex} }\\
  \hline
 BA  \cite{ghadimi2018approximation} & & $f(\widehat{\xx}_{K},\yy^*(\widehat{\xx}_{K}))-f^*$ & $\mathcal{O}\left( \epsilon^{-5/4}\right)$&\\
AccBiO \cite{ji2021lower}& &  &  $\mathcal{O}\left( \epsilon^{-1}\right)$&
\\ \hline
  \textbf{DAGM} (Decentralized)  & &$\frac{1}{n}1^{\top}\mathbf{f}(\widehat{\x}_{K},\y^*(\widehat{\x}_{K}))-\frac{1}{n}1^{\top}\mathbf{f}^*$  & $\mathcal{O}(n^{-2} \epsilon^{-2}/(1-\sigma))$&
\\
\hline
 \multicolumn{5}{c}{\textbf{Non-convex  }}
 \\
\hline
BA  \cite{ghadimi2018approximation}  & &  & $\mathcal{O}\left(\epsilon^{-5/4}\right)$&
\\
 AID-BiO \cite{ji2021bilevel} & & $\frac{1}{K}\sum_{k=0}^{K-1}\|\nabla f (\xx_k,\yy^*(\xx_k))\|^2$ & $\mathcal{O}\left(\epsilon^{-1}\right)$&
\\
AmIGO \cite{arbel2021amortized}& &  & $\mathcal{O}\left(\epsilon^{-1}\right)$&
\\
DBO \cite{chen2022decentralized} (Decentralized)& &  & $\mathcal{O}\left(\epsilon^{-1}/(1-\sigma)^2\right)$&
\\
 \hline
 \textbf{DAGM} (Decentralized)  &  & $\frac{1}{K}\sum_{k=0}^{K-1}\|\frac{1}{n}1^{\top}\nabla\mathbf{f}(\bar\x_k,\y^*(\bar\x_k))\|^2$ &$\mathcal{O}\left((n^{-2}\epsilon^{-2})+(n^{-1}\epsilon^{-1}/(1-\sigma)^2)\right)$&
\\
\hline
    \end{tabular}
  \caption{Comparisons between different \textbf{deterministic bilevel  optimization} approaches with computation to achieve an $\epsilon$-stationary point in the non-convex setting and an $\epsilon$-optimal solution in the convex setting. Here,
$\bar\x_k :=(1/n)\sum_{i=1}^n\xx_{i,k}$, $\widehat{\x}_{K}:=(1/K)\sum_{k=1}^K \bar\x_k $, and $\sigma$ denotes the spectral gap of the network topology, with $n$ being the number of nodes.   $f$ and $f^*$ represent the main objective function and its optimal value, respectively, while $\mathbf{f}$ denotes the concatenation of the local objective functions $f_i$, and $\mathbf{f}^*$ denotes its optimal value; see Section \ref{sec:Preliminaries and Notations}.}
\label{tab:1}
 \end{table*}

 \section{Penalty-Based Decentralized  Bilevel Optimization}\label{sec:Preliminaries and Notations}
Consider the bilevel optimization problem \cite{tarzanagh2022fednest}:
\begin{subequations}\label{eqn:main:dblo}
\begin{align}
 &  \m{x}^* \in \argmin_{\xx\in \mathbb R^{d_1}} \frac{1}{n}\sum_{i=1}^n f_i\left(\xx,\yy^*(\xx)\right),& \label{pr: BLO1}\\
& \qquad \mbox{subj. to} \quad  \yy^*(\xx) \in \argmin_{\yy\in \mathbb{R}^{d_2}} \frac{1}{n}\sum_{i=1}^n g_i\left(\xx,\yy\right).&\label{pr: BLO2}
\end{align}
\end{subequations}
In bilevel optimization over a decentralized network \cite{chen2022decentralized,yang2022decentralized}, $n$ agents form a connected undirected network $\mathcal{G}=\{\mathcal{V},\mathcal{E}\}$, and cooperatively solve Problem \eqref{eqn:main:dblo}. Note that $\mathcal{V}=\{1,\ldots,n \}$ denotes the set of agents, $\mathcal{E} \subseteq \mathcal{V}\times \mathcal{V}$ the set of edges in the network, and $f_i$, $g_i$ are the local objective functions available only to agent $i$; a depiction is given in Figure~\ref{fig:1}(b). For each $i$, we denote by $\mathcal{N}_i$ the set of \textit{neighbors} of agent $i$ in the underlying network, i.e., $j \in \mathcal{N}_i$, if and only if $(i,j)\in \mathcal{E}$. Each edge $(i, j) \in \mathcal{E}$ has an associated weight $w_{ij}\geq 0$, which measures how much agent $i$ values the information received by agent $j$. We impose the following assumptions on the network matrix $\mr W$ and on the inner and outer objective functions. 

\begin{assumption}\textnormal{(\textbf{Mixing matrix})}\label{assu:netw}
The  nonnegative symmetric matrix $\mr W=[w_{ij}]\in \mathbb{R}^{n\times n}_{+}$ associated with network $\mathcal{G}$ encodes its connectivity structure so that:
\begin{enumerate}[label={\textnormal{\textbf{{A}\arabic*.}}}]
    \item \label{assu:netw:item1}
     $w_{ij}=0$ if agents $i$ and $j$ are not connected.
    \item \label{assu:netw:item2} $\mr W$ is doubly stochastic, i.e., $\mr W1_n=1_n$ and $\mr W^{\top}1_n=1_n$.
    \item  \label{assu:netw:item3} $\textnormal{null}\{\mr I_n-\mr W\}=\textnormal{span}\{1_n\}$.
    \item \label{assu:netw:item4} There exist positive scalars $\theta$  and $\Theta$ with $0< \theta \leq \Theta<1$, such that for all $i\in \mathcal{V}:\,\,\theta  \leq w_{ii} \leq \Theta $.
\end{enumerate}
\end{assumption}

Assumption \ref{assu:netw} is widely used in the decentralized optimization literature \cite{boyd2004fastest,nedic2018network,nedic2009distributed}. The mixing matrix $\mr W$ that satisfies this assumption can be constructed using the Metropolis method \cite{shi2015extra,xiao2004fast}; for more detailed information, refer to \eqref{eq:mahan}. Under Assumptions~\ref{assu:netw:item1}--\ref{assu:netw:item3}, the Perron-Frobenius theorem \cite{pillai2005perron} implies that the eigenvalues of $\mr W$ lie in $(-1,1]$, the multiplicity of eigenvalue $1$ is one, and
\begin{equation}\label{eqn:mixing}
\sigma:=\|\mr W-\frac{1}{n}1_n 1_n^{\top}\|=\max\{|\lambda_2| ,|\lambda_n| \} \in (0,1),
\end{equation}
where $\sigma$ is the mixing rate of the network and $\lambda_n\leq \lambda_{n-1}\leq \dots \leq \lambda_2 \leq \lambda_1$ are the eigenvalues of $\mr W$.
{Assumption \ref{assu:netw:item4} is fulfilled by several common consensus matrices \cite{nedic2009distributed,tsitsiklis1986distributed,xiao2005scheme}.
\begin{exmp}
For the Maximum-degree weights configuration, we assign a uniform weight of $\frac{1}{n}$ to all edges and adjust self-weights to ensure each node's incoming weights sum to 1:
    \begin{equation*}
    w_{ij} = \begin{cases}
    \frac{1}{n}, & \text{if } \{i,j\} \in \mathcal{E}, \\
    1 - \frac{\deg(i)}{n}, & \text{if } i = j,\\
    0, & \text{otherwise},
    \end{cases}
    \end{equation*}
    where $\deg(i)$ denotes the degree of node $i$. The maximum degree, $\deg(i) = n-1$, occurs when a node is connected to all others. Conversely, the minimum degree is 1 (assuming every node has at least one connection). Consequently, for $n > 2$, $\theta = {1}/{n}$ and $\Theta = 1 - (1/n)$. The designed weight matrix is symmetric and doubly stochastic by construction \cite{sayed2014diffusion}.
\end{exmp}
\begin{exmp}
The Metropolis weights are given by  \cite{boyd2004fastest}
     \begin{equation*}
    w_{ij} = \begin{cases}
    \frac{1}{1+\max\{\degree(i),\degree(j)\}}, & \text{if } \{i,j\} \in \mathcal{E}, \\
    1-\sum\limits_{\{i,k\}\in\mathcal{E}} w_{ik}, & \text{if } i = j,\\
    0, & \text{otherwise},
    \end{cases}
    \end{equation*}
 where $\deg(i)=| \mathcal{N}_i|$ is the degree of agent $i$. The maximum degree, $\deg(i) = n-1$, occurs when a node is connected to all other ones. Conversely, the minimum degree is 1 (assuming every node has at least one connection). Consequently, $\Theta = 1  -(1/n)$, since
$  w_{ii}\leq 1-(1/n)$. The resulting weight matrix is doubly stochastic.
\end{exmp}

\begin{assumption}\label{assu:lip}
For all $i\in \mathcal{V} :$
\begin{enumerate} [label={\textnormal{\textbf{{B}\arabic*.}}}]
\item \label{assu:f:L} For any $\xx\in \mathbb{R}^{d_1}$, $\nabla_{\xx} f_i(\xx,\cdot)$, $\nabla_{\yy} f_i(\xx,\cdot)$, $\nabla_{\yy} g_i(\xx,\cdot)$,
$\nabla^2_{\xx \yy} g_i(\xx,\cdot)$, $\nabla^2_{\yy} g_i(\xx,\cdot)$
are $L_{f_{\xx}}$, $L_{f_{\yy}}$, $L_{g}$, $L_{g_{\xx\yy}}$, $L_{g_{\yy\yy}}$-Lipschitz continuous  for some constants $L_{f_{\xx}}>0$, $L_{f_{\yy}}>0$, $L_{g}>0$, $L_{g_{\xx\yy}}>0$ and $L_{g_{\yy\yy}}>0$.
\item For any $\yy\in \mathbb{R}^{d_2}$, $\nabla_{\yy} f_i(\cdot,\yy)$, $\nabla^2_{\xx \yy} g_i(\cdot,\yy)$,  $\nabla_{\yy}^2 g_i(\cdot,\yy)$ are $\tilde{L}_{f_{\yy}}$, $\tilde{L}_{g_{\xx\yy}}$, $\tilde{L}_{g_{\yy\yy}}$-Lipschitz continuous.
\item \label{assu:f:grad} For any $\xx\in \mathbb{R}^{d_1}$ and $\yy\in \mathbb{R}^{d_2}$, we have $\| \nabla_{\yy} f_i(\xx,\yy)\|\leq C_{f_{\yy}}$ and $\| \nabla_{\xx} f_i(\xx,\yy)\|\leq C_{f_{\xx}}$ for some positive constants $C_{f_{\yy}}$ and $C_{f_{\xx}}$.
\item \label{assu:g:Cgxy} For any $\xx\in \mathbb{R}^{d_1}$ and $\yy\in \mathbb{R}^{d_2}$, we have  $\nabla^2_{\xx \yy} g_i(\xx,\yy)\preceq C_{g_{\xx\yy}} \mr I_{d_2}$ for some positive constant $C_{g_{\xx\yy}}$.
\item \label{assu:g:strongly} For any $\xx\in \mathbb{R}^{d_1}$,
the inner function $g_i(\xx,\cdot)$ is $\mu_{g}$-strongly convex and $C_{g_{\yy\yy}}$-smooth for some constants $0<\mu_{g} \leq C_{g_{\yy\yy}} < \infty$, i.e., $\mu_{g} \mr I_{d_2}\preceq\nabla^2_{\yy} g_i(\xx,\yy)\preceq  C_{g_{\yy\yy}} \mr I_{d_2}$.
\end{enumerate}
\end{assumption}
Assumption \ref{assu:lip} is widely used in bilevel optimization research \cite{chen2021closing,ghadimi2018approximation,hong2020two,ji2021bilevel}. It shows the smooth behavior of first- and second-order derivatives for the local functions $f_i(\xx,\yy)$ and $g_i(\xx,\yy)$, as well as for the solution mapping $\yy^*(\xx)$.
Typically, problems such as hyperparameter learning, where each client has local validation and training datasets associated with objectives $(f_i, g_i)_{i=1}^m$ representing validation and training losses respectively, fit these criteria. The goal is to find hyperparameters $\xx$ that lead to learning model parameters $\yy$ that minimize the global validation loss. In cases where the validation and training losses are smooth, such as cross-entropy, logistic loss, or exponential losses, Assumption \ref{assu:lip} holds. Further discussions and proofs
can be found in Lemma \ref{lem:lip}\eqref{L2} of the Appendix.

\subsection{A Concatenated Formulation and its Properties}\label{sec:Concatenation Formulation}
As a prelude to understanding the key properties of bilevel gradient descent in a decentralized setting, we begin by providing a concatenation formulation of Eq. \eqref{eqn:main:dblo}. Let $\mathrm{x}_i\in \mathbb{R}^{d_1}$ and $\mathrm{y}_i\in \mathbb{R}^{d_2}$ denote the local copy of $\mathrm{x}$ and $\yy$ at node $i$. Let
\begin{equation*}
\begin{split}
    \x &:= [\mathrm{x}_1; \ldots ; \mathrm{x}_n]\in \mathbb{R}^{nd_1}, \\
    \x^* &:= [\mathrm{x}_1^*; \ldots ; \mathrm{x}_n^*]\in \mathbb{R}^{nd_1}, \\
    \y &:= [\yy_{1}; \ldots ; \yy_n]\in\mathbb{R}^{nd_2},\\
    \y^*(\x)&:=[\yy_1^*(\xx_1);\ldots;\yy_n^*(\xx_n)]\in \mathbb{R}^{nd_2},\\
  \mathbf{g}(\x,\y)&:=[ g_1(\xx_1,\yy_1) ;\ldots;g_n(\xx_n,\yy_n)]\in\mathbb{R}^{n d_2},\\
      \mathbf{f}(\x,\y^*(\x))&:=[ f_1(\xx_1,\yy_1^*(\xx_1));\ldots; f_n(\xx_n,\yy_n^*(\xx_n))]\in\mathbb{R}^{nd_1}.
\end{split}
\end{equation*}
Since the network $\mc{G}$ is connected, the bilevel problem \eqref{eqn:main:dblo} is equivalent to
\begin{subequations}\label{eqn:obj:cbo}
\begin{align}
&\x^* \in \underset{\{\xx_i\}_{i=1}^n }{\argmin}~ \frac{1}{n}\sum_{i=1}^n f_i\left(\xx_i,\yy_i^*(\xx_i)\right)
\label{eqn:obj:cbo:out}\\
\nonumber
& \quad\text{subj.~to} \quad \xx_i=\xx_j, \quad \forall j\in  \mathcal{N}_i,\quad \forall i, \nonumber \\
\vspace{-1cm}
& \quad      \y^*(\x) \in \underset{\{\yy_i\}_{i=1}^n }{\argmin}~~\frac{1}{n}\sum_{i=1}^n g_i\left(\xx_i,\yy_i\right) \label{eqn:obj:cbo:inn}\\
& \quad  ~\textnormal{\text{subj.~to}~~}  \yy_i=\yy_j, \quad \forall j\in  \mathcal{N}_i,\quad \forall i.\nonumber
\end{align}
\end{subequations}
The following lemma establishes the above claim and characterizes a penalized version of \eqref{eqn:obj:cbo}, which is more suitable for decentralized implementation.
\begin{lm}[Penalized Bilevel Problem]\label{lem: equivalent}
Under \ref{assu:netw:item1}- \ref{assu:netw:item3},
\begin{enumerate}
\item \label{itm:eqi:itm1}  The bilevel problem in \eqref{eqn:main:dblo} is equivalent to \eqref{eqn:obj:cbo}.
\item \label{itm:eqi:itmw}
For any given penalty parameters $\beta>0$ and $\alpha>0$, the penalized problem associated with \eqref{eqn:obj:cbo} is given by
\begin{subequations}\label{eqn:reform:dblo}
\begin{align}\label{eqn:approximate:prob}
\nonumber& \check{\x}^* \in \argmin_{\x\in \mathbb{R}^{nd_1}}~~ \mathbf{F}(\x,\check{\y}^*(\x)):=\frac{1}{2\alpha }{\x}^\top (\textbf{I}_{nd_1}-\acute{\W}){\x}
\\&\qquad\qquad \qquad \qquad \qquad ~~~~+ 1^{\top}\mathbf{f}(\x,\check{\y}^*(\x))\\\nonumber
& ~\textnormal{s.t.} \quad \check{\y}^*(\x)\in \argmin_{\y\in \mathbb{R}^{nd_2}}\big\{\mathbf{G}(\x,\y):=\frac{1}{2\beta}{\y}^\top (\textbf{I}_{nd_2}-\W){\y}
\\&\qquad \qquad \qquad \qquad \qquad  \qquad~+1^{\top}\mathbf{g}(\x,\y)\big\},\label{eqn:approximate:y}
\end{align}
\end{subequations}
where $\W:=\mr W\otimes \mr I_{d_2} \in \mathbb{R}^{nd_2 \times nd_2}$ and $\acute{\W}:=\mr W\otimes \mr I_{d_1} \in \mathbb{R}^{nd_1 \times nd_1}$ are the \textit{extended} mixing matrices;  $1^{\top}\mathbf{f}(\x,\check{\y}^*(\x))=\sum_{i=1}^n f_i\left(\xx_i,\check{\yy}_i^*(\xx_i)\right)$ and $1^{\top}\mathbf{g}(\x,\y)=\sum_{i=1}^{n}g_i(\xx_i,\yy_i)$.
\end{enumerate}
\end{lm}
Lemma~\ref{lem: equivalent} allows us to compute the inner/outer gradients by exchanging information between neighboring nodes. Specifically, it allows to develop an efficient algorithm to approximate the hyper-gradient via decentralized computation of {matrix-vector} products and few {vector communications}.
Throughout this paper we also assume that the sets of minimizers of the original problem in Eq.
\eqref{eqn:main:dblo} and the penalized consensus problem in Eq. \eqref{eqn:reform:dblo} are nonempty.
\\
Before providing the formal statement of the algorithm, we highlight the inherent challenge of directly applying the decentralized gradient method to the bilevel problem in Eq. \eqref{eqn:reform:dblo}.
To elucidate this point, we deduce the gradient of the outer objective function $\mathbf{F}(\x,\check{\y}^*(\x))$
in the subsequent lemma.

\begin{lm}\label{lem:grad}
Suppose Assumptions \ref{assu:netw} and \ref{assu:g:strongly} hold.
\begin{enumerate}
\item For any $\x\in \mathbb{R}^{nd_1}$, $\check{\y}^*(\x)$ is unique and differentiable, and
\begin{equation}\label{eqn:grad:y}
\nabla \check{\y}^*(\x)=- \beta \nabla^2_{\x \y} \mathbf{g}\left(\x,\check{\y}^*(\x)\right) [\textbf{H}(\x,\check{\y}^*(\x))]^{-1},
\end{equation}
where $\textbf{H}(\x,\check{\y}^*(\x)):=(\textbf{I}_{n d_2}-\W)+\beta \nabla_{\y}^2 \mathbf{g}(\x,\check{\y}^*(\x))$.
\item For any $\x\in \mathbb{R}^{nd_1}$, the gradient of $\mathbf{F}$ as a function of $\x$, is given by
\begin{align}\label{eqn:grad:F}
\nonumber \nabla \mathbf{F}(\x,\check{\y}^*(\x))&=\frac{1}{\alpha}(\textbf{I}_{n d_1}-\acute{\W})\x+\nabla_{\x} \mathbf{f}(\x,\check{\y}^*(\x))
\\&+ \nabla \check{\y}^*(\x) \nabla_{\y}\mathbf{f}\left(\x,\check{\y}^*(\x)\right).
\end{align}
\end{enumerate}
\end{lm}

Next, we utilize the above lemma to describe the core difficulties of developing a decentralized optimization method for solving \eqref{eqn:reform:dblo}.  In order to develop a decentralized bilevel gradient-type method, one needs to estimate the outer gradient $\nabla \mathbf{F}(\x,\check{\y}^*(\x))$. Note that the inner optimizer $\check{\y}^*(\x)$ is not available in general. To circumvent this bottleneck, following existing non-decentralized works \cite{ghadimi2018approximation,hong2020two}, we replace $\check{\y}^*(\x)$ with an approximation $\y\in \mathbb{R}^{n d_2}$ and define the following surrogate for \eqref{eqn:grad:F}:
\begin{align}\label{eqn:grad:tildF}
\nonumber
\tilde{\nabla} \mathbf{F}(\x,\y)&=\frac{1}{\alpha}(\textbf{I}_{n d_1}-\acute{\W})\x+\nabla_{\x} \mathbf{f}(\x,\y)
\\&+\beta \nabla^2_{\x \y} \mathbf{g}\left(\x,\y\right)  \textbf{h},
\end{align}
where
\begin{align}\label{eqn:tru:hgp}
\nonumber\textbf{h}&:=-  [\textbf{H}(\x,\y)]^{-1}\nabla_{\y} \mathbf{f}(\x,\y),~~~\textnormal{and}
\\\textbf{H}(\x,\y)&:=(\textbf{I}_{n d_2}-\W)+\beta \nabla_{\y}^2 \mathbf{g}(\x,\y).
\end{align}

Here, $\nabla_{\y}^2 \mathbf{g}(\x,\y)\in \mathbb{R}^{nd_2\times nd_2}$ is a block diagonal matrix where its $i$-th diagonal block is given by the $i$-th local Hessian  $\nabla_{\yy}^2 g_i(\xx_{i}, \yy_{i})\in \mathbb{R}^{d_2\times d_2}$. Similarly,
the matrix $\nabla^2_{\x\y} \mathbf{g} \left(\x,\y\right) \in \mathbb{R}^{nd_1\times nd_2}$ is a block diagonal matrix where its $i$-th diagonal block is given by
$
\nabla^2_{\xx \yy}g_i(\xx_{i}, \yy_{i})\in \mathbb{R}^{d_1 \times d_2}$. Further, the matrix $\textbf{I}_{n d_2}-\W$ is a \textit{block neighbor sparse} in the sense that the $(i,j)$-th block is non-zero if and only if $j\in \mathcal{N}_i$ or $j = i$.
Consequently, the Hessian matrix $\textbf{H}(\x,\y)$ follows the sparsity pattern of the graph $\mathcal{G}$ and can be computed by exchanging information with neighboring nodes. Specifically, the $i$-th diagonal block is  $\textbf{H}_{ii}(\x,\y)=(1-w_{ii})\textbf{I}_{n d_2}+\beta\nabla_{\yy}^2 g_i(\xx_{i}, \yy_{i})$ and the  $(i,j)$-th off diagonal block is $\textbf{H}_{ij}(\x,\y)=w_{ij}\textbf{I}_{n d_2}$ when $j\in \mathcal{N}_{j}$ and $\textbf{0}$ otherwise.

\section{DAGM Algorithm}\label{sec:A Decentralized Alternating Method}
We develop a communication-efficient Decentralized Alternating Gradient Method (DAGM) that relies on Hessian splitting and Decentralized estimation of the Inverse Hessian-Gradient Product (DIHGP) for solving the bilevel problem presented in Eq. \eqref{eqn:reform:dblo}.
\subsection{Hessian Splitting and DIHGP}\label{spl}

Note that the formulation of Eq. \eqref{eqn:tru:hgp} does not guarantee the sparsity of the inverse Hessian $[\textbf{H}(\x,\y)]^{-1}$ in the decentralized setting. To overcome this issue, we represent the Hessian inverse as a convergent series of matrices, where each term can be computed using local information. In particular, we approximate the inverse Hessian-gradient-product $\textbf{h}$ defined in \eqref{eqn:tru:hgp} by a \textit{local} Neumann series approximation.  To do so, inspired by \cite{mokhtari2016network,zargham2013accelerated}, we split the Hessian into matrices $\textbf{D}$ and  $\textbf{B}$ as follows:
\begin{align}\label{hhh}
 \nonumber \textbf{H}&=(\textbf{I}_{nd_2}-\W)+\beta \nabla_{\y}^2 \mathbf{g}(\x,\y) \\\nonumber
    &=\underbrace{\beta \nabla_{\y}^2 \mathbf{g}(\x,\y)+2(\textbf{I}_{n d_2}-\textnormal{diag}(\W))}_{=:\textbf{D}}
    \\&-\underbrace{\left(\textbf{I}_{n d_2}-2\textnormal{diag}(\W)+\W\right)}_{=:\textbf{B}},
\end{align}
where $\textbf{D}$ is a block diagonal
positive definite matrix and $\textbf{B}$ is a neighbor sparse positive
semidefinite matrix.
 By factoring $\textbf{D}^{1/2}$ from both sides of Eq. \eqref{hhh}, we get
\begin{align*}
  \textbf{H}=\textbf{D}^{1/2}(\textbf{I}_{nd_2}-\textbf{D}^{-1/2}\textbf{B}\textbf{D}^{-1/2})\textbf{D}^{1/2}.
\end{align*}
Hence,
\begin{align*}
  \textbf{H}^{-1}=\textbf{D}^{-1/2}(\textbf{I}_{nd_2}-\textbf{D}^{-1/2}\textbf{B}\textbf{D}^{-1/2})^{-1}\textbf{D}^{-1/2}.
\end{align*}
In \cite{mokhtari2016network}, the authors present the following lemma ensuring that the spectral
radius of the matrix
$\textbf{D}^{-1/2} \textbf{B} \textbf{D}^{-1/2}$ is strictly less than 1.
\begin{lm}\label{symmetric_term_bounds11}
Under Assumptions  \ref{assu:netw} and \ref{assu:g:strongly},  we have
\begin{equation}
   0 \  \preceq \  \bbD^{-{1}/{2}}  \mB   \bbD^{-{1}/{2}} \ \preceq\ \rho\bbI_{nd_2},
\end{equation}
where $\rho:= 2(1-\theta)/(2(1-\Theta)+{\beta \mu_g} ) < 1$ and the constants $\theta, \Theta,\beta,$ and $\mu_g$  are defined in  Assumptions  \ref{assu:netw} and \ref{assu:lip}.
\end{lm}
Next, using Lemma~\ref{symmetric_term_bounds11}, we consider the Taylor's series $(\mr I-\mr X)^{-1}=\sum_{j=0}^{\infty} \mr X^j$ with $\mr X=\textbf{D}^{-1/2} \textbf{B} \textbf{D}^{-1/2}$ to write the Hessian inverse $\textbf{H}^{-1}$ as
\begin{equation}\label{olhij}
  \textbf{H}^{-1}=\textbf{D}^{-1/2}\sum_{u=0}^{\infty}(\textbf{D}^{-1/2} \textbf{B} \textbf{D}^{-1/2})^u\textbf{D}^{-1/2}.
\end{equation}
Note that the computation of the above series requires
global communication, which is not affordable in decentralized
settings. Hence, we consider the first $U+ 1$ (where $U\geq 0$) terms of the series for defining the approximate Hessian inverse as follows:
\begin{equation}\label{gh}
\hat{\textbf{H}}_{{(U)}}^{-1}=\textbf{D}^{-1/2}\sum_{u=0}^{U}(\textbf{D}^{-1/2} \textbf{B} \textbf{D}^{-1/2})^u\textbf{D}^{-1/2}.
\end{equation}
Since the matrix $\textbf{D}$ is block diagonal and $\textbf{B}$ is block neighbor sparse, it turns out that the approximate Hessian inverse  $\hat{\textbf{H}}_{{(U)}}^{-1}$ is $U$-hop block neighbor sparse, i.e., the $(i, j)$-th block is nonzero, if and only if there is at least one path between nodes $i$ and
$j$ with length $U$ or shorter.
Substituting the explicit expression for $\hat{\textbf{H}}_{{(U)}}^{-1}$ in \eqref{gh} into \eqref{eqn:tru:hgp}, we get
\begin{equation}\label{mkk}
 \textbf{h}_{(U)}=-\hat{\textbf{H}}_{{(U)}}^{-1}\p
 =-\textbf{D}^{-1/2}\sum_{u=0}^{U}(\textbf{D}^{-1/2} \textbf{B} \textbf{D}^{-1/2})^u\textbf{D}^{-1/2} \p,
\end{equation}
where $\p:= \nabla_{\y}\mathbf{f}(\x, \y)$. Then, Eq. \eqref{mkk} can be computed by the recursive expression as follows:
\begin{subequations}
\begin{equation}\label{bhb}
 \textbf{h}_{(s+1)}=\textbf{D}^{-1} \textbf{B} \textbf{h}_{(s)}-\textbf{D}^{-1} \p
 =\textbf{D}^{-1} (\textbf{B} \textbf{h}_{(s)}-\p).
\end{equation}
By using the sparsity pattern of $\textbf{B}$, we can further decompose \eqref{bhb} into local components as
\begin{align}\label{eqn:hh}
\mr h_{i,(s+1)}=\mr D_{ii}^{-1} (\sum_{j\in \mathcal{N}_i,j=i} \mr B_{ij} \mr h_{j,(s)}-\mathrm{p}_{i}).
\end{align}

\end{subequations}
Here, $\mr D_{ii}:=\beta \nabla_{\yy}^2 g_i(\xx_{i}, \yy_{i})+2(1_{d_2}-w_{ii})\mr I_{d_2}\in \mathbb{R}^{d_2}$
is the $i$-th diagonal block of matrix $\textbf{D}$ in \eqref{hhh}.
Note that the block $D_{ii}$ is locally available at node $i$.
Furthermore, the diagonal blocks $\mr B_{ii}=(1_{d_2}-w_{ii})\mr I_{d_2}$ and the off-diagonal blocks $\mr B_{ij}=w_{ij}\mr I_{d_2}$  can be computed and stored by node $i$.
The gradient component $\nabla_{\yy}f_i(\xx_{i}, \yy_{i})$ is also stored and computed at node $i$.
Therefore, node $i$ can execute
the recursion in \eqref{eqn:hh} by having access to the $ \mr h_{j,(s)}$ of its neighbors $j\in \mathcal{N}_i$.
The Decentralized  estimation of Inverse Hessian-Gradient-Product (DIHGP) method, executed by agent $i$, is summarized in Algorithm \ref{algo:estimate}.

\begin{algorithm}[t]
\begin{small}
\caption{
$\mr h_{i,(U)} =\pmb{\textnormal{DIHGP}}(\mr x_i,  \mr y_i, U,\mr W)$ at each node $i$
}
\begin{algorithmic}[1]
\label{algo:estimate}
\STATE  $\mr B_{ii}=(1_{d_2}-w_{ii})\mr I_{d_2}$ and $\mr B_{ij}=w_{ij}\mr I_{d_2}$.
\STATE  $\mr D_{ii}= \beta\nabla_{\yy }^2 g_i(\xx_{i}, \yy_{i})+2(1_{d_2}-w_{ii})\mr I_{d_2}$.
\STATE  $\mathrm{p}_{i}=\nabla_{\yy}f_i(\xx_{i}, \yy_{i})$.
\STATE Find $\mr h_{i,(0)}$ such that $\mr D_{ii} \mr h_{i,(0)} +\mathrm{p}_{i}=0$.
\FOR{$s=0, \dots, U-1$}
\STATE Exchange the iterate $\mr h_{i,(s)}$ with neighbors $j\in \mathcal{N}_i$.
\STATE Set $\mathrm{b}_i=-\mathrm{p}_{i} + \sum_{j\in \mathcal{N}_i,j=i}\mr B_{ij} \mr h_{j,(s)}$.
\STATE  Find $\mr h_{i,(s+1)}$ such that  $\mr D_{ii} \mr h_{i,(s+1)}= \mathrm{b}_i$.
\ENDFOR
 \end{algorithmic}
 \end{small}
\end{algorithm}

The following lemma provides bounds for
the eigenvalues of the $U$-th order inverse approximation $\hat{\mathbf H}_{(U)}^{-1}$ in \eqref{gh}, adopted from \cite{mokhtari2016network}.
\begin{lm}\label{Hessian_inverse_eigenvalue_bounds_lemma}
Suppose Assumptions  \ref{assu:netw} and \ref{assu:g:strongly} hold.
Then, the eigenvalues of the approximate Hessian inverse $\hat{\mathbf H}_{(U)}^{-1}$ are bounded as
\begin{equation*}
\lambda \bbI_{nd_2}\ \preceq\  \hat{\mathbf H}_{(U)}^{-1}  \preceq\   \Lambda \bbI_{nd_2},
\end{equation*}
where the constants $\lambda$ and $\Lambda$ are defined as
\begin{equation*}
\!\!\! \lambda\!:=\! \frac{1}{2(1-\theta)+\beta  C_{g_{\yy\yy}} }, \quad
\ \ \Lambda\! :=\! {\frac{1-\rho^{U+1}}{(1-\rho)((2(1-\Theta)+{\beta \mu_g} ) )}},
\end{equation*}
 $\rho$ is defined in Lemma \ref{symmetric_term_bounds11} and the remaining constants are defined in  Assumptions  \ref{assu:netw} and \ref{assu:lip}.
Moreover, let
$\bbE:= \bbI_{nd_2}- {\hat{\hbH}_{(U)}^{-{1}/{2}}} \bbH  {\hat{\hbH}_{(U)}^{-{1}/{2}}}$ be
the error of the Hessian inverse approximation. Then,
\begin{equation*}
    \mathbf 0\  \preceq\ \bbE\ \preceq\ \rho^{U+1}\bbI_{nd_2}.
\end{equation*}
\end{lm}
Lemma \ref{Hessian_inverse_eigenvalue_bounds_lemma} indicates that the error matrix $\bbE$ decreases exponentially as $U$ increases. Note that the matrix $\bbE$ measures the closeness between the Hessian inverse approximation matrix $\hat{\hbH}_{(U)}^{-1}$ and the exact Hessian inverse $\bbH^{-1}$. Hence, as $U$  increases, the Hessian inverse approximation $\hat{\hbH}_{(U)}^{-1}$ approaches the exact Hessian inverse $\bbH^{-1}$.
\subsection{The Structure of the DAGM Algorithm}
The iteration indices of the outer and inner loops  are denoted by $k\in \{0,\ldots,K-1 \}$ and $t\in \{0,\ldots,M \}$, respectively. For all $k\geq 0$ and $t\geq 0$,
\begin{equation*}
\textbf{q}_{k}^{t}:=\nabla_{\y} \mathbf{G}(\x_{k},\y_k^t)=\frac{1}{\beta}(\textbf{I}_{nd_2}-\W){\y}_k^t+ \nabla_{\y} \mathbf{g}(\x_{k},\y_k^t)
\end{equation*}
denotes the gradient of the inner function at the current point $(\x_{k},\y_k^t)$.  Given the above gradient, we use decentralized gradient descent (DGD) \cite{nedic2009distributed} to minimize (approximately) the inner function as follows:
\begin{equation}\label{qqq}
\y_{k}^{t+1}=\y_k^{t}-\beta \textbf{q}_k^{t},
\end{equation}
where $\textbf{q}_k^t=[\mathrm{q}_{1,k}^t;\ldots;\mathrm{q}_{n,k}^t]$ and
its $i$-th element is given by
\begin{subequations}
\begin{align}\label{ee}
\nonumber \mathrm{q}_{i,k}^{t}&=\frac{1}{\beta}(1-w_{ii}){ \yy_{i,k}^{t}}
\\&-\frac{1}{\beta}\sum_{j\in \mathcal{N}_i}w_{ij} \yy_{j,k}^{t}+ \nabla_{\yy} g_i(\xx_{i,k}, \yy_{i,k}^{t}).
\end{align}
Hence,
\begin{equation}\label{eqn:yi}
\yy_{i,k}^{t+1}= \yy_{i,k}^{t}-\beta \mathrm{q}_{i,k}^{t}.
\end{equation}
\end{subequations}

DAGM runs $M$ steps of Eq. \eqref{qqq} to  obtain an approximate solution $\y_k^M$ for the inner problem in Eq. \eqref{eqn:approximate:y}.
\begin{algorithm}[t]
\caption{ \pmb{DAGM}  }
\begin{algorithmic}[1]
\label{algo:D-BLGD}
\STATE{\textbf{Inputs:} $M,K,U\in \mathbb{N}$; $\{(\xx_{i,0},\yy_{i,0})\}_{i=1}^n \in \mathbb{R}^{d_1 } \times \mathbb{R}^{d_2}$; $( \beta, \alpha)\in \mathbb{R}_{++}$; and  a mixing matrix $\mr{W}$.}
\FOR{$k=0,1,2, \dots,K-1$}
\STATE Set $\tilde{\yy}_{i,k-1}=\yy_{i,k}^{0}=\yy_{i,k-1}^{M}$ if $k>0$
\\\quad and $\yy_{i,0}$ otherwise.
\FOR{$t=0, \dots, M-1$}
\STATE Exchange iterate $\yy_{i,k}^{t}$ with neighbors $j\in \mathcal{N}_i$.
\STATE $\mathrm{q}_{i,k}^{t}=\frac{1}{\beta}(1-w_{ii}){ \yy_{i,k}^{t}}-\frac{1}{\beta}\sum_{j\in \mathcal{N}_i}w_{ij} \yy_{j,k}^{t}+\nabla_{\yy} g_i(\xx_{i,k}, \yy_{i,k}^{t}).
$
\STATE
$\yy_{i,k}^{t+1}= \yy_{i,k}^{t}-\beta \mathrm{q}_{i,k}^{t}.$
\ENDFOR
\STATE  $ \tilde{\yy}_{i,k}=\yy_{i,k}^{M}$.
\STATE Exchange the iterate $\xx_{i,k}$ with neighbors $j\in \mathcal{N}_i$.
\STATE $\mr h_{i,k,(U)} =\pmb{\textnormal{DIHGP}}(\xx_{i,k}, \tilde{\yy}_{i,k}, U,\mr{W}).$
\STATE $\mathrm{d}_{i,k,(U)} =\frac{1}{\alpha}(1-w_{ii}){\xx_{i,k}}-\frac{1}{\alpha}\sum_{j\in \mathcal{N}_i}w_{ij}\xx_{j,k}$\qquad $~~~~+ \nabla_{\xx} f_i(\xx_{i,k},\tilde \yy_{i,k})+\beta\nabla^2_{\xx \yy}g_i(\xx_{i,k},\tilde \yy_{i,k})\mr h_{i,k,(U)}$.
\STATE Update local iterate:
$\xx_{i,k+1}=\xx_{i,k}-   \alpha\mathrm{d}_{i,k,(U)}$.
\ENDFOR
\end{algorithmic}
\end{algorithm}

Next, we provide the outer-level update using the estimator  $\y_k^M$ from the inner loop.  Each outer-level iteration requires hyper-gradient estimation.  As described in the previous section, the DAGM algorithm truncates the first $U$ summands of the Hessian inverse Taylor series in Eq. \eqref{olhij} to approximate the inverse of the Hessian matrix.  Then, we set
\begin{subequations}
\begin{equation}\label{eqn:update:DAGM}
\x_{k+1}=\x_{k}-\alpha   \widehat{\nabla} \mathbf{F}(\x_k,\y_k^M),
\end{equation}
 where $\widehat{\nabla} \mathbf{F}(\x_k,\y_k^M)$ is an estimate of the hyper-gradient by using \eqref{mkk} and setting  $\y=\y_k^M$ in \eqref{eqn:grad:tildF} as
\begin{align}
 \nonumber \widehat{\nabla} \mathbf{F}(\x_k,\y_k^M)&:=\frac{1}{\alpha}(\textbf{I}_{n d_1}-\acute{\W}) \x_k+\nabla_{\x} \mathbf{f}(\x_k,\y_k^M)
  \\&+\beta
  \nabla^2_{\x \y} \mathbf{g}\left(\x_k,\y_k^M\right) \textbf{h}_{k,(U)},\quad \textnormal{and}\label{eqn:d-k-U}\\\nonumber
  \textbf{h}_{k,(U)}&:=-\textbf{D}_k^{-1/2}\sum_{u=0}^{U}(\textbf{D}_k^{-1/2} \textbf{B} \textbf{D}_k^{-1/2})^u\textbf{D}_k^{-1/2} \p_k.
\end{align}
\end{subequations}
Note that DAGM relies on the DIHGP algorithm, which estimates the hyper-gradient by truncating Taylor's expansion for the DIHGP step. By leveraging DIHGP, we can implement Algorithm~\ref{algo:D-BLGD} using matrix-vector products and communication of vectors.
Additionally, each node $i$ needs to broadcast $U+1$ vectors of dimension $d_1$ per iteration, namely $\xx_{i,k}$, $\{\mathrm{d}_{i,k,(s)}\}_{s=0}^{U-1}\in \mathbb{R}^{d_1}$, to its
neighbors. Hence, at each iteration of DAGM, node $i$ sends $(U+1)|\mathcal{N}_i|$ vectors of the same dimension $d_1$ to the neighboring nodes. Thus, the per-iteration communication complexity of the DAGM algorithm increases linearly with the approximation order $U$. Consequently, there is a complexity-accuracy trade-off for the choice of $U$. Specifically, increasing $U$ will boost the approximation accuracy (and thus per-iteration performance) of the DAGM
and incur a higher communication burden.
\section{Convergence Analysis} \label{sec:convergence}
Next, we provide theoretical results for DAGM in the strongly convex, convex, and non-convex settings, respectively. All
proofs are provide in Appendix \ref{apen}.

Throughout this section, we also assume $\xx_{i,0}=0$ for simplicity in the analysis. To present the convergence results, it
is necessary to define the following average sequence:
\begin{align}\label{eqn:barx}
 \bar{\x}_k:=1_n \otimes \bar{\mr \xx}_k\in \mathbb{R}^{nd_1\times 1},  \quad \bar{\xx}_k :=\frac{1}{n}\sum_{i=1}^n\xx_{i,k},
\end{align}
Moreover, for notational convenience, let us define $\mathbf{f}^*:=\mathbf{f}(\x^*,\y^*(\x^*))$, and
\begin{equation}
\begin{aligned}\label{dfc}
 D_{F}&:=\F(\x_0,\check{\y}^*(\x_0))-\F(\check{\x}^*,\check{\y}^*(\check{\x}^*)),\\
 P_0&:=\|\y_{0}-\check{\y}^*(\x_{0}) \|^2.
\end{aligned}
\end{equation}
When the objective is (strongly) convex, we can obtain an upper bound on the suboptimality of our algorithm's output by computing the function difference at the current iteration and at a minimizer. On the other hand, when the objective is nonconvex, approximating local/global minima generally becomes intractable \cite{nemirovskij1983problem}. Consequently, as is common in the literature, we demonstrate that the algorithm can find an approximate first-order stationary point of the objective. Using $\m{f}^*=\mathbf{f}(\x^*,\y^*(\x^*))$, the rates stated in Table \ref{tab:1} provide upper bounds for the following expressions: $(1/n)1^{\top}\left(\mathbf{f}(\bar\x_{K},\y^*(\bar\x_{K})) - \mathbf{f}^*\right)$ in the strongly convex case, $(1/n)1^{\top}\left(\mathbf{f}(\widehat{\x}_{K},\y^*(\widehat{\x}_{K})) - \mathbf{f}^*\right)$ in the convex case, and $(1/K)\sum_{k=0}^{K-1}\|(1/n)1^{\top}\nabla\mathbf{f}(\bar\x_k,\y^*(\bar\x_k))\|^2$ in the nonconvex case. Note that we track the objective function computed at the average of the iterations over all nodes $\widehat{\x}_{K} := (1/K)\sum_{k=1}^K \bar\x_k$, where $\bar\x_k := (1/n)\sum_{i=1}^n \x_{i,k}$. This can be further simplified in the strongly convex setting to the objective computed at $\bar\x_k$.
The following Theorems \ref{thm:Strongly-Convex}, \ref{thm:convex}, and \ref{thm:Non-Convex} detail these results. Throughout these theorems, we set
\begin{align}\label{eqn:beta}
\beta &\leq \bar{\beta}:=\min\left\{\frac{b_g}{\lambda_{\max}(\textbf{I}_{nd_2}-\W) L_{g}},\frac{2}{\mu_{g}+L_{g}},\frac{1}{b_g},1\right\}, 
\end{align}
where $b_g=\hat{\lambda}_{\min}(\textbf{I}_{nd_2}-\W)+\frac{\mu_{g}L_{g}}{\mu_{g}+L_{g}}$ and $\hat{\lambda}_{\min}$ represents the nonzero minimum eigenvalue.
\begin{thm}[\textbf{Strongly-Convex}]\label{thm:Strongly-Convex}
Suppose Assumption~\ref{assu:netw} on the weight matrix $\mr W$ and Assumption~\ref{assu:lip} on the local functions hold.  Further, assume  $\{f_i\}_{i=1}^n$ are strongly convex with parameter $\mu_{f} > 0$. If $\beta$ satisfies \eqref{eqn:beta},
\begin{align*}
\alpha  &\leq \frac{1}{2L_F},~M \geq \max\left\{k+1,\frac{\log (\alpha)}{ \beta}\right\} ,
\\U &= \left|\left\lceil\frac{1}{2}\log_{1/\rho} \left(\eta^2K/(1-\beta b_g)^{k+1}\right) \right\rceil\right|,
\end{align*}
where $\rho$, $L_F$ and $\eta$ are defined in Lemmas \ref{symmetric_term_bounds11}, \ref{lem:lip} and \ref{lem:error}, respectively,
then for any $K\geq 1$:
\begin{align*}
&\quad \frac{1}{n}1^{\top}\left(\mathbf{f}(\bar\x_{K},\y^*(\bar\x_{K}))-\mathbf{f}^*\right) 
\\&\leq \frac{1}{n}\left( D_{F}  +\frac{1}{ L_F} \left(\frac{1}{2}
+C^2  P_0\right)\right) (1-\nu)^{K}\\
& ~ +\frac{ \hat{C}\tilde{C}  }{n(1-\sigma)} \alpha+\frac{ 2\hat{L}_gC_{\F}}{n(1-\sigma)^{\frac{1}{2}}}\left(\frac{\beta^{\frac{1}{2}}}{(1-\sigma)^{\frac{1}{2}}}+\mu_g^\frac{-1}{2}\right) \beta^{\frac{1}{2}},
\end{align*}
 where  $\bar{\x}_k$ is defined in \eqref{eqn:barx}, and $D_{F}$ and $P_0$ are defined in \eqref{dfc},
\begin{equation}\label{gk}
\nu:=\min \left\{ \alpha \mu_F , \beta b_g\right\},~\textnormal{and}~
 C_{\F}:=C_f+(2\alpha)^{-1}(1-\sigma),
 \end{equation}
 with $\mu_{F}$ given by $\mu_{F}:=\mu_f+(2\alpha)^{-1}(1-\sigma)$;   $L_F$, $\hat{L}_g$ and  $\hat{C}$, $\tilde{C}$ are defined in \eqref{lf}, Lemma \ref{cl} and \eqref{eqn:po}, respectively.
\end{thm}
A key part of our theoretical result is that although the algorithm is applied to the penalized problem, the convergence bound is obtained for the original problem. Specifically, to obtain the bound on $({1}/{n})1^{\top} \left(\mathbf{f}(\bar{\mathbf{x}}_K, \mathbf{y}^*(\bar{\mathbf{x}}_K)) - \mathbf{f}^*\right)$ in Theorem \ref{thm:Strongly-Convex}, we need the following inequality using \eqref{eqn:reform:dblo}:
\begin{align*}
     \frac{1}{n}&1^{\top}\left(\mathbf{f}(\bar{\x}_{K},\y^*(\bar{\x}_{K}))-\mathbf{f}^*\right)
      \leq \|{\y}^*(\bar{\x}_{K})- \check{\y}^*(\bar{\x}_{K})\|
      \\&+\F(\bar{\x}_{K},\check{\y}^*(\bar{\x}_{K}))-\mathbf{F}(\check{\x}^*,\check{\y}^*(\check{\x}^*))
       \\&+\frac{1}{2\alpha} {{\x}^*}^\top (\textbf{I}_{nd_1}-\acute{\W}){{\x}^*}- \frac{1}{2\alpha} {\bar{\x}_{K}}^\top (\textbf{I}_{nd_1}-\acute{\W}){\bar{\x}_{K}}.
\end{align*}
The first term on the R.H.S., which captures the gap between $\check{\y}^*(\x)$ and $\y^*(\x)$ --- the solutions of the penalized inner Problem \eqref{eqn:approximate:y} and the constrained inner bilevel Problem \eqref{eqn:obj:cbo:inn}, respectively --- is bounded by Lemma \ref{cl}. The second term represents the gap between the objective at iterates generated by an optimization algorithm applied to the penalized Problem \eqref{eqn:approximate:prob}, i.e., $\F(\bar{\x}_{K}, \check{\y}^*(\bar{\x}_{K}))$, and its optimal objective $\mathbf{F}(\check{\x}^*, \check{\y}^*(\check{\x}^*))$, which is bounded by Lemma \ref{f2}. The third term, bounded by Lemma \ref{lem:up}, represents the difference between the penalty terms computed at ${\x}^*$ and $\bar{\x}_{K}$, the optimum and the output of Algorithm~\ref{algo:D-BLGD}, respectively.
\begin{cor}\label{cor:Strongly-Convex}
 If $\alpha = \mathcal{O}(\delta^K)$ for some $\delta<1$ and  $\beta = \mathcal{O}(n^{-1}\delta^{2(b+K)})$ with
$b\geq \log(n\bar{\beta})^{1/2}$ where $\bar{\beta}$ is defined in \eqref{eqn:beta}, in Theorem \ref{thm:Strongly-Convex}, then 
$({1}/{n})1^{\top}\left(\mathbf{f}(\bar\x_{K},\y^*(\bar\x_{K}))-\mathbf{f}^*\right) \leq \epsilon$ needs  $\mathcal{O}\big(\log \frac{n^{-1}\epsilon^{-1}}{1-\sigma}\big)$ iterations of Algorithm~\ref{algo:D-BLGD}.
\end{cor}

}
\begin{thm}[\textbf{Convex}]\label{thm:convex}
Suppose Assumption~\ref{assu:netw} on the weight matrix $\mr W$ and Assumption~\ref{assu:lip} on the local functions hold. Further, assume  $\{f_i\}_{i=1}^n$ are convex.  If $\beta$ satisfies \eqref{eqn:beta},
\begin{align*}
\alpha\leq \frac{1}{L_F},\quad U=\left|\lceil\log_{1/\rho} \eta K \rceil\right|, \quad
M=\mathcal{O}\left(\frac{K\alpha}{\beta}\right),
\end{align*}
where $\rho$, $L_F$ and $\eta$ are defined in Lemmas \ref{symmetric_term_bounds11}, \ref{lem:lip} and \ref{lem:error}, respectively,
then for any $K\geq 1$ with $\|\check{\x}^* \|\leq R$, we have
\begin{align*}
\nonumber&\quad \frac{1}{n}1^{\top}\left(\mathbf{f}(\widehat{\x}_{K},\y^*(\widehat{\x}_{K}))-\mathbf{f}^*\right)\\&\leq
\frac{1}{n K}\left( \frac{1}{2\alpha}R^2+\tau (\tau+R) \right)+\frac{  \tilde{C} \hat{C}  \alpha }{n(1-\sigma)}
\\& +\frac{2\hat{L}_gC_{\F}}{n(1-\sigma)^{\frac{1}{2}}}\Big(\frac{\beta^{\frac{1}{2}}}{(1-\sigma)^{\frac{1}{2}}}+\mu_g^{\frac{-1}{2}}\Big) \beta^{\frac{1}{2}},
\end{align*}
 where  $\widehat{\x}_{K}=({1}/{K})\sum_{k=1}^{K}\bar{\x}_{k}$, and $\bar{\x}_k$, $ C_{\F}$, $\tau$, and $D_{F}$, $P_0$ are, respectively, defined in \eqref{eqn:barx}, \eqref{gk}, Lemma \ref{lem:dif:convex}, and \eqref{dfc}. Moreover, $L_F$, $\hat{L}_g$, and $\hat{C}$, $\tilde{C}$ are defined in \eqref{lf}, Lemma \ref{cl}, and \eqref{eqn:po}, respectively.
\end{thm}
\begin{cor}\label{cor:convex}
If $\alpha=\mathcal{O}(\sqrt{1-\sigma}/\sqrt{K})$, and $\beta= \mathcal{O}(1-\sigma/(b+K))$ for some $b \geq (1-\sigma)/\bar{\beta}$ where $\bar{\beta}$ is defined in \eqref{eqn:beta}, in Theorem \ref{thm:convex}, then 
$({1}/{n})1^{\top}\left(\mathbf{f}(\widehat{\x}_{K},\y^*(\widehat{\x}_{K}))-\mathbf{f}^*\right) \leq \epsilon$ needs  $\mathcal{O}\big(\log \frac{n^{-2}\epsilon^{-2}}{1-\sigma}\big)$ iterations of Algorithm~\ref{algo:D-BLGD}.
\end{cor}
\begin{thm}[\textbf{Non-convex}]\label{thm:Non-Convex}
Suppose Assumption~\ref{assu:netw} on the weight matrix $\mr W$ and Assumption~\ref{assu:lip} on  $\{f_i\}_{i=1}^n$ hold. If $\beta$ satisfies \eqref{eqn:beta},
\begin{align*}
\alpha\leq \frac{1}{8L_F},\quad U=\left|\left\lceil\frac{1}{2}\log_{1/\rho} (\eta^2K) \right\rceil\right|, \quad
M=\mathcal{O}\left(\frac{1+\alpha^2}{\beta}\right),
\end{align*}
where $\rho$, $L_F$ and $\eta$ are defined in Lemmas \ref{symmetric_term_bounds11}, \ref{lem:lip} and \ref{lem:error}, respectively,
then for any $K\geq 1$, we have
\begin{align*}
&\quad\frac{1}{K}\sum_{k=0}^{K-1}\|\frac{1}{n}1^{\top}\nabla\mathbf{f}(\bar\x_k,\y^*(\bar\x_k))\|^2
\\
&\leq \frac{\jmath_116}{nK\alpha} \left(D_{F}
+\jmath_2\right)+\frac{4L_{f}^2\alpha^2 \hat{C}^2}{n(1-\sigma)^2}
\\&+\frac{\jmath_1 2C^2}{n}\left(\frac{4\hat{L}_g\beta}{(1-\sigma)}+\frac{4\hat{L}_g\beta^{1/2}}{(1-\sigma)^{1/2}\mu_g^{1/2}}\right),
\end{align*}
with
\begin{align}\label{eqn:conca:j}
\nonumber   \jmath_1&:=2+\frac{16 (L_{f_{\xx}}+L_{f_{\yy}} )}{(1-\sigma)^2},~
    \jmath_2:=5\big((2+\frac{3\varrho^2}{16 L_F^2} )
+ P_0\big),
\end{align}
where $\hat{L}_g$, $\varrho$, $L_F$, $\hat{C}$, and  $D_{F}$, $P_0$ are, respectively, defined in Lemma \ref{cl}, Lemma  \ref{lem:lip}, \eqref{lf}, \eqref{eqn:po}, and \eqref{dfc}.
\end{thm}


\begin{cor}\label{cor:nonconvex}
If $\alpha = \mathcal{O}(1/\sqrt{K})$ and  $\beta = \mathcal{O}((1-\sigma)/(b+K))$ for some $b \geq (1-\sigma)/\bar{\beta}$ where $\bar{\beta}$ is defined in \eqref{eqn:beta}, in Theorem \ref{thm:Non-Convex}, then $({1}/{K})\sum_{k=0}^{K-1}\|({1}/{n})1^{\top}\nabla\mathbf{f}(\bar\x_k,\y^*(\bar\x_k))\|^2\leq \epsilon$ needs $\mathcal{O}\left(n^{-2}\epsilon^{-2}+(n^{-1}\epsilon^{-1}/(1-\sigma)^2)\right)$ iterations of Algorithm~\ref{algo:D-BLGD}.
\end{cor}

\subsection{Comparison of Communication Complexity with Other Decentralized Bilevel Methods}
\label{sec:commun}
Since existing decentralized methods such as DGBO \cite{yang2022decentralized} and DGTBO \cite{chen2022decentralized} address nonconvex problems, for a fair comparison, we compare the communication complexity of DAGM with these methods in a nonconvex setting. In this context, the communication complexity is defined as the total number of communication rounds required by an algorithm to achieve an $\epsilon$-stationary point, meaning a point such that $\|\nabla f(\mathbf{x})\| \leq \epsilon$. Our algorithm, DAGM, significantly reduces the number of communication rounds and minimizes communication costs per round by approximating the hyper-gradient through decentralized computation of matrix-vector products and minimal vector communications. Table~\ref{tab:complexity_comparison} provides details of our communication complexity and a comparison with existing literature.
\begin{table}[th]
    \centering
    \begin{tabular}{|>{\centering\arraybackslash}m{1.3cm}|>{\centering\arraybackslash}m{6.3cm}|}
        \hline
        \textbf{Algorithm} & \textbf{Communication Complexity} \\ \hline
        DAGM & $\mathcal{O}\left( \frac{1}{n\epsilon(1-\sigma)^2}\left((d_1 +d_2)\log (\frac{1}{\epsilon})+ d_1\right)\right)$ \\ \hline
        DGBO \cite{yang2022decentralized} & $\mathcal{O}\left(\frac{1}{\epsilon(1-\sigma)^2}\left(d_2^2\log (\frac{1}{\epsilon}) + d_1d_2\right)\right)$ \\ \hline
        DGTBO \cite{chen2022decentralized} & $\mathcal{O}\left(\frac{1}{\epsilon(1-\sigma)^2}\left(d_1d_2\log (\frac{1}{\epsilon})+d_1\right)\right)$ \\ \hline
    \end{tabular}
    \caption{ \textbf{Comparison of communication complexity} to achieve an $\epsilon$-stationary point; here $n$ is the number of nodes, $1-\sigma$ is the spectral gap of the network, and $d_1$ and $d_2$ are the dimensions of outer and inner optimization variables.}
          \label{tab:complexity_comparison}
\end{table}

It is worth noting that if $W$ is a matrix of all ones, then the communication complexity of our algorithm matches that of centralized bilevel algorithms, as both this algorithm and centralized methods require $M$ inner iterations. However, $W$ can be very sparse, as considered in our experiments, and only neighborhood communication is needed. Hence, for some standard choices of $W$, such as the Metropolis weight matrix considered in this paper, the communication complexity of our algorithm is much smaller than that of existing decentralized algorithms. We observe that DGBO \cite{yang2022decentralized} shares similarities with our algorithm, including a subalgorithm with $b$ iterations for global Hessian computation, and $K$ outer loops (main iterations). In Table~\ref{tab:complexity_comparison} (details in Appendix \ref{ap:complexity}), we also provide the total communication complexity of their algorithm. Similarly, the decentralized bilevel optimization algorithm (DGTBO) in \cite{chen2022decentralized} tackles Problem \eqref{eqn:main:dblo}, for which we have provided the total communication complexity in Table~\ref{tab:complexity_comparison}. DAGM is theoretically superior to DGBO and DGTBO in terms of communication complexity, as it includes a $1/n$ term in its bound, indicating linear acceleration, and its communication dependency on dimensions is $d_1+d_2$. In contrast, for DGBO and DGTBO, the dimension dependencies are $d_2^2+ d_1d_2$ and $d_1d_2$, respectively.
\begin{figure*}[t]
\centering
\includegraphics[scale=0.12]{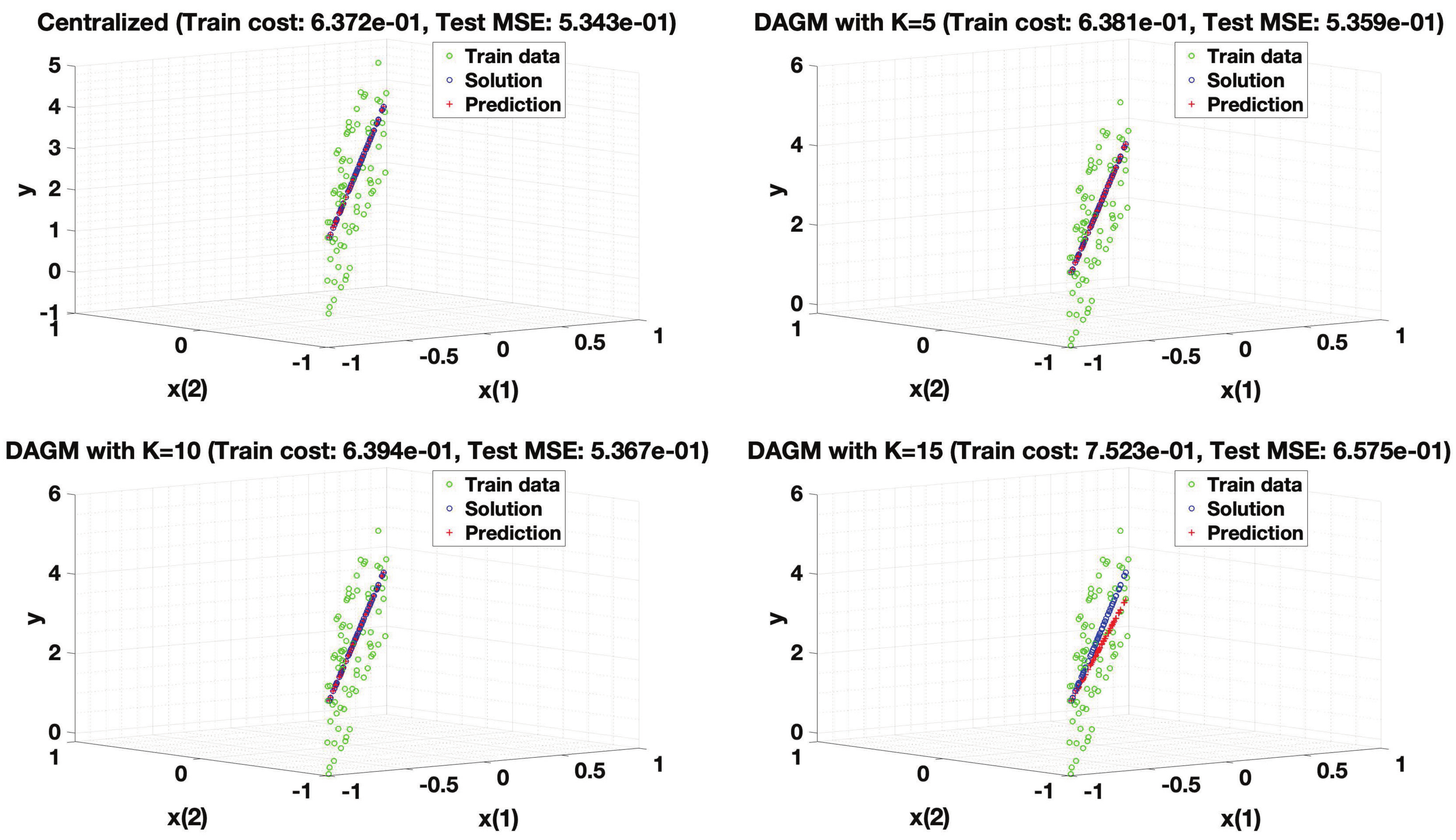}
\caption{Training cost and test MSE of DAGM for solving a regularized linear regression problem over 100 epochs for a synthetic dataset.}\label{fig:linear:regression}
\end{figure*}
\begin{figure*}[t]
\centering
    \begin{subfigure}{0.45\textwidth}
      \includegraphics[scale=0.055]{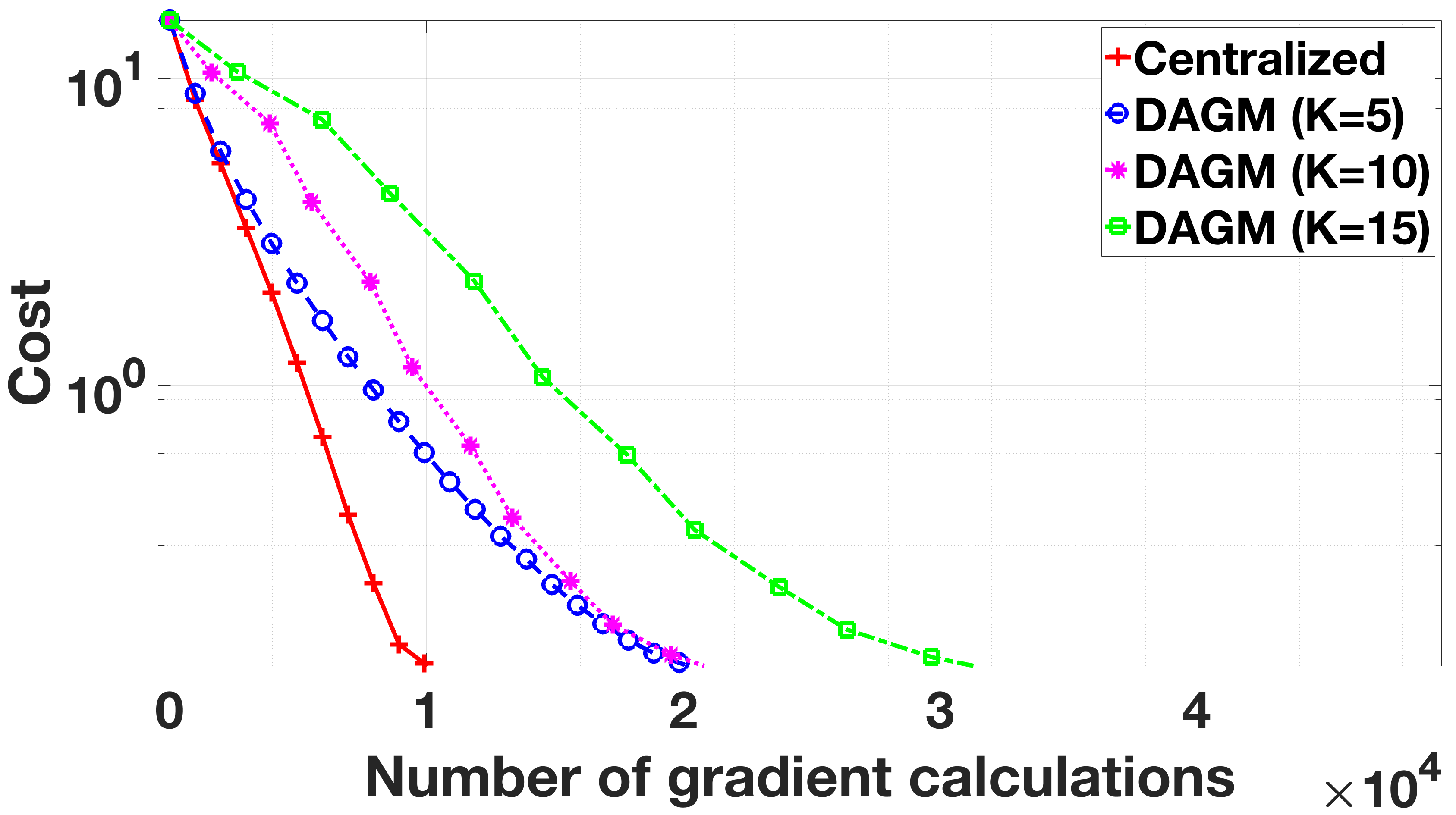}
       \caption{Regularized softmax problem applied to MNIST dataset.}
       \label{fig:minist}
   \end{subfigure}
   \hspace{1cm}
   \begin{subfigure}{0.45\textwidth}
\includegraphics[scale=0.055]{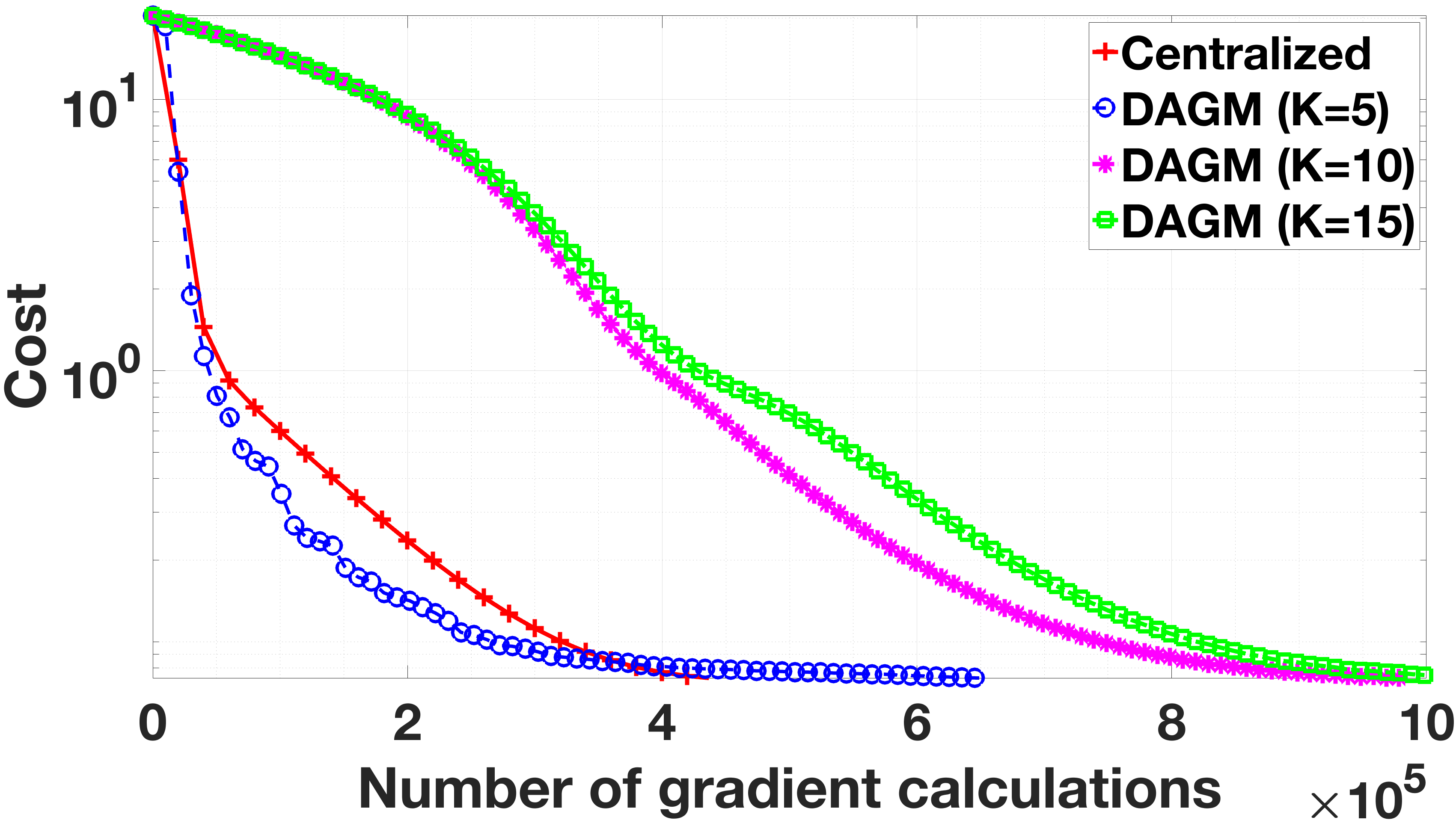}
       \caption{Regularized support vector machine (SVM) problem applied to {Mushroom} dataset.}
       \label{fig:mushroom}
   \end{subfigure}
   \caption{Convergence of DAGM algorithm over 100 epochs on the real datasets.
   }\label{fig:minist:mushroom}
\end{figure*}

\section{Numerical Experiments on Bilevel Type Problems}\label{sec:numerical results}
\subsection{Decentralized Hyper-parameter optimization (HO)} HO is the process of finding the best set of hyper-parameter values that cannot be learned using the training data alone \cite{franceschi2018bilevel}. An HO problem can be formulated as a bilevel optimization problem as follows: the outer objective $f(\yy^*(\xx);\mc{D}^{\text{val}})$ aims to minimize the validation loss with respect to the hyper-parameters $\xx$, and the inner objective function $g(\xx,\yy;\mc{D}^{\text{tr}})$ optimizes a learning algorithm by minimizing the training loss with respect to the model parameters $(\xx, \yy)$, where $\mathcal D^{\text{tr}}$ and $\mathcal D^{\text{val}}$ denote the training and validation datasets, respectively. Given a loss function $\ell$, a set of $n$ agents, $\mc{D}^{\text{val}}=\{\mc{D}^{\text{val}}_1, \ldots, \mc{D}^{\text{val}}_n\}$, and $\mc{D}^{\text{tr}}=\{\mc{D}^{\text{tr}}_1, \ldots, \mc{D}^{\text{tr}}_n\}$, the decentralized HO problem can be formulated as  \eqref{eqn:main:dblo} where $ \sum_{i=1}^n f_i(\yy^*(\xx);\mc{D}^{\text{val}})=\sum_{i=1}^n \ell(\yy^*(\xx);\mc{D}^{\text{val}}_i)$ and $ \yy^*(\xx)\in\argmin_{\yy\in\mathbb{R}^{d_2}}~~ \sum_{i=1}^n g_i(\xx,\yy;\mc{D}^{\text{tr}}):=\ell(\yy;\mc{D}^{\text{tr}}_i)+\yy^\top \exp(\xx)$.
Note that $\exp(\cdot)$ represents the element-wise exponential function. Furthermore, utilizing the $\exp(\cdot)$ function ensures non-negative outputs, with $ \exp (\xx) \geq 0$, which is  desirable for the inner optimizer's outputs. In our experiments, for any $(\zz_i,b_i) \in \mc{D}^{\text{tr}}_i$, we consider the following choices of loss functions:
\begin{itemize}
    \item \textbf{Linear}: $ \ell(\yy;\mc{D}^{\text{tr}}_i)=(\yy^{\top}\zz_i-b_i)^2$;
    \item \textbf{Logistic}: $\ell(\yy;\mc{D}^{\text{tr}}_i)=\log\left(1 + \exp(-b_i\yy^{\top}\zz_i)\right)$;
    \item \textbf{Support Vector Machine (SVM)}: $$\ell(\yy;\mc{D}^{\text{tr}}_i)=\max\left\{0,1-b_i\yy^{\top}\zz_i\right\};$$
    \item \textbf{Softmax}: $ \ell(\yy,u;\mc{D}^{\text{tr}}_i)=-\text{log}\frac{e^{\yy_{b_i}^{\top}\zz_i + u_{b_i}}}{\sum_{j=1}^Ce^{\yy_j^{\top}\zz_i + u_j}};$
\end{itemize}
where $C$ represents the number of classes, and $b_i\in [C]$.
The inner and outer step-sizes $(\alpha,\beta)$ are chosen from the set: $\{1e-2, 5e-3, 1e-3, 5e-4, 1e-4\}$. The number of DIHGP updates, $U$, is set to $3$. The mixing matrix $\mr W=[w_{ij}]$ is defined as
\begin{eqnarray}\label{eq:mahan}
w_{ij}=\left\{
      \begin{array}{ll}
      \frac{1}{1+\max\{\degree(i),\degree(j)\}},&\text{if } \{i,j\}\in\mathcal{E}, \\
      1-\sum\limits_{\{i,k\}\in\mathcal{E}} w_{ik},&\text{if } i=j,\\
      0, &\text{otherwise},
      \end{array}\right.
\end{eqnarray}
where $\text{deg}(i)$ represents the degree of agent $i$, that is, the number of neighbors the agent has. This corresponds to the Metropolis weight matrix in which the weight on each edge is one over one plus the larger degree at its two incident vertices. The self-weights are chosen so that the sum of weights at each node is $1$ \cite{boyd2004fastest}. The connected network is generated randomly with a connectivity ratio of $r=0.5$.

\begin{figure*}[t]
\centering
      \includegraphics[scale=0.35]{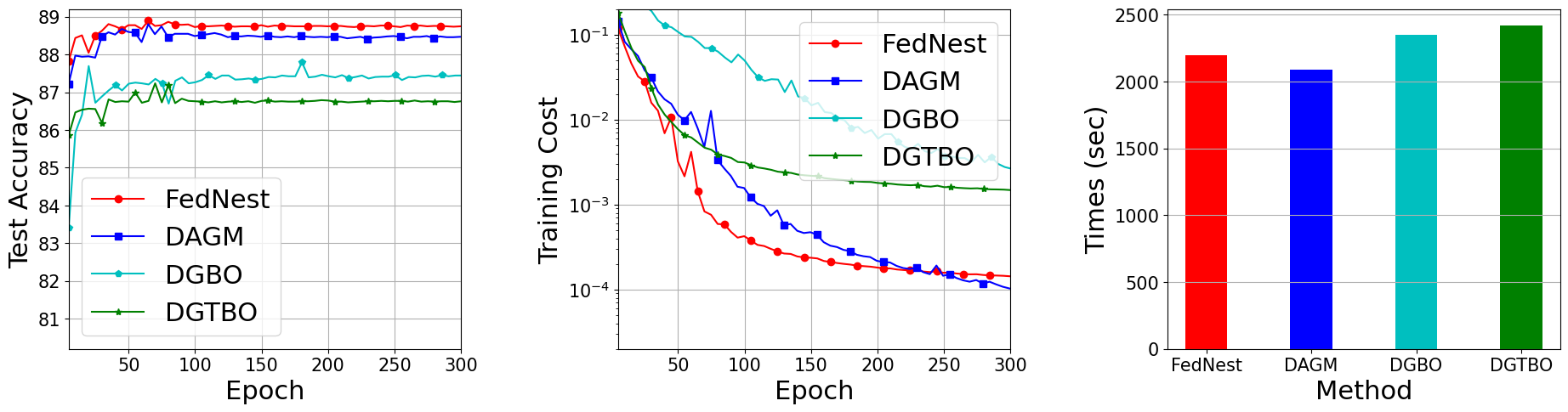}
   \caption{Distributed Hyper-representation experiments on a 2-layer MLP and MNIST dataset.
   }\label{fig:hyper2}
\end{figure*}
\textbf{Synthetic Data}.
First, we illustrate the performance of Algorithm~\ref{algo:D-BLGD} on a synthetic dataset using regularized linear regression. We set $d_1=d_2=2$ and $n=100$. The underlying true signal $\yy^*$ is generated from a standard normal distribution. Each data sample $\zz_i$ is generated as  $\zz_i =  (\yy^*)^\top \zz_i+\sigma\zz_i^\top \yy^* + \epsilon_i $, where $\sigma=0.25$, and each $\epsilon_i$ is drawn from a standard normal distribution. Figure~\ref{fig:linear:regression} provides the training cost and test mean square error (MSE) of DAGM while solving a regularized linear regression problem across 100 epochs and 10 replicates. 
From Figure~\ref{fig:linear:regression}, we can see that DAGM with $K=1, 5$, and $10$ gives accurate predictions, while there's a gradual increase in both the training cost and MSE from $K=1$ to $K=15$.

\begin{figure*}[t]
  \centering
  \includegraphics[width=.34\textwidth]{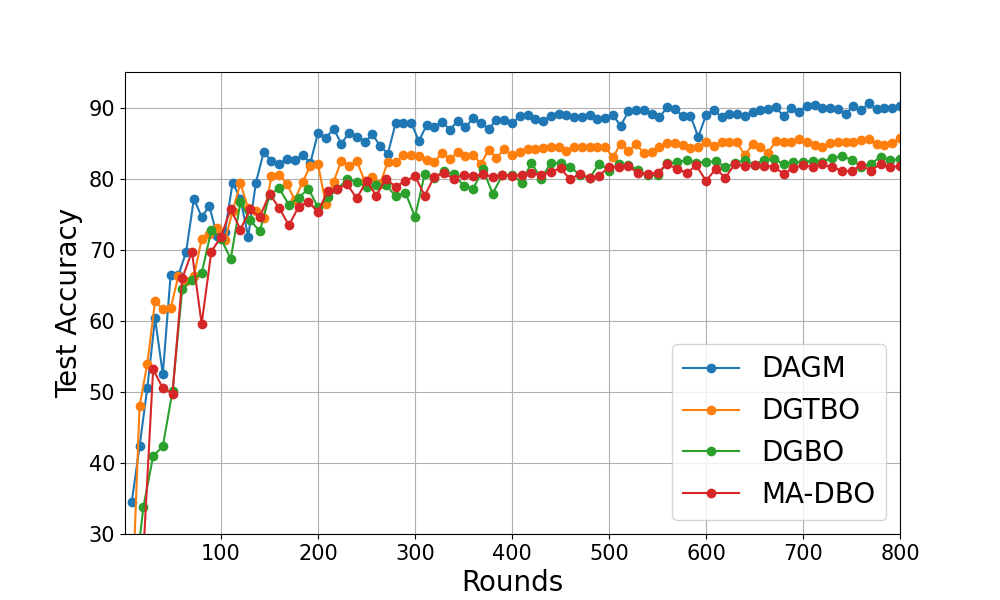} 
  \includegraphics[width=.34\textwidth]{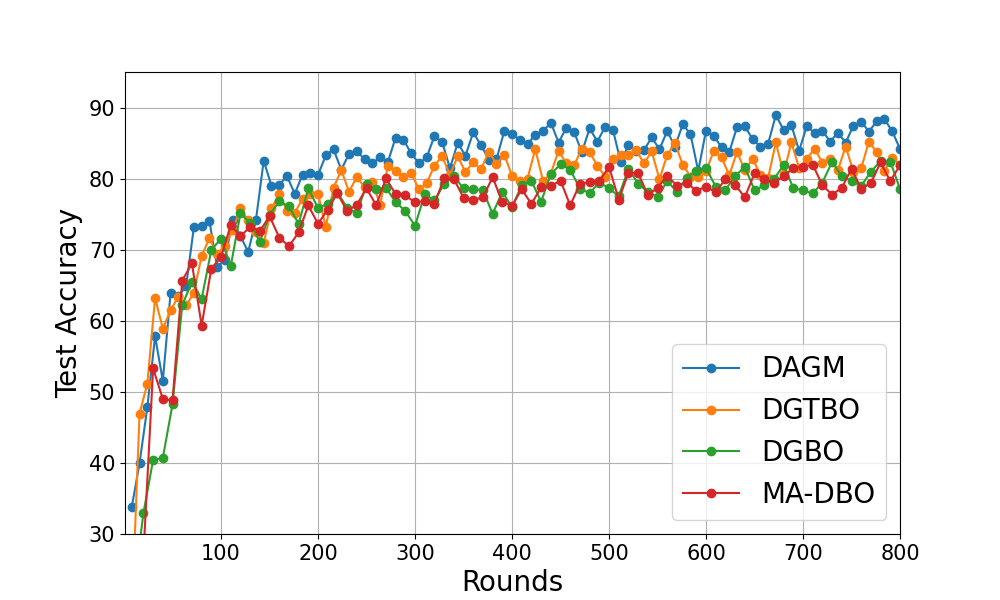} 
  \includegraphics[width=.31\textwidth]{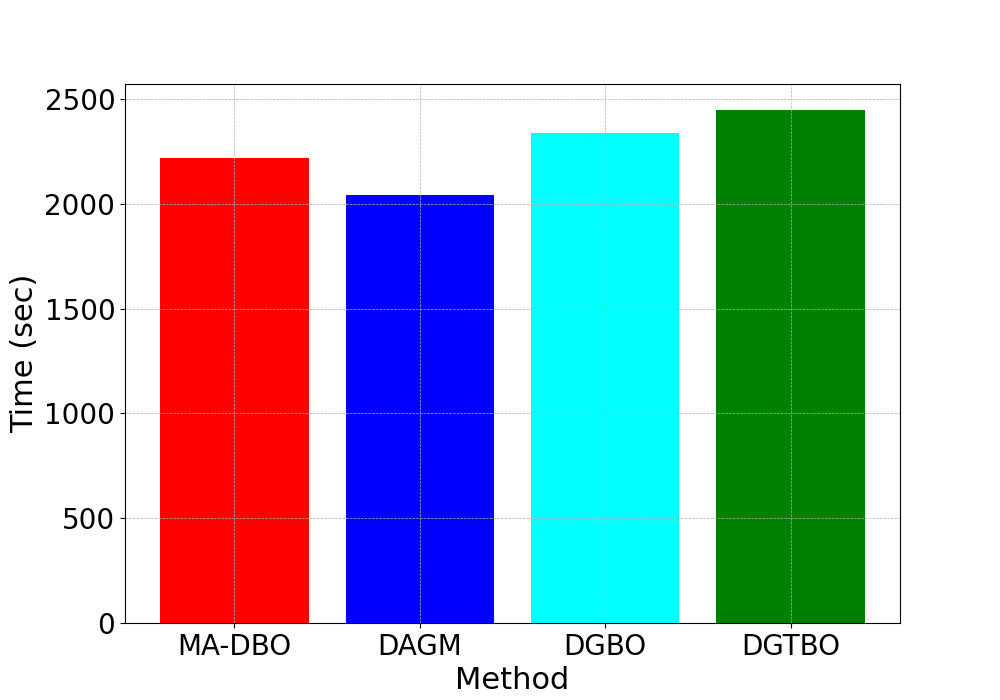} 
\caption{Test accuracy of decentralized algorithms across communication rounds on the long-tail MNIST dataset for heterogeneity levels \(q=0.1\) (left) and \(q=0.5\) (middle). The time comparison for \(q=0.5\) is shown in the right figure.}
\label{fig:heter}
\end{figure*}

\textbf{Real Data Applications}.~The numerical results on real datasets are shown in Figure~\ref{fig:minist:mushroom} for the DAGM algorithm. Figures~\ref{fig:minist:mushroom}(a)  and \ref{fig:minist:mushroom}(b) show the results for a regularized softmax regression applied to the MNIST dataset \cite{lecun2010mnist} and a regularized support vector machine (SVM) applied to the Mushroom dataset \cite{asuncion2007uci}, respectively. We observe that centralized DAGM performs the best in terms of the number of gradient computations. There is a gradual decrease in performance from $K=1$ and $K=15$. This indicates the trade-off between communication and convergence rate in decentralized bilevel optimization, as shown in our theoretical analysis. We note that centralized DAGM also does vector communications, but suffers from a high communication cost on the central node.

\subsection{Decentralized Representation Learning}\label{sec:exp:rl}

Next, we consider decentralized representation learning. Modern approaches in meta-learning such as model-agnostic meta-learning~\cite{finn2017model} and reptile \cite{nichol2018reptile} learn representations that are shared across all tasks in a bilevel manner. The hyper-representation problem optimizes a classification model in a two-phased process. The outer objective optimizes the model backbone to obtain better feature representation on validation data ($\{\mc{D}^t_i\}_{i=1}^n$), while the inner problem optimizes a header for downstream classification tasks on training data ($\{\mc{D}^t_{i}\}_{i=1}^n$). In this experiment, we use a 2-layer multilayer perceptron (MLP) with 200 hidden units. The outer problem optimizes the hidden layer with 157,000 parameters, and the inner problem optimizes the output layer with 2,010 parameters. We study a  non-i.i.d. partitioning of the MNIST data set following~ federated nested optimization (FedNest)~\cite{tarzanagh2022fednest}, and split each client's data evenly to train and validation datasets. Thus, each client has 300 train and 300 validation samples. We compare the proposed DAGM algorithm to decentralized Gossip-type bilevel optimization (DGBO)  \cite{yang2022decentralized}, decentralized gradient-tracking bilevel optimization (DGTBO) \cite{chen2022decentralized}, and FedNest~\cite{tarzanagh2022fednest} with only one local training. Note that the main difference between these algorithms is on the hyper-gradient updates.

Figure~\ref{fig:hyper2} demonstrates the impact of  different algorithms on test accuracy, training loss, and CPU time. It is evident that both the centralized FedNest and the decentralized DAGM algorithms perform well in terms of test accuracy and training loss. Additionally, both algorithms significantly outperform DGBO and DGTBO in CPU times. Specifically, the DAGM algorithm exhibits the best timing performance among all four algorithms due to its decentralized matrix-vector product and vector communication. These findings align with the discussion in Section~\ref{sec:A Decentralized Alternating Method} and show that the proposed algorithm is robust and scalable.

\subsection{Heterogeneous Fair Loss Tuning}\label{sec:exp:fairloss}

Our experiment follows the setup for loss function tuning on an imbalanced dataset, as described in \cite{tarzanagh2022fednest}. The objective is to maximize class-balanced validation accuracy while training on the imbalanced dataset. We adopt the same network architecture, long-tail MNIST dataset, and train-validation strategy as in \cite{tarzanagh2022fednest}. However, unlike their approach of partitioning the dataset into 10 clients using FedAvg \cite{mcmahan17fedavg} with either i.i.d. or non-i.i.d. distribution, we introduce fine-grained control over partition heterogeneity, inspired by \cite{murata2021bias}.

In a dataset with $n$ imbalanced classes, each containing $C_i$ samples, we aim to achieve a specified heterogeneity level $q \in [0,1]$. The dataset is divided into $n$ clients, each designed to have an equal number of samples, $n^{-1}\sum_{i=1}^{n} C_i$. For each client $i$, we include $q \times 100\%$ of the data from class $i$, or all $C_i$ examples if the class size is insufficient. If a client has fewer samples than $n^{-1}\sum_{i=1}^{n} C_i$, we fill the gap by uniformly sampling from the remaining samples across all classes.  This ensures an equal sample count across clients and a specified level of class heterogeneity, even within the context of an imbalanced dataset.  In our experiments, we split the imbalanced MNIST dataset into $n=10$ clients according to $q$.

 Figure \ref{fig:heter} provides the test accuracy of four different algorithms—DAGM, DGTBO \cite{chen2022decentralized}, DGBO \cite{yang2022decentralized}, and MA-DBO \cite{chen2023decentralized}—over a series of rounds and for two heterogeneity levels ($q=0.1$ and $q=0.5$).  In  Figure \ref{fig:heter} (left), $q=0.1$ indicating a lower level of heterogeneity in the distributed data, the test accuracy increases sharply during the initial rounds and reaches a plateau as the number of rounds progresses. DAGM exhibits a slightly higher test accuracy, maintaining a consistent lead above the other methods. The right graph is marked with $q=0.5$, which suggests a greater level of heterogeneity. Similar to the left graph, all algorithms show an initial rapid improvement in test accuracy, which stabilizes after approximately $400$ rounds. DAGM competes closely with DGTBO and DGBO in performance, with MA-DBO trailing slightly. DAGM's robustness to the varying levels of heterogeneity is evident as it maintains a high test accuracy across both scenarios. Furthermore, it significantly outperforms all three methods (DGTBO, DGBO and MA-DBO) in terms of runtime.

\section{Conclusion}\label{sec:conclusion}
This work introduces a novel decentralized algorithm -DAGM-for bilevel optimization problems. DAGM approximates the hyper-gradient by employing decentralized computation involving \textit{matrix-vector} products and \textit{vector communications}. Convergence rates are established for DAGM under diverse convexity assumptions for the objective function. Notably, our method achieves a linear acceleration (an $n^{-1}$ acceleration in complexity) even when incorporating vector computation/communication, distinguishing it from existing approaches. Numerical evaluations across various problems corroborate the theoretical findings and demonstrate the strong performance of the proposed method in real-world applications.
\section*{Acknowledgements}
The authors thank the Editor and reviewers for their constructive comments and suggestions.
The work of GM was partially supported by NSF grant DMS 2348640. The work of DAT was partially supported by NIH grants NIH U01 AG066833 and U01 AG068057.
\bibliographystyle{plain}
 \bibliography{autosam}

\newpage
\clearpage

\renewcommand{\thesection}{S\arabic{section}} 

\setcounter{section}{0}

\begin{center}
\Large\textbf{Supplementary Material for \\ \textquotedblleft A Penalty-Based Method for Communication-Efficient Decentralized Bilevel Programming \textquotedblright}
\end{center}

\section{Comparison in Communication Complexity}\label{ap:complexity}
In this section, we compare the communication complexities of DAGM in Algorithm \ref{algo:D-BLGD}, DGBO in \cite{yang2022decentralized}, and  DGTBO in \cite{chen2022decentralized}.
Compared to existing decentralized algorithms with the same standard choices of $W$, such as the Metropolis weight matrix considered  in our experiments, DAGM reduces the required
rounds of communications (i.e., communication complexity) and minimizes communication costs per
round by approximating the hyper-gradient through decentralized computation of matrix-vector products and minimal
vector communications.
\\
$\bullet$ \textbf{Communication complexity of DAGM (Algorithm 2 ) :}
   To illustrate this, we can divide the communication complexity of our algorithm into two parts: Algorithm 2 (DAGM) and Subalgorithm 1 (DIHGP)
   \begin{enumerate}
    \item[I).]\textit{Communication complexity of DIHGP}: As we can see, this algorithm operates in $U$ inner iterations, which are common in bilevel optimization. Each of these $U$ iterations requires the exchange of the vector $\mathrm{h}_{i,(s)} \in \mathbb{R}^{d_1}$ with neighbors $j \in \mathcal{N}_i$. Hence, the total inner communication of DIHGP is $U$ vectors of  dimension $d_1$.
    \item[II).]\textit{Communication complexity of DAGM}: We first consider the computational complexity of each iteration.
    \begin{itemize}
        \item    Step~5 exchanges the iterate $\mathrm{y}_{i,k}^{t} \in \mathbb{R}^{d_2}$ with neighbors $j \in \mathcal{N}_i$, which requires one communication of a $d_2$-dimensional vector and operates over $M$ iterations, resulting in $Md_2$ communication cost.
        \item    Step 10 exchanges the iterate $\mathrm{x}_{i,k}$ with neighbors $j \in \mathcal{N}_i$, which also requires one communication of a $d_2$-dimensional vector, resulting in $d_1$ communication cost.
        \item Finally, Step 11 calls the DIHGP sub-algorithm, which requires $U$ vectors of $d_1$ dimension, resulting in $Ud_1$ communication cost.
    \end{itemize}
\end{enumerate}
Overall, since the algorithm operates over $K$ iterations, the total number of iterations is $K((U+1)d_1 +M d_2)$.  Note that to achieve $\epsilon$ stationary we need $K=\mathcal{O}\left(\frac{1}{n\epsilon(1-\sigma)^2} \right)$ outer iterations, where $1-\sigma$ denotes the spectral gap of the
communication network, and the number of inner iterations $U=\mathcal{O}\left(\log (\frac{1}{\epsilon})\right)$. Hence, the communication complexity of DAGM is $\mathcal{O}\left( \frac{1}{n\epsilon(1-\sigma)^2}\left((d_1 +d_2)\log (\frac{1}{\epsilon}) + d_1\right)\right)$. As we can see, the communication complexity of our algorithm is linear in both inner and outer dimensions, $d_1$ and $d_2$, hence it only requires vector communication.

$\bullet$ \textbf{Communication complexity of DGBO \cite{yang2022decentralized} :}
We can divide the communication complexity of Gossip-Based Decentralized Stochastic Bilevel Optimization algorithm  (DGBO) in \cite{yang2022decentralized} into two parts:
Algorithm 1 and its inner loop (Lines 10-13).
\begin{enumerate}
    \item [I).] Communication complexity of inner loop (Lines 10-13): As we can see, these steps operate in $b$ inner iterations, which are computing estimator $\nu_{t,j}^k\in \mathbb{R}^{d_2\times d_2}$ for Hessian  $\nabla^2_{\m{y}}g$ via consensus and
stochastic approximation. Each of these $b$ iterations requires the exchange of the matrix $\nu_{t,i}^j$ with neighbors $j\in \mathcal{N}_k$. Hence, the total inner communication of this inner loop is $b$ matrices of $d_2\times d_2$ dimension.
    \item [II).] Communication complexity of DGBO:
    We first consider the computational complexity of each iteration. Step 4 involves exchanging the iterate $\mathrm{x}_t^j\in \mathbb{R}^{d_1}$ with neighbors $j\in \mathcal{N}_k$, which requires a single communication of a $d_1$-dimensional
vector. Step 5 exchanges the iterate $\mathrm{y}_t^j\in \mathbb{R}^{d_2}$ with neighbors $j\in \mathcal{N}_k$, which requires one communication of a $d_2$-dimensional
vector. Step 6 exchanges the iterate $\mathrm{s}_t^j\in \mathbb{R}^{d_1}$ (estimator of $\nabla_{\mathrm{x}} f$) with neighbors $j\in \mathcal{N}_k$, which requires one communication of a $d_1$-dimensional
vector. Step 7 exchanges the iterate $\mathrm{h}_t^j\in \mathbb{R}^{d_2}$ (estimator of $\nabla_{\mathrm{y}} f$) with neighbors $j\in \mathcal{N}_k$, which requires one communication of a $d_2$-dimensional
vector. Step 8 exchanges the iterate $\mathrm{u}_t^j\in \mathbb{R}^{d_1\times d_2}$ ( estimator of full Jacobian matrices $\nabla^2_{\mathrm{x}\mathrm{y}}g$) with neighbors $j\in \mathcal{N}_k$, which requires one communication of a $d_1\times d_2$-dimensional
matrix.
 Finally, Steps 10-13 call the inner loop, which requires b matrices of $d_2\times d_2$ dimension.
   Overall, the total number of iterations is
    $K(bd_2^2+2(d_1+d_2)+d_1d_2)$. Finally, by choosing the total number of outer iterations $K=\mathcal{O}\left(\frac{n}{\epsilon(1-\sigma)^2} \right)$ and the number of inner iterations $b=\mathcal{O}\left(\log (\frac{1}{\epsilon})\right)$ to estimate the global Hessian via a Neumann series-based approach,
the communication complexity of  DGBO is
$\mathcal{O}\left(\frac{1}{\epsilon(1-\sigma)^2}\left(d_2^2\log (\frac{1}{\epsilon}) + d_1 + d_2 + d_1d_2\right)\right).$
\end{enumerate}
$\bullet$ \textbf{Communication complexity of DGTBO \cite{chen2022decentralized} :}
     Similarly,  we can divide the communication complexity of  decentralized bilevel optimization algorithm (DGTBO) in \cite{chen2022decentralized} into three parts: Algorithm 3 (Decentralized Bilevel Optimization (DGTBO)) and
Subalgorithm 2 (Hypergradient estimate) which contains Subalgorithm 1 (Jacobian-Hessian-Inverse Product oracle (JHIP)).
\begin{enumerate}
 \item[I).]
Communication complexity of (JHIP): As we can see, this algorithm operates in $n\times N$ inner iterations, which is essentially a
decentralized algorithm.
In Step 3,
each of these $n\times N$ iterations requires the exchange of the matrix $Z_j^{(t)}$ with
neighbors $j\in \mathcal{N}_i$, which requires one communication of a $d_2\times d_1$-dimensional matrix.
In Step 5,
each of these $n \times N$ iterations requires the exchange of the matrix $Y_j^{(t)}$ with
neighbors $j\in \mathcal{N}_i$, which requires one communication of a $d_2\times d_1$-dimensional matrix.
Hence, the total inner communication of JHIP is $n\times N$ matrices of $2(d_2\times d_1)$ dimension.
\item[II).]
Communication complexity of Algorithm 2 (Hypergradient estimate): Step 11 or Step 14 of this subalgorithm calls Subalgorithm 1, which requires $n\times N$ matrices of $d_2\times d_1$ dimension.

\item[III).]
Communication complexity of DGTBO:
  Step 8 exchanges the iterate
$\nu_{j,k}^{(t-1)}$ with neighbors $j\in \mathcal{N}_{i}$, which requires one communication of a
$d_2$-dimensional vector.
Step 9 exchanges the iterate
$\mathrm{y}_{j,k}^{(t)}$ with neighbors $j\in \mathcal{N}_{i}$, which requires one communication of a
$d_2$-dimensional vector.
These steps operate in $n\times M $. Thus,  the total inner communication of steps 5-10 is $n\times M$ vectors of  $2d_2$ dimension.
Step 14 calls Algorithm 2, which requires $n\times N$ matrices of $d_2\times d_1$ dimension.
Step 15 exchanges the iterate
$\mathrm{x}_{j,k}$ with neighbors $j\in \mathcal{N}_{i}$, which requires one communication of a
$d_1$-dimensional vector.
Overall, since the algorithm operates over $K$ iterations, the total number of iterations is $Kn\left( Md_2+d_1+nNd_1d_2\right)$.
Finally, by choosing the total number of outer iterations $K=\mathcal{O}\left(\frac{1}{n\epsilon (1-\sigma)^2} \right)$, the number of inner iterations $M=\mathcal{O}\left(\log (\frac{1}{\epsilon})\right)$, and the number of JHIP oracle calls $N=\mathcal{O}\left(\log (\frac{1}{\epsilon})\right)$ to compute the global Jacobian-Hessian-Inverse
product using a decentralized optimization approach,
the communication complexity of  DGTBO is $\mathcal{O}\left(\frac{1}{\epsilon(1-\sigma)^2 }\left(( d_2+d_1d_2)\log (\frac{1}{\epsilon})+d_1\right)\right)$.
\end{enumerate}

\section{Proofs of the Main Theorems}\label{apen}
\begin{table}[H]
	{\footnotesize
  \begin{tabular}{| c || l | }
  \hline
     $\mathbb{R}^d$ &  The $d$-dimension Euclidean space
     \\
    \hline
    $\mathbb{R}^{d_1 \times d_2}$ &  The set of $d_1$-by-$d_2$ real matrices
     \\
    \hline
    $\mathbb{R}_{+}$ & \stackanchor{The set of non-negative}{\footnotesize real numbers}
    \\
    \hline
    $\mathbb{R}_{++}$ &  The set of positive real numbers
     \\
    \hline
     $\vphantom{\sum^N} [n]$ &  \stackanchor{The set $\{1,2,...,n\}$}{\footnotesize for any integer $n$}
     \\ \hline
     $\lceil x \rceil$ & \stackanchor{ The least integer greater than} {\footnotesize or equal to $x$}
     \\ \hline
     $\vphantom{\sum^N} {\mr x}^\top$ & Transpose of the vector $\xx$
     \\ \hline
     $\vphantom{\sum^N} \x\in \mathbb{R}^{nd}$ & \stackanchor{Concatenation of local vectors}{\footnotesize  $\mr x_i \in \mathbb{R}^d$ as $[\mathrm{x}_1; \ldots ; \mathrm{x}_n]$}
     \\ \hline
     $\vphantom{\sum^N} \mr I_d$ &  Identity matrix of size $d$ \\ \hline
     $\vphantom{\sum^N} \mathrm{1}_d$ &  The all-one vector \\ \hline
     $\vphantom{\sum^N} \langle\cdot, \cdot \rangle$ &  Standard inner product operator
     \\ \hline
     $|\cdot |$ & \stackanchor{The
absolute value of a real }{\footnotesize number or the cardinality of a set}
     \\ \hline
     $\otimes$ &  Kronecker product of matrices \\ \hline
     $\vphantom{\sum^N} \|\cdot\|$ &  The $\ell_2$--norm of a vector \\ \hline
     $\vphantom{\sum^N} \| \x\|$ &  $\sum_{i=1}^{n}\|\mathrm{x}_i\|$
     \\ \hline
     $\textnormal{diag}(\mr W)$& \stackanchor{The diagonal components of the}{\footnotesize matrix $\mr W$}
     \\ \hline
     $\vphantom{\sum^N} \textnormal{null}\{\mr W\}$ &  The null space of matrix $\mr W$ \\ \hline
     $\vphantom{\sum^N} \textnormal{span}\{\xx\}$ &  The span of the vector $\xx$
     \\ \hline $a \mr I_d\preceq \mr W\preceq b \mr I_d$& \stackanchor{The eigenvalues of $\mr W$ lie }{\footnotesize in $[a,b]$ interval}
     \\\hline
     $\vphantom{\sum^N} \lambda_i(\mr W)$ &  \stackanchor{The $i$-th largest eigenvalue of }{\footnotesize  matrix $\mr W$ }\\ \hline
   \stackanchor{  $\lambda_{\max}(\mr W)$, $\lambda_{\min}(\mr W)$,}{\footnotesize and $\hat{\lambda}_{\min}(\mr W)$ }&
    \stackanchor{ The largest, smallest, and smallest }{\footnotesize nonzero eigenvalues of matrix $\mr W$} \\ \hline
  \end{tabular}}
\vspace{.2cm}
    \caption{Summary of Notations and Terminologies}
	\label{table:summary:notation}
\end{table}
In this section, we present the proofs of Theorems \ref{thm:Strongly-Convex}, \ref{thm:convex}, and \ref{thm:Non-Convex}.
First, we introduce some technical lemmas used in our analysis. The proofs of these lemmas are deferred to Section \ref{sec:aux:lem}.
\begin{lm}\label{lem:lip}
Let $\mu_{\mathbf{G}}=\hat{\lambda}_{\min}(\textbf{I}_{nd_2}-\W)+ \mu_g$ and $\beta<1$.
Consider the definitions of objective function $\F$ and $\check{\y}^*(\x)$ in \eqref{eqn:reform:dblo}.
The following statements hold.
\begin{enumerate}
\item \label{L1} Under Assumptions \ref{assu:netw} and \ref{assu:lip}, we have
 \begin{equation} \label{eq:lipschitz}
    \left\|
\tilde{\nabla} \mathbf{F}(\x,\y)-
\nabla \mathbf{F}(\x,\check{\y}^*(\x)) \right\|\le C\|\check{\y}^*(\x)- \y\|,
\end{equation}
where $\tilde{\nabla} \mathbf{F}$ is defined in \eqref{eqn:grad:tildF}, and
\begin{align*}
\nonumber C &:=  L_{f_{\xx}}+\frac{ C_{g_{\xx\yy}}L_{f_{\yy}}}{ \mu_{\mathbf{G}}}
+  C_{f_{\yy}}\left(\frac{ L_{g_{\xx\yy}}}{\mu_{\mathbf{G}}}+\frac{  C_{g_{\xx\yy}}L_{g_{\yy\yy}}}{\mu_{\mathbf{G}}^2}\right).
\end{align*}
\item \label{L2} Under Assumptions  \ref{assu:netw}, \ref{assu:g:Cgxy} and \ref{assu:g:strongly}, $\check{\y}^*(\x)$ defined in \eqref{eqn:reform:dblo} is Lipschitz continuous in $\x$ with constant
$\varrho:={  C_{g_{\xx\yy}}}/{ \mu_g}.$
\item \label{L3} Under Assumptions \ref{assu:netw} and \ref{assu:lip}, $\nabla \mathbf{F}$ is Lipschitz continuous in $\x$ with constant $L_{F}$ i.e., for any
given $\x_1, \x_2 \in \mathbb{R}^{nd_1}$, we have
\begin{equation*}
  \|\nabla \mathbf{F}(\x_2,\check{\y}^*(\x_2))-\nabla \mathbf{F}(\x_1,\check{\y}^*(\x_1)) \|\leq L_{F}\|\x_2-\x_1 \|,
\end{equation*}
with
\begin{align}\label{lf}
\nonumber L_{F}&:=\frac{(\tilde{L}_{f_{\yy}}+ C)\cdot C_{g_{\xx\yy}}}{\mu_{\mathbf{G}}}+ L_{f_{\xx}}
\\&+  C_{f_{\yy}}\left(\frac{ \tilde{L}_{g_{\xx\yy}}C_{f_{\yy}}}{\mu_{\mathbf{G}}}
+\frac{  C_{g_{\xx\yy}}\tilde{L}_{g_{\yy\yy}}}{\mu_{\mathbf{G}}^2}\right).
\end{align}
\end{enumerate}
\end{lm}
\begin{lm}\label{lem:error}
Suppose  Assumptions \ref{assu:netw} and \ref{assu:lip} hold.
Set parameter $\beta$ as in Eq. \eqref{gabb}. If we select $M= \mathcal{O}(1/\beta)$, then we have
\begin{align}\label{L111}
\nonumber&\quad \|\widehat{\nabla} \mathbf{F}(\x_k,\y^M_k)-\nabla \mathbf{F}(\x_k,\check{\y}^*(\x_k))\|^2\\\nonumber&\leq
2\eta^2\rho^{2(U+1)}+2C^2 \left(1- \beta b_g\right)^{M}
\\&\cdot\left(\big(\frac{1}{2}\big)^k P_0+2\varrho^2 \sum_{j=0}^{k-1}\left(\frac{1}{2}\right)^{k-1-j}\|\x_{j}-\x_{j+1} \|^2\right),
\end{align}
where $\widehat{\nabla} \mathbf{F}(\x_k,\y^M_k)$ is defined in Eq. \eqref{eqn:d-k-U},
 $\rho$ is defined in Lemma \ref{symmetric_term_bounds11}, and
\begin{align}\label{var}
\eta:=\frac{\beta n^2 C_{g_{\xx\yy}} C_{f_{\yy}}}{(2(1-\Theta)+{\beta \mu_g} )(1-\rho)}.
\end{align}

\end{lm}

\begin{lm}\label{lem:up}
Consider the definition of the objective function $\F$ in \eqref{eqn:approximate:prob}.
If Assumptions \ref{assu:netw}, \ref{assu:f:grad}, \ref{assu:g:Cgxy}, and \ref{assu:g:strongly} hold, then we have
\begin{align*}
 1^{\top}\mathbf{f}(\bar \x,\y^*(\bar \x))-1^{\top}\mathbf{f}^*&\leq \F(\x,\y^*(\x))-\mathbf{F}({\x}^*,{\y}^*({\x}^*))
 \\&+\frac{  \tilde{C}\hat{C}  \alpha}{1-\sigma},
\end{align*}
where $\tilde{C}$ and $\hat{C}$ are defined as
\begin{equation}\label{eqn:po}
    \begin{split}
   \tilde{C}&:=C_{f_{\xx}}+\frac{2C_{g_{\xx\yy}} \Lambda C_{f_{\yy}}}{\mu_{g}+L_{g}},\quad \textnormal{and}\\
   \hat{C}&:= C_{f_{\xx}}+C_{f_{\yy}}.
    \end{split}
\end{equation}
\end{lm}
\begin{lm}\label{f2}
Consider the definition of objective function $\F$ in \eqref{eqn:approximate:prob}.
Let $\beta$ satisfies \eqref{eqn:beta},
\begin{align*}
&\alpha  \leq \frac{1}{2L_F},~ U=\left|\left\lceil\frac{\log ((1-\beta b_g)^{k+1}/K\eta^2)}{2\log (\rho)} \right\rceil\right|,~
\\&M \geq \max\left\{k+1,\frac{\log (\alpha)}{ \beta}\right\},
\end{align*}
where $\eta$ and $\rho$ are defined in \eqref{var} and Lemma \ref{symmetric_term_bounds11}, respectively.
Let $\mathbf{F}^*:= \mathbf{F}(\check{\x}^*,\check{\y}^*(\check{\x}^*))$.
If Assumptions \ref{assu:netw}, \ref{assu:f:grad} and \ref{assu:g:Cgxy} hold then we have
\begin{align}\label{q1}
 \nonumber  &\quad \F(\x_{K},\check{\y}^*(\x_{K}))-\F^*
\\&\leq
\Gamma_{K} \left(  \F(\x_{0},\check{\y}^*(\x_{0}))-\F^*  +\frac{1}{L_F} (\frac{1}{2}+ C^2 P_0)\right),
\end{align}
where $\Gamma_k:=(1-\nu)^{k}$ and $\nu$ is defined in \eqref{gk}. 
$P_0$ is defined in \eqref{dfc}. Constants $L_F$ and $C$ are defined in Lemma \ref{lem:lip}.
\end{lm}
\begin{proof}
By using the smoothness of $\mathbf{F}$ due to  Lemma \ref{lem:lip}--(\ref{L3}), we have
  \begin{align}\label{nkz}
   \nonumber &\quad \F(\x_{k+1},\check{\y}^*(\x_{k+1}))\leq \F(\x_{k},\check{\y}^*(\x_{k}))
    \\&+\langle \nabla \F(\x_{k},\check{\y}^*(\x_{k})),\x_{k+1}-\x_k  \rangle+\frac{L_{F}}{2}\|\x_{k+1}-\x_k \|^2.
  \end{align}
  At this point, we would proceed by bounding the second term, for all $\uu\in\mathbb{R}^{nd_1}$, as
  \begin{align}\label{nv}
 \nonumber  &\quad \langle \nabla \F(\x_{k},\check{\y}^*(\x_{k})),\x_{k+1}-\x_k  \rangle
 \\\nonumber & =
    \langle \nabla \F(\x_{k},\check{\y}^*(\x_{k})),\x_{k+1}-\uu  \rangle
    \\\nonumber &+\langle \nabla \F(\x_{k},\check{\y}^*(\x_{k})),\uu-\x_k  \rangle\\\nonumber
    &=\langle \nabla \F(\x_{k},\check{\y}^*(\x_{k}))-\widehat{\nabla} \F(\x_{k},\y_k^M),\x_{k+1}-\uu  \rangle
   \\\nonumber & +\langle \widehat{\nabla} \F(\x_{k},\y_k^M),\x_{k+1}-\uu  \rangle+\langle \nabla \F(\x_{k},\check{\y}^*(\x_{k})),\uu-\x_k  \rangle\\\nonumber
   &=\langle \nabla \F(\x_{k},\check{\y}^*(\x_{k}))-\widehat{\nabla} \F(\x_{k},\y_k^M),\x_{k+1}-\uu  \rangle\\\nonumber
   &+\frac{1}{2\alpha}\big(\|\uu-\x_k \|^2-\|\uu-\x_{k+1} \|^2\big)\\
   &-\frac{1}{2\alpha}\|\x_{k+1}-\x_{k} \|^2+\langle \nabla \F(\x_{k},\check{\y}^*(\x_{k})),\uu-\x_k  \rangle,
  \end{align}
 where the last equality holds since by \eqref{eqn:update:DAGM} and identity $(\mathrm{a}-\mathrm{b})^\top  (\mathrm{a}-\mathrm{c})
=(1/2)\Big\{\|  \mathrm{a}-\mathrm{c}\|^2-\| \mathrm{ c}-\mathrm{b}\|^2+\| \mathrm{ a}-\mathrm{b}\|^2 \Big\}$, we have
 \begin{align*}
 &\quad \langle \widehat{\nabla} \F(\x_{k},\y_k^M),\x_{k+1}-\uu  \rangle  =\frac{1}{\alpha} \langle \x_{k}-\x_{k+1},\x_{k+1}-\uu  \rangle\\
  &=\frac{1}{2\alpha}\big(\|\uu-\x_k \|^2-\|\uu-\x_{k+1} \|^2-\|\x_{k+1}-\x_{k} \|^2\big).
 \end{align*}
Next, by denoting $\widehat{\Delta}_k:= \widehat{\nabla} \mathbf{F}(\x_k,\y_k^M)- \nabla \mathbf{F}(\x_k,\check{\y}^*(\x_k))$ and substituting \eqref{nv} into \eqref{nkz}, we
have the following inequality
\begin{align}\label{nkxx}
 \nonumber&\quad \F(\x_{k+1},\check{\y}^*(\x_{k+1}))
 \\\nonumber& \leq \F(\x_{k},\check{\y}^*(\x_{k}))+\langle \nabla \F(\x_{k},\check{\y}^*(\x_{k})),\uu-\x_k  \rangle
 \\\nonumber&+\frac{1}{2\alpha}\big(\|\uu-\x_k \|^2-\| \uu-\x_{k+1}\|^2\big)\\
  &-\frac{(1-L_{F}\alpha)}{2\alpha}\|\x_{k+1} -\x_k\|^2+\langle \widehat{\Delta}_k,\uu-\x_{k+1} \rangle,
\end{align}
for which, invoking the fact that
\begin{align*}
  \langle \widehat{\Delta}_k,\uu-\x_{k+1} \rangle &\leq \|\widehat{\Delta}_k \|\cdot \| \uu-\x_{k+1}\|
  \\&\leq \frac{\alpha}{2}\|\widehat{\Delta}_k \|^2+\frac{1}{2\alpha}\| \uu-\x_{k+1} \|^2,
\end{align*}
it can be
obtained that
\begin{align*}
\nonumber& \quad \F(\x_{k+1},\check{\y}^*(\x_{k+1}))
\\&\leq \F(\x_{k},\check{\y}^*(\x_{k}))+\langle \nabla \F(\x_{k},\check{\y}^*(\x_{k})),\uu-\x_k \rangle
\\&
+\frac{1}{2\alpha}\| \uu-\x_k\|^2
-\frac{(1-L_{F}\alpha)}{2\alpha}\|\x_{k+1} -\x_k\|^2
+\frac{\alpha}{2}\|\widehat{\Delta}_k \|^2.
\end{align*}
Subsequently, by setting $\uu=\zeta\check{\x}^*+(1-\zeta)\x_k$ for some $\zeta\in [0,1]$ in the above inequality, one has that
\begin{align*}
\nonumber &\quad  \F(\x_{k+1},\check{\y}^*(\x_{k+1}))
 \leq (1-\zeta)\F(\x_{k},\check{\y}^*(\x_{k}))
 \\& +\zeta\big(\F(\x_{k},\check{\y}^*(\x_{k}))
+\langle \nabla \F(\x_{k},\check{\y}^*(\x_{k})),\check{\x}^*-\x_k \rangle\big)
  \\\nonumber &+\frac{\zeta^2}{2\alpha}\|\check{\x}^*-\x_k \|^2-\frac{(1-L_{F}\alpha)}{2\alpha}\|\x_{k+1} -\x_k\|^2+\frac{\alpha}{2}\|\widehat{\Delta}_k \|^2.
\end{align*}
As $\sum_{i=1}^n f_i\left(\xx_i,\check{\yy}_i^*(\xx_i)\right)$ is $\mu_f$-strongly convex, then $\F$ is strongly convex with parameter:
\begin{equation*}
\begin{split}
\mu_{F}&:=\mu_f+(2\alpha)^{-1}\lambda_{\min}(\textbf{I}_{nd_1}-\acute{\W})
\\&=\mu_f+(2\alpha)^{-1}(1-\sigma).
\end{split}
\end{equation*}
Let $\mathbf{F}^*:= \mathbf{F}(\check{\x}^*,\check{\y}^*(\check{\x}^*))$.
Then, we have
\begin{align*}
\nonumber &\quad \F(\x_{k+1},\check{\y}^*(\x_{k+1}))
\\& \leq (1-\zeta)\F(\x_{k},\check{\y}^*(\x_{k})) +\zeta \F^*-\frac{\zeta\mu_{F}}{2}\|\check{\x}^*-\x_k \|^2
  \\ &+\frac{\zeta^2}{2\alpha}\|\check{\x}^*-\x_k \|^2-\frac{(1-L_{F}\alpha)}{2\alpha}\|\x_{k+1} -\x_k\|^2 +\frac{\alpha}{2}\|\widehat{\Delta}_k \|^2.
\end{align*}
Then, with $\zeta=\alpha\mu_{F}$ which is less than $1$ since $\alpha\leq 1/(2L_F)$, this becomes:
\begin{align*}
\nonumber  &\quad \F(\x_{k+1},\check{\y}^*(\x_{k+1}))
 \leq (1-\alpha\mu_{F})\F(\x_{k},\check{\y}^*(\x_{k}))\\ &+\alpha\mu_{F} \F^*-\frac{(1-L_{F}\alpha)}{2\alpha}\|\x_{k+1} -\x_k\|^2+\frac{\alpha}{2}\|\widehat{\Delta}_k \|^2.
\end{align*}
We proceed by subtracting $\mathbf{F}^*$ from both sides of the above inequality to obtain
\begin{align}\label{sd}
&\quad \nonumber  \F(\x_{k+1},\check{\y}^*(\x_{k+1}))-\F^*
 \\\nonumber& =  (1-\alpha\mu_{F}) \left(\F(\x_{k},\check{\y}^*(\x_{k}))-\F^*\right)
  \\\nonumber&-\frac{(1-L_{F}\alpha)}{2\alpha}\|\x_{k+1} -\x_k\|^2+\frac{\alpha}{2}\|\widehat{\Delta}_k \|^2
  \\\nonumber& \leq  (1-\nu) \left(\F(\x_{k},\check{\y}^*(\x_{k}))-\F^*\right)
  \\&-\frac{(1-L_{F}\alpha)}{2\alpha}\|\x_{k+1} -\x_k\|^2+\frac{\alpha}{2}\|\widehat{\Delta}_k \|^2,
\end{align}
where $\nu=\min \left( \alpha \mu_F ,\beta b_g\right)$.
Let $\Gamma_k:=(1-\nu)^{k}$.
Dividing both sides of Eq. \eqref{sd} by $\Gamma_{k+1}$, summing over $k = 0,\ldots,K-1$,  and invoking Lemma \ref{lem:error}, yield
\begin{align}\label{cxz}
\nonumber  &\quad \sum_{k=0}^{K-1}\frac{1}{\Gamma_{k+1}}\big( \F(\x_{k+1},\check{\y}^*(\x_{k+1}))-\F^*\big)
\\\nonumber&\leq
\sum_{k=0}^{K-1}\frac{1}{\Gamma_{k}}\big(\F(\x_{k},\check{\y}^*(\x_{k}))-\F^*\big)
\\\nonumber&-\frac{(1-L_{F}\alpha)}{2\alpha}\sum_{k=0}^{K-1}\frac{1}{\Gamma_{k+1}}\|\x_{k+1} -\x_k\|^2
\\\nonumber &+\sum_{k=0}^{K-1}\frac{\alpha \rho^{2(U+1)}\eta^2}{\Gamma_{k+1}}
\\\nonumber &+\alpha C^2 \left( 1-\beta b_g\right)^{M} \left(\sum_{k=0}^{K-1}\frac{1}{\Gamma_{k+1}}\left(\frac{1}{2}\right)^k P_0
\right.\\&\left.+2\varrho^2 \sum_{k=1}^{K-1}\sum_{j=0}^{k-1}\frac{1}{\Gamma_{k+1}}\left(\frac{1}{2}\right)^{k-1-j}\|\x_{j}-\x_{j+1} \|^2\right).
\end{align}
For the last term on the right-hand side of Eq. \eqref{cxz}, we have
\begin{subequations}
\begin{align}\label{njk}
\nonumber &\quad\sum_{k=1}^{K-1}\sum_{j=0}^{k-1}\frac{1}{\Gamma_{k+1}}\big(\frac{1}{2}\big)^{k-1-j}\|\x_{j}-\x_{j+1} \|^2
\\\nonumber&\leq \sum_{k=0}^{K-1}\frac{1}{2^k}\sum_{k=0}^{K-1}\frac{1}{\Gamma_{k+1}}\|\x_{k}-\x_{k+1} \|^2
\\&\leq2 \sum_{k=0}^{K-1}\frac{1}{\Gamma_{k+1}}\|\x_{k}-\x_{k+1} \|^2.
\end{align}
Furthermore, take $M$ such that
\begin{align}\label{bbnlb}
 4 \varrho^2\alpha C^2\left( 1-\beta b_g\right)^{M}\leq \frac{L_F}{2}.
\end{align}
\end{subequations}
It is easy to verify that \eqref{bbnlb} holds when
\begin{align*}
M\geq \frac{\log \left( {L_F}/{8 \varrho^2\alpha C^2} \right)}{\log\left( 1-\beta b_g\right)}=\mathcal{O}\left(\frac{\log (\alpha)}{ \beta}\right).
\end{align*}
Plugging \eqref{njk} and \eqref{bbnlb} into \eqref{cxz} and noting that $\alpha \leq 1/(2L_F)$ yields
\begin{align*}
 \nonumber  &\quad \F(\x_{K},\check{\y}^*(\x_{K}))-\F^*
\leq
\Gamma_{K} \left(  \F(\x_{0},\check{\y}^*(\x_{0}))-\F^*
\right.\\&\left.+\frac{1}{2L_F}\sum_{k=0}^{K-1}\left(\frac{ \rho^{2(U+1)}\eta^2}{\Gamma_{k+1}}+\frac{ C^2 P_0}{\Gamma_{k+1}2^k}\left( 1-\beta b_g\right)^{M}\right)\right).
\end{align*}
Let $M\geq k+1$. Since $\nu=\min \Big( \alpha \mu_F ,\beta b_g\Big)$, $\Gamma_{k+1}:=(1-\nu)^{k+1}$, we get
\begin{align*}
   \sum_{k=0}^{K-1} \frac{\left( 1-\beta b_g\right)^{M}}{\Gamma_{k+1}2^k}&\leq \sum_{k=0}^{K-1} \frac{\left( 1-\beta b_g\right)^{M}}{\left( 1-\beta b_g\right)^{k+1}2^k}
  \\&\leq \sum_{k=0}^{K-1} \frac{1}{2^k}< 2.
\end{align*}
This result yields
\begin{align}\label{e:1}
 \nonumber  &\quad \F(\x_{K},\check{\y}^*(\x_{K}))-\F^*
\\\nonumber &\leq
\Gamma_{K} \left(  \F(\x_{0},\check{\y}^*(\x_{0}))-\F^*
\right.\\&\left.+\frac{1}{2L_F}\sum_{k=0}^{K-1} \frac{\rho^{2(U+1)}\eta^2}{\left( 1-\beta b_g\right)^{k+1}}
+\frac{C^2  P_0}{L_F}\right).
\end{align}
Let
$$U=\left|\left\lceil\frac{\log ((1-\beta b_g)^{k+1}/K\eta^2)}{2\log (\rho)} \right\rceil\right|.$$
Then, since $\rho<1$ in Lemma \ref{symmetric_term_bounds11}, we have
\begin{align}\label{r:2}
\eta^2\rho^{2(U+1)}\leq \eta^2\rho^{2U}= \frac{\left( 1-\beta b_g\right)^{k+1}}{K},
\end{align}
which implies that
\begin{align*}
\sum_{k=0}^{K-1} \frac{\rho^{2(U+1)}\eta^2}{\left( 1-\beta b_g\right)^{k+1}}< 1.
\end{align*}
Combining this with \eqref{e:1}, we have
\begin{align*}
 \nonumber  &\quad \F(\x_{K},\check{\y}^*(\x_{K}))-\F^*
\\&\leq
\Gamma_{K} \left(  \F(\x_{0},\check{\y}^*(\x_{0}))-\F^*
+\frac{1}{2L_F}
+\frac{C^2  P_0}{L_F}\right).
\end{align*}
\end{proof}
%

\subsection{Proof of Theorem~ \ref{thm:Strongly-Convex}}
\begin{proof}
Note that the bounded gradient holds if $f_i$ is Lipschitz continuous.
Since $\sum_{i=1}^n f_i\left(\xx_i,\check{\yy}_i^*(\xx_i)\right)$ is $C_f:=C_{f_{\xx}}+C_{f_{\yy}}$-lipschitz continuous, it follows that $\F$ is lipschitz continuous with the parameter:
\begin{equation}\label{fsa}
\begin{split}
C_{\F}&:=C_f+(2\alpha)^{-1}\lambda_{\min}(\textbf{I}_{nd_1}-\acute{\W})
\\&=C_f+(2\alpha)^{-1}(1-\sigma).
\end{split}
\end{equation}
\begin{align}
 \nonumber    &\quad \F(\x_{K},{\y}^*(\x_{K}))-\mathbf{F}({\x}^*,{\y}^*({\x}^*))
 \\\nonumber &\leq \F(\x_{K},{\y}^*(\x_{K}))-\mathbf{F}(\check{\x}^*,\check{\y}^*(\check{\x}^*))
\\\nonumber &= \F(\x_{K},\y^*(\x_{K}))-\F(\x_{K},\check{\y}^*(\x_{K}))
\\\nonumber &+\F(\x_{K},\check{\y}^*(\x_{K}))-\mathbf{F}(\check{\x}^*,\check{\y}^*(\check{\x}^*))
\\\nonumber &\leq C_{\F}\|\y^*(\x_{K})-\check{\y}^*(\x_{K}) \|
\\\nonumber&+\F(\x_{K},\check{\y}^*(\x_{K}))-\mathbf{F}(\check{\x}^*,\check{\y}^*(\check{\x}^*))
\\\nonumber &\leq C_{\F}\left(\frac{2\hat{L}_g\beta}{(1-\sigma)}+\frac{2\hat{L}_g\beta^{1/2}}{(1-\sigma)^{1/2}\mu_g^{1/2}}\right)
\\\nonumber &+\Gamma_{K} \left(  \F(\x_{0},\check{\y}^*(\x_{0}))-\F^*
\right.\\&\left.+\frac{1}{2L_F} (2\rho^{2}\eta^2+ C^2 P_0)\right),
\end{align}
where the last inequality is by Lemma \ref{cl} and \eqref{q1}.

Thus, from Lemma \ref{lem:up}, we have
\begin{align*}
   &\quad 1^{\top}\mathbf{f}(\bar\x_{K},{\y}^*(\bar\x_{K}))-1^{\top}\mathbf{f}^*
   \\&\leq \F(\x_{K},{\y}^*(\x_{K}))-\mathbf{F}({\x}^*,{\y}^*({\x}^*))+\frac{ \hat{C}   \tilde{C}\alpha}{1-\sigma}
    \\&\leq \Gamma_{K} \left( D_{F}  +\frac{1}{2 L_F} \big(2\rho^{2}\eta^2+ C^2 P_0\big)\right)
    \\&+C_{\F}\left(\frac{2\hat{L}_g\beta}{(1-\sigma)}+\frac{2\hat{L}_g\beta^{1/2}}{(1-\sigma)^{1/2}\mu_g^{1/2}}\right)+\frac{ \hat{C}   \tilde{C}\alpha}{1-\sigma},
\end{align*}
which
verifies the conclusion.
\end{proof}
\subsection{Proof of Corollary~ \ref{cor:Strongly-Convex}}
Substituting $\alpha = \mathcal{O}(\delta^{K})$ for some $\delta<1$ and  $\beta = \mathcal{O}(n^{-1}\delta^{2(b+K)})$ with
$b\geq \log(n\bar{\beta})^{1/2}$ into the following inequality,
\begin{align*}
&\quad  \frac{1}{n}1^{\top}\left(\mathbf{f}(\bar\x_{K},\y^*(\bar\x_{K}))-\mathbf{f}^*\right)
\\& \leq \frac{1}{n}\left( D_{F}  +\frac{1}{ L_F} \left(\frac{1}{2}
+C^2  P_0\right)\right) (1-\nu)^{K}\\
& ~ +\frac{ \hat{C}\tilde{C}  }{n(1-\sigma)} \alpha+\frac{ 2\hat{L}_gC_{\F}}{n(1-\sigma)^{\frac{1}{2}}}\left(\frac{\beta^{\frac{1}{2}}}{(1-\sigma)^{\frac{1}{2}}}+\mu_g^\frac{-1}{2}\right) \beta^{\frac{1}{2}}
\\&=\mathcal{O}\left( \log (n^{-1}\epsilon^{-1}/(1-\sigma))\right),
\end{align*}
gives the desired result.
\begin{lm}\label{lem:dif:convex}
Under the same assumptions and the parameters of Theorem~\ref{thm:convex}, we have
\begin{align*}
\nonumber &\quad \sum_{k=0}^{K-1}\|\widehat{\nabla} \mathbf{F}(\x_k,\y^M_k)- \nabla \mathbf{F}(\x_k,\check{\y}^*(\x_k)) \|\leq\tau,
\end{align*}
where $\tau:= 2
+ 2 \sqrt{P_0}+2\varrho
+   \sqrt{2L_FD_F}$, and $\varrho$, and $L_{F}$, are defined in Lemma \ref{lem:lip}. $D_{F}$ and $P_0$ are defined in \eqref{dfc}.
\end{lm}


\subsection{Proof of Theorem~\ref{thm:convex} }
\begin{proof}
We let $\uu=\check{\x}^*$ in \eqref{nkxx} to conclude that
\begin{align*}
 \nonumber&\quad \F(\x_{k+1},\check{\y}^*(\x_{k+1}))
 \\& \leq \F(\x_{k},\check{\y}^*(\x_{k}))+\langle \nabla \F(\x_{k},\check{\y}^*(\x_{k})),\check{\x}^*-\x_k  \rangle
 \\&+\frac{1}{2\alpha}\left(\|\check{\x}^*-\x_k \|^2-\| \check{\x}^*-\x_{k+1}\|^2\right)
 \\\nonumber
  &-\frac{(1-L_{F}\alpha)}{2\alpha}\|\x_{k+1} -\x_k\|^2+\langle \widehat{\Delta}_k,\check{\x}^*-\x_{k+1} \rangle\\\nonumber
  &\leq \mathbf{F}(\check{\x}^*,\check{\y}^*(\check{\x}^*))+\frac{1}{2\alpha}\left(\|\check{\x}^*-\x_k \|^2-\| \check{\x}^*-\x_{k+1}\|^2\right)
  \\&+\| \widehat{\Delta}_k\| \ \|\check{\x}^*-\x_{k+1} \|,
\end{align*}
where $\widehat{\Delta}_k:= \widehat{\nabla} \mathbf{F}(\x_k,\y_k^M)- \nabla \mathbf{F}(\x_k,\check{\y}^*(\x_k))$; the second inequality is by the convexity of $\F$, assumption $\alpha\leq 1/L_F$, and Cauchy-Schwarz.

Let $\mathbf{F}^*:= \mathbf{F}(\check{\x}^*,\check{\y}^*(\check{\x}^*))$. Summing up both sides of the above inequality, we obtain
\begin{align}\label{cz}
\nonumber &\quad \sum_{k=0}^{K-1}\left( \F(\x_{k+1},\check{\y}^*(\x_{k+1}))-\F^*\right)
\\\nonumber & \leq \sum_{k=0}^{K-1}\Big(\frac{1}{2\alpha}\left(\|\check{\x}^*-\x_k \|^2-\| \check{\x}^*-\x_{k+1}\|^2\right)
\\\nonumber &+\| \widehat{\Delta}_k\| \ \|\check{\x}^*-\x_{k+1} \|\Big)
\\\nonumber & =\frac{1}{2\alpha}\left(\|\check{\x}^*-\x_0 \|^2-\| \check{\x}^*-\x_{K}\|^2\right)
\\&+\sum_{k=0}^{K-1} \| \widehat{\Delta}_k\|  \|\check{\x}^*-\x_{k+1} \|,
\end{align}
which implies
\begin{align}\label{eqn:thm:conv}
\nonumber &\quad\sum_{k=0}^{K-1}\big( \F(\x_{k+1},\check{\y}^*(\x_{k+1}))-\F^*\big) + \frac{1}{2\alpha} \| \check{\x}^*-\x_{K}\|^2
\\& \leq \frac{1}{2\alpha}\|\check{\x}^*-\x_0 \|^2+\sum_{k=0}^{K-1} \| \widehat{\Delta}_k\| \|\check{\x}^*-\x_{k+1} \|.
\end{align}
We now need to bound the quantity $\|\check{\x}^*-\x_{k+1} \|$ in terms of $\|\check{\x}^*-\x_0 \|$. Dropping
the first term in Eq. \eqref{eqn:thm:conv}, which is positive due to the optimality of $\F^*$, we have:
\begin{align*}
  \| \check{\x}^*-\x_{K}\|^2 &\leq \|\check{\x}^*-\x_0 \|^2+2\alpha\sum_{k=0}^{K-1} \| \widehat{\Delta}_k\|  \|\check{\x}^*-\x_{k+1} \|
  \\&\leq \|\check{\x}^*-\x_0 \|^2+\frac{2}{L_F}\sum_{k=0}^{K-1} \| \widehat{\Delta}_k\|  \|\check{\x}^*-\x_{k+1} \|,
\end{align*}
where the second inequality is by $\alpha\leq \frac{1}{L_F}$.
\\
We now use Lemma  \ref{lem:sequ}  with $S_K=\|\check{\x}^*-\x_0 \|^2$ and $\lambda_k=\| \widehat{\Delta}_k\|$ obtain:
\begin{align*}
 \|\check{\x}^*-\x_{K+1} \| & \leq \frac{1}{2}\sum_{k=0}^{K-1} \| \widehat{\Delta}_k\|
 \\&+\Big(\|\check{\x}^*-\x_0 \|^2+\big(\frac{1}{2}\sum_{k=0}^{K-1} \| \widehat{\Delta}_k\|\big)^2 \Big)^{1/2}.
\end{align*}
Since $\sum_{k}\| \widehat{\Delta}_k\|$ is an increasing sequence, we have:
\begin{align}\label{ds}
\nonumber \|\check{\x}^*-\x_{k+1} \| & \leq \frac{1}{2}\sum_{k=0}^{K-1} \| \widehat{\Delta}_k\|
\\\nonumber &+\Big(\|\check{\x}^*-\x_0 \|^2+\big(\frac{1}{2}\sum_{k=0}^{K-1} \| \widehat{\Delta}_k\|\big)^2 \Big)^{1/2}
\\\nonumber &\leq \sum_{k=0}^{K-1} \| \widehat{\Delta}_k\|+\|\check{\x}^*-\x_0 \|
\\&\leq  \tau+\|\check{\x}^*-\x_0 \|,
\end{align}
where the second inequality follows from $\sqrt{a+b}\leq \sqrt{a}+\sqrt{b}$ for $a,b \geq 0$ and the last inequality follows from Lemma \ref{lem:dif:convex}.

Plugging \eqref{ds} into \eqref{cz} and dividing both sides by $K$ yields:
\begin{align}\label{sa1}
\nonumber &\quad \frac{1}{K}\sum_{k=0}^{K-1}\left( \F(\x_{k+1},\check{\y}^*(\x_{k+1}))-\F^*\right)
\\\nonumber & \leq \frac{1}{2\alpha K}\left(\|\check{\x}^*-\x_0 \|^2-\| \check{\x}^*-\x_{K}\|^2\right)
\\\nonumber &+\frac{1}{K}\sum_{k=0}^{K-1} \| \widehat{\Delta}_k\|  \left(\tau+\|\check{\x}^*-\x_0 \|\right)
\\\nonumber & \leq \frac{1}{2\alpha K}\|\check{\x}^*-\x_0 \|^2+\frac{\tau}{K} \left(\tau+\|\check{\x}^*-\x_0 \|\right)
\\&=\frac{1}{2\alpha K}\|\check{\x}^* \|^2+\frac{\tau}{K} \left(\tau+\|\check{\x}^* \|\right),
\end{align}
where the last line is due to the assumption $\xx_{i,0}=0$.

Since $\F$ is Lipschitz continuous with parameter $C_{\F}$ in \eqref{fsa}, we have
\begin{align}\label{dd}
\nonumber &\quad \F(\x_{k+1},\y^*(\x_{k+1}))-\mathbf{F}({\x}^*,{\y}^*({\x}^*))
\\\nonumber &\leq \F(\x_{k+1},\y^*(\x_{k+1}))-\mathbf{F}(\check{\x}^*,\check{\y}^*(\check{\x}^*))
\\\nonumber & =\F(\x_{k+1},\y^*(\x_{k+1}))-\F(\x_{k+1},\check{\y}^*(\x_{k+1}))
\\\nonumber&+\F(\x_{k+1},\check{\y}^*(\x_{k+1}))-\mathbf{F}(\check{\x}^*,\check{\y}^*(\check{\x}^*))
\\\nonumber &\leq C_{\F} \|\y^*(\x_{k+1})- \check{\y}^*(\x_{k+1})\|
\\\nonumber &+ \F(\x_{k+1},\check{\y}^*(\x_{k+1}))-\mathbf{F}(\check{\x}^*,\check{\y}^*(\check{\x}^*))
\\\nonumber &\leq C_{\F}\left(\frac{2\hat{L}_g\beta}{(1-\sigma)}+\frac{2\hat{L}_g\beta^{1/2}}{(1-\sigma)^{1/2}\mu_g^{1/2}}\right)
\\&+ \F(\x_{k+1},\check{\y}^*(\x_{k+1}))-\mathbf{F}(\check{\x}^*,\check{\y}^*(\check{\x}^*)),
\end{align}
where the last inequality follows from Lemma \ref{cl}.

Then, from \eqref{sa1}, we have
\begin{align}\label{dd0}
\nonumber &\quad \frac{1}{K}\sum_{k=0}^{K-1}\left(\F(\x_{k+1},\y^*(\x_{k+1}))-\mathbf{F}({\x}^*,{\y}^*({\x}^*))\right)
\\\nonumber &\leq C_{\F}\left(\frac{2\hat{L}_g\beta}{(1-\sigma)}+\frac{2\hat{L}_g\beta^{1/2}}{(1-\sigma)^{1/2}\mu_g^{1/2}}\right)
\\&+\frac{1}{2\alpha K}\|\check{\x}^* \|^2+\frac{\tau}{K} (\tau+\|\check{\x}^* \|).
\end{align}
Finally, by the convexity of $\F$ and the average iterate $\widehat{\x}_{K}=(1/K)\sum_{k=1}^{K}\bar{\x}_{k}$, we have:
\begin{align*}
   &\quad \frac{1}{n}1^{\top}\mathbf{f}(\widehat{\x}_{K},\y^*(\widehat{\x}_{K}))-\frac{1}{n}1^{\top}\mathbf{f}^*
   \\ &\leq \frac{1}{nK}\sum_{k=0}^{K-1}\left(1^{\top}\mathbf{f}(\bar{\x}_{k+1},\y^*(\bar{\x}_{k+1}))-1^{\top}\mathbf{f}^*\right)\\
    &\leq  \frac{1}{nK}\sum_{k=0}^{K-1}\left(\F(\x_{k+1},\y^*(\x_{k+1}))-\mathbf{F}({\x}^*,{\y}^*({\x}^*)) \right)
    \\&+\frac{ \hat{C}  \tilde{C} \alpha}{n(1-\sigma)}
    \\
    & \leq \frac{C_{\F}}{n}\left(\frac{2\hat{L}_g\beta}{(1-\sigma)}+\frac{2\hat{L}_g\beta^{1/2}}{(1-\sigma)^{1/2}\mu_g^{1/2}}\right)
    \\&+\frac{1}{n K}\left(  \frac{1}{2\alpha}\|\check{\x}^* \|^2+\tau (\tau+\|\check{\x}^* \|) \right)+\frac{ \hat{C}   \tilde{C}\alpha}{n(1-\sigma)}
    \\&\leq \frac{C_{\F}}{n}\left(\frac{2\hat{L}_g\beta}{(1-\sigma)}+\frac{2\hat{L}_g\beta^{1/2}}{(1-\sigma)^{1/2}\mu_g^{1/2}}\right)
    \\&+\frac{1}{n K}\left( \frac{1}{2\alpha}R^2+\tau (\tau+R) \right)+\frac{  \tilde{C} \hat{C}  \alpha }{n(1-\sigma)},
\end{align*}
where the second inequality follows from Lemma \ref{lem:up}; the third inequality follows from \eqref{dd0}, and the last inequality is by $\|\check{\x}^* \|\leq R$.
\end{proof}
\subsection{Proof of Corollary~ \ref{cor:convex}}
Substituting $\alpha=\mathcal{O}(\sqrt{1-\sigma}/\sqrt{K})$ and  $\beta= \mathcal{O}(1-\sigma/(b+K))$ for some $b \geq (1-\sigma)/\bar{\beta}$ into the following inequality,
\begin{align*}
&\quad\frac{1}{n}1^{\top}\mathbf{f}(\widehat{\x}_{K},\y^*(\widehat{\x}_{K}))-\frac{1}{n}1^{\top}\mathbf{f}^*
\\&\leq  \frac{C_{\F}}{n}\left(\frac{2\hat{L}_g\beta}{(1-\sigma)}+\frac{2\hat{L}_g\beta^{1/2}}{(1-\sigma)^{1/2}\mu_g^{1/2}}\right)
    \\&+\frac{1}{n K}\left( \frac{1}{2\alpha}R^2+\tau (\tau+R) \right)+\frac{  \tilde{C} \hat{C}  \alpha }{n(1-\sigma)}
\\&=\mathcal{O}\left(\log \frac{n^{-2}\epsilon^{-2}}{(1-\sigma)}\right)
,
\end{align*}
gives the desired result.
\begin{lm}\label{thm33}
Under the same conditions stated in Theorem \ref{thm:Non-Convex}, we have
\begin{align*}
&\quad\frac{1}{K}\sum_{k=0}^{K-1}\|\frac{1}{n} 1^{\top}\nabla \F(\x_k,\y^*(\x_k))\|^2
\\&\leq \frac{16}{nK\alpha} \left(D_{F}
+\frac{5}{64L_{F}}\left(2+\frac{3\varrho^2}{16 L_F^2}
+ P_0\right)\right)
\\\nonumber &+\frac{2C^2}{n}\left(\frac{4\hat{L}_g\beta}{(1-\sigma)}+\frac{4\hat{L}_g\beta^{1/2}}{(1-\sigma)^{1/2}\mu_g^{1/2}}\right),
\end{align*}
where $C$, $\varrho$, are defined in Lemma \ref{lem:lip}. $L_F$ and $D_{F}$, $P_0$ are defined in \eqref{lf} and \eqref{dfc}, respecttively.

\end{lm}
Next, we show that the DAGM method achieves consensus across different agents in
the graph.
\begin{lm}\label{vbb}
Under the same conditions stated in Theorem \ref{thm:Non-Convex}, we have
\begin{align*}
&\quad\frac{1}{K}\sum_{k=0}^{K-1}\|\frac{1}{n} 1^{\top}(\x_{k} -\bar{\x}_k)\|^2
\\&\leq \frac{2\alpha^2}{(1-\sigma)^2} \frac{1}{K}\sum_{k=0}^{K-1} \|\frac{1}{n} 1^{\top}\nabla \mathbf{F}(\x_k,{\y}^*(\x_k))\|^2
+\frac{2\alpha^2 \hat{C}^2}{n(1-\sigma)^2},
\end{align*}
where $\hat{C}$ is defined in Eq. \eqref{eqn:po}.
\end{lm}

\subsection{Proof of Theorem~\ref{thm:Non-Convex} }
\begin{proof}
From Assumption~\ref{assu:netw}, we have
 \begin{align}
\nonumber  &\quad\| 1^{\top}\nabla\mathbf{f}(\bar\x_k,\y^*(\bar\x_k))\|^2
\\\nonumber &=\|\sum_{i=1}^n \nabla f_i(\bar{\xx}_k,\yy_i^*(\bar\xx_k))+\sum_{i=1}^n\frac{1}{\alpha}(\xx_{i,k}-\sum_{j\in \mathcal{N}_i}\acute{w}_{ij}\xx_{j,k}) \|^2\\\nonumber &\leq
\sum_{i=1}^n\| \nabla f_i(\bar{\xx}_k,\yy_i^*(\bar\xx_k))+\frac{1}{\alpha}(\xx_{i,k}-\sum_{j\in \mathcal{N}_i}\acute{w}_{ij}\xx_{j,k}) \|^2
\\\nonumber
&= \|\nabla \mathbf{f}(\bar\x_k,\y^*(\bar\x_k))+\frac{1}{\alpha}(\mathbf{I}_{nd_1}-\acute{\W})\x_k\|^2
\\\nonumber &\leq 2\|\nabla \mathbf{f}(\x_k,\y^*(\x_k))+\frac{1}{\alpha}(\mathbf{I}_{nd_1}-\acute{\W})\x_k\|^2
\\\nonumber&+2\|\nabla \mathbf{f}(\bar\x_k,\y^*(\bar\x_k))-\nabla \mathbf{f}(\x_k,\y^*(\x_k))\|^2,
  \end{align}
  where the last inequality is by $\|a+b\|^2\leq 2\|a\|^2+2\|b\|^2$ for all $a,b\in \mathbb{R}^d$.

  From Lemma 2.2--c in \cite{ghadimi2018approximation}, $\nabla \mathbf{f}$ is Lipschitz continuous with constant
  \begin{align*}
    L_{f}&:= \frac{(\tilde{L}_{f_{\yy}}+ C_f)\cdot C_{g_{\xx\yy}}}{ \mu_g}+ L_{f_{\xx}}
\\&+  C_{f_{\yy}}\left(\frac{ \tilde{L}_{g_{\xx\yy}}C_{f_{\yy}}}{\mu_g}
+\frac{  C_{g_{\xx\yy}}\tilde{L}_{g_{\yy\yy}}}{\mu_g^2}\right),
  \end{align*}
where
 \begin{equation*}
    C_{f}:= L_{f_{\xx}}+\frac{ C_{g_{\xx\yy}}L_{f_{\yy}}}{ \mu_g}
+  C_{f_{\yy}}\left(\frac{ L_{g_{\xx\yy}}}{ \mu_g}+\frac{  C_{g_{\xx\yy}}L_{g_{\yy\yy}}}{\mu_g^2}\right).
  \end{equation*}
  Then, from \eqref{eqn:approximate:prob} and $L_{f}$-Lipschitz continuity of $\nabla \mathbf{f}$, we have
 \begin{align}
\nonumber  &\quad\frac{1}{K}\sum_{k=0}^{K-1}\|\frac{1}{n} 1^{\top}\nabla\mathbf{f}(\bar\x_k,\y^*(\bar\x_k))\|^2
\\\nonumber&\leq \frac{2}{K}\sum_{k=0}^{K-1}\|\frac{1}{n} 1^{\top}\nabla \F(\x_k,\y^*(\x_k))\|^2
\\\nonumber&+\frac{2L_{f}^2}{K}\sum_{k=0}^{K-1}\|\frac{1}{n} 1^{\top}(\x_k-\bar\x_k)\|^2
\\\nonumber &\leq \left(\frac{4L_{f}^2\alpha^2}{(1-\sigma)^2}+2\right) \frac{1}{K}\sum_{k=0}^{K-1} \|\frac{1}{n} 1^{\top}\nabla \mathbf{F}(\x_k,{\y}^*(\x_k))\|^2
\\&+\frac{4L_{f}^2\alpha^2 \hat{C}^2}{n(1-\sigma)^2},
\end{align}
where the second inequality follows from Lemma \ref{vbb}.

Hence,
from Lemma \ref{thm33}, we have
\begin{align*}
&\quad\frac{1}{K}\sum_{k=0}^{K-1}\|\frac{1}{n}1^{\top}\nabla\mathbf{f}(\bar\x_k,\y^*(\bar\x_k))\|^2
\\&\leq \jmath_1 \frac{16}{nK\alpha} \left(D_{F}
+\jmath_2\right)+\frac{4L_{f}^2\alpha^2 \hat{C}^2}{n(1-\sigma)^2}
\\&+\jmath_1 2C^2\frac{1}{n}\left(\frac{4\hat{L}_g\beta}{(1-\sigma)}+\frac{4\hat{L}_g\beta^{1/2}}{(1-\sigma)^{1/2}\mu_g^{1/2}}\right)
,
\end{align*}
where $\hat{C}$ is defined in \eqref{eqn:om}, and
\begin{align}\label{eqn:conca:j}
\nonumber   \jmath_1&:=\frac{4L_{f}^2\alpha^2}{(1-\sigma)^2}+2,\qquad
   \\ \jmath_2&:=\frac{5}{64L_{F}}\left(2+\frac{3\varrho^2}{16 L_F^2}
+ P_0\right).
\end{align}

\end{proof}
\subsection{Proof of Corollary~ \ref{cor:nonconvex}}
Substituting $\alpha = \mathcal{O}(1/\sqrt{K})$ and  $\beta = \mathcal{O}((1-\sigma)/(b+K))$ for some $b \geq (1-\sigma)/\bar{\beta}$ into the following inequality,
\begin{align*}
&\quad\frac{1}{K}\sum_{k=0}^{K-1}\|\frac{1}{n}1^{\top}\nabla\mathbf{f}(\bar\x_k,\y^*(\bar\x_k))\|^2
\\&\leq  \frac{\jmath_116}{nK\alpha} \left(D_{F}
+\jmath_2\right)+\frac{4L_{f}^2\alpha^2 \hat{C}^2}{n(1-\sigma)^2}
\\&+\frac{\jmath_1 2C^2}{n}\left(\frac{4\hat{L}_g\beta}{(1-\sigma)}+\frac{4\hat{L}_g\beta^{1/2}}{(1-\sigma)^{1/2}\mu_g^{1/2}}\right)
\\&=\mathcal{O}\left((n^{-2}\epsilon^{-2})+(n^{-1}\epsilon^{-1}/(1-\sigma)^2)\right)
,
\end{align*}
gives the desired result.




\section{Proofs of Auxiliary Lemmas}\label{sec:aux:lem}

\subsection{Proof of Lemma~\ref{lem: equivalent}}
\begin{proof}
Let $(\dot{\xx}^*, \dot{\yy}^*(\dot{\xx}^*))$ and $\{(\xx_i^*,\yy_i^*(\xx_i^*))\}_{i=1}^n$ denote the  minmizers of \eqref{eqn:main:dblo} and \eqref{eqn:obj:cbo}, respectively.
The constraints $\xx_i=\xx_j$ and $\yy_i=\yy_j$ enforce that $\yy_i^*(\xx_i)=\yy_i^*(\xx_j)=\yy_j^*(\xx_j)$ for all $i \in \mathcal{V}$ and $j\in \mathcal{N}_i$.
Since the network is connected, these conditions imply that
two sets of variables $\{\xx_1,\ldots,\xx_n \}$ and $\{\yy_1^*(\xx_1),\ldots,\yy_n^*(\xx_n) \}$ are feasible for Problem \eqref{eqn:obj:cbo} if and only if $\xx_1=\cdots=\xx_n$ and $\yy_1^*(\xx_1)=\cdots=\yy_n^*(\xx_n)$. Therefore,
Problems \eqref{eqn:main:dblo} and \eqref{eqn:obj:cbo} are equivalent in the sense that for all $i$ and $j$ the minmizers of \eqref{eqn:obj:cbo} satisfy $\xx_i^*=\dot{\xx}^*$ and $\yy_i^*(\xx_i^*)=\dot{\yy}^*(\dot{\xx}^*)$ for all $i \in \mathcal{V}$.

We now prove the second part.
Using Assumption \ref{assu:netw:item3}, we have
 $\textnormal{null}\{\mr {I}_{n}-\mr W\}=\textnormal{span}\{1_n\}$. Then, for the extended weight matrices $\W:=\mr W\otimes \mr I_{d_2} \in \mathbb{R}^{nd_2 \times nd_2}$ and $\acute{\W}:=\mr W\otimes \mr I_{d_1} \in \mathbb{R}^{nd_1 \times nd_1}$, we get  $\textnormal{null}\{\textbf{I}_{nd_1}-\acute{\W}\}=\textnormal{span }\{1_n\otimes \mr{I}_{d_1}\}$ and $\textnormal{null}\{\textbf{I}_{nd_2}-{\W}\}=\textnormal{span }\{1_n\otimes \mr{I}_{d_2}\}$. Thus, we have that
$(\textbf{I}_{nd_1}-\acute{\W}){\x}=0$ and $(\textbf{I}_{nd_2}-\W){\y}=0$ hold if and only if $\xx_1=\cdots=\xx_n$ and $\yy_1=\cdots=\yy_n$, respectively.
From Assumption \ref{assu:netw:item2},
the matrix $\textbf{I}_{nd_1}-\acute{\W}$ is positive semidefinite, which yields that its square root matrix $(\textbf{I}_{nd_1}-\acute{\W})^{1/2}$
is also positive semidefinite.
As a consequence, the bilevel optimization problem
in \eqref{eqn:obj:cbo} is equivalent to the optimization problem
\begin{subequations}\label{37ab}
\begin{align}\label{pr BLO4:}
 &\min_{{\x}\in \mathbb R^{nd_1}}  1^{\top}\mathbf{f}(\x,{\y}^*(\x))=\sum_{i=1}^n f_i\left({\xx}_i,{\yy}_i^*({\xx}_i)\right)\\\nonumber
&\,\,\,\,  \mbox{s.t.}   \ (\textbf{I}_{nd_1}-\acute{\W})^{1/2} {\x}=0,
\\ \label{pr: BLO3}
 &\qquad {\y}^*({\x}):=\argmin_{\y\in \mathbb R^{nd_2}}  1^{\top}\mathbf{g}(\x,\y)=\sum_{i=1}^n g_i\left({\xx}_i,{\yy}_i\right)\\\nonumber
 &  \qquad \qquad \qquad ~\textnormal{\text{subj.~to}~~}    (\textbf{I}_{nd_2}-\W)^{1/2} {\y}=0,
\end{align}
\end{subequations}
where ${\y}^*({\x})=[{\yy}_1^*({\xx}_1);\ldots;{\yy}_n^*({\xx}_n)]$.

Note that the minimizer of \eqref{pr BLO4:} is $\x^*=[\xx^*_1;\ldots;\xx^*_n]$, where each $\{ \xx_i^*\}$  is the solution of \eqref{eqn:obj:cbo}. Let  $\beta$ and $\alpha$ denote the penalty coefficients for the equality constraints in \eqref{pr BLO4:} and \eqref{pr: BLO3}, respectively. Then, using the penalty method, we can reformulate \eqref{pr BLO4:} and \eqref{pr: BLO3} as follows:

\begin{subequations}
\begin{align*}
&\min_{\x\in \mathbb{R}^{nd_1}}\ \mathbf{F}(\x,\check{\y}^*(\x)):=\frac{1}{2\alpha}{\x}^\top (\textbf{I}_{nd_1}-\acute{\W}){\x}
\\&\qquad\qquad\qquad\qquad\qquad+ \sum_{i=1}^n f_i\left(\xx_i,\check{\yy}_i^*(\xx_i)\right)\\
& ~~~\textnormal{\text{subj.~to}~~}   \ \check{\y}^*(\x)=\argmin_{\y\in \mathbb{R}^{nd_2}}\big\{\mathbf{G}(\x,\y):=\frac{1}{2\beta}{\y}^\top (\textbf{I}_{nd_2}-\W){\y}
\\&\qquad\qquad\qquad\qquad\qquad+\sum_{i=1}^{n}g_i(\xx_i,\yy_i)\big\}.
\end{align*}
\end{subequations}
\end{proof}

We consider the orthogonal decomposition of $\check{\y}^*(\x)$ w.r.t.
$\textnormal{null}\{\textbf{I}_{nd_2}-{\W}\}=\left\{\y\in \mathbb{R}^{nd_2 }:(\textbf{I}_{nd_2}-{\W})\y=0_{ nd_2}\right\}$ and its orthogonal complement $\textnormal{null}\{\textbf{I}_{nd_2}-{\W}\}^\bot$; i.e.,
\begin{align}\label{ortt}
\check{\y}^*(\x)=\check{\y}^*_{1}(\x)+\check{\y}^*_{2}(\x),
\end{align}
where $\check{\y}^*_{1}(\x)\in \textnormal{null}\{\textbf{I}_{nd_2}-\W\}$ and $\check{\y}^*_{2}(\x) \in \textnormal{null}\{\textbf{I}_{nd_2}-\W\}^\bot$.
\begin{lm}\label{cn}
Suppose Assumption~\ref{assu:netw} on the weight matrix $\mr W$ holds.
Further, assume  $\{g_i\}_{i=1}^n$ are strongly convex with parameter $\mu_{g} > 0$.
Then, there exists a constant $\hat{L}_g>0$, which is independent of $\beta$, such that
\begin{align}\label{177}
|\mathbf{g}(\x,\check{\y}^*(\x))-\mathbf{g}(\x,\check{\y}^*_1(\x)) |\leq \hat{L}_g \|\check{\y}^*(\x)-\check{\y}^*_1(\x)\|.
\end{align}
\end{lm}
\begin{proof}
Since $\check{\y}^*(\x)$ is the minimizer of the penalized Problem
\eqref{eqn:approximate:y}, we have
\begin{align}\label{611}
\nonumber  &\quad \beta1^{\top}\mathbf{g}(\x,\check{\y}^*(\x))+\frac{1}{2}{\check{\y}^*(\x)}^\top (\textbf{I}_{nd_2}-\W){\check{\y}^*(\x)}
\\\nonumber &\leq  \beta1^{\top}\mathbf{g}(\x,{\y}^*(\x))+\frac{1}{2}{{\y}^*(\x)}^\top (\textbf{I}_{nd_2}-\W){{\y}^*(\x)}
\\ &= \beta1^{\top}\mathbf{g}(\x,{\y}^*(\x)),
\end{align}
where we use $(\textbf{I}_{nd_2}-\W)^{1/2}{{\y}^*(\x)}=0$ due to \eqref{pr: BLO3}.
Thus, it follows that
\begin{align}\label{jhgg}
\nonumber&\quad {\check{\y}^*(\x)}^\top (\textbf{I}_{nd_2}-\W){\check{\y}^*(\x)}
\\&\leq
2 \beta\left(1^{\top}\mathbf{g}(\x,{\y}^*(\x))-1^{\top}\mathbf{g}(\x,\check{\y}^*(\x)) \right)
\leq \beta R,
\end{align}
where $R$ is a constant independent of $\beta$ such that $2\left(1^{\top}\mathbf{g}(\x,{\y}^*(\x))-1^{\top}\mathbf{g}(\x,\check{\y}^*(\x)) \right)\leq R$.
We need to verify the existence of such a
constant. Under our assumption, $\mathbf{g}(\x,{\y}^*(\x))$ is finite and independent
of $\beta$. Below we show that $\mathbf{g}(\x,\check{\y}^*(\x))$ has a lower bound that is
finite and independent of $\beta$.
Indeed, since $\mathbf{g}(\x,\cdot)$ is strongly
convex with constant $\mu_g$ for any fixed $\x\in \mathbb{R}^{nd_1 }$, we have
\begin{align}\label{fgg}
\nonumber \mathbf{g}(\x,\check{\y}^*(\x))&\geq \mathbf{g}(\x,0)+\langle \nabla \mathbf{g}(\x,0),\check{\y}^*(\x) \rangle
\\&+ \frac{\mu_g}{2}\|\check{\y}^*(\x) \|^2.
\end{align}
Observe that
\begin{align}\label{dff}
\nonumber &\quad \mathbf{g}(\x,0)+\langle \nabla \mathbf{g}(\x,0),\check{\y}^*(\x) \rangle+ \frac{\mu_g}{2}\norm{\check{\y}^*(\x) }^2
\\\nonumber &\geq \min_{\y}\left\{\mathbf{g}(\x,0)+\langle \nabla \mathbf{g}(\x,0),{\y}\rangle+ \frac{\mu_g}{2}\|{\y} \|^2\right\}
\\&=\mathbf{g}(\x,0)-\frac{1}{2\mu_g}\norm{ \nabla \mathbf{g}(\x,0) }^2.
\end{align}
Substituting \eqref{dff} into \eqref{fgg}, we have
\begin{align}\label{jhff}
\mathbf{g}(\x,\check{\y}^*(\x))\geq  \mathbf{g}(\x,0)-\frac{1}{2\mu_g}\| \nabla \mathbf{g}(\x,0) \|^2.
\end{align}
Without loss of generality, we assume that $0\in \textnormal{dom }(\mathbf{g})$ such that $\mathbf{g}(\x,0)<\infty$. Otherwise, we can replace $0$ by any
solution in $\textnormal{dom }(\mathbf{g})$ and \eqref{jhff} still holds. Thus, we can choose
\begin{align}\label{gf}
\nonumber R&=2\left(1^{\top}\mathbf{g}(\x,\y^*(\x)) -\mathbf{g}(\x,0)
\right.\\&\left.+\frac{1}{2\mu_g}\| \nabla \mathbf{g}(\x,0) \|^2\right),
\end{align}
which is finite
and independent of $\beta$.

Then, according to the orthogonal decomposition of $\check{\y}^*(\x)$ in \eqref{ortt}, we have
\begin{align}\label{khgg}
\nonumber&\quad\left({\check{\y}^*(\x)}^\top (\textbf{I}_{nd_2}-\W){\check{\y}^*(\x)}\right)^{1/2}
\\\nonumber &=\left({\check{\y}^*_2(\x)}^\top (\textbf{I}_{nd_2}-\W){\check{\y}^*_2(\x)}\right)^{1/2}
\\&\geq (1-\sigma)^{1/2}\|\check{\y}^*_2(\x) \|.
\end{align}
Substituting \eqref{jhgg} into \eqref{khgg} leads to
\begin{align}\label{ss}
 \|\check{\y}^*_2(\x)\|\leq \frac{(\beta R)^{1/2}}{(1-\sigma)^{1/2}}.
\end{align}
By \eqref{ss}, the distance between $\check{\y}^*(\x)$ and $\check{\y}^*_1(\x)$ (i.e., $\|\check{\y}^*_2(\x)\|$) is
bounded. Since $\check{\y}^*(\x)$ is optimal for the penalized problem \eqref{eqn:approximate:y},
we have
\begin{align}\label{fgbdd}
\nonumber\mathbf{g}(\x,\check{\y}^*(\x))&\leq\mathbf{g}(\x,0)-\frac{1}{2\beta}\left({\check{\y}(\x)^{*\top}} (\textbf{I}_{nd_2}-\W){\check{\y}^*(\x)}\right)^{1/2}
\\&\leq \mathbf{g}(\x,0).
\end{align}
Thus, $\check{\y}^*(\x)\in \mathcal{C}_0(\mathbf{g})$, where $\mathcal{C}_0(\mathbf{g})=\{\y:\mathbf{g}(\x,\y)\leq \mathbf{g}(\x,0) \}$ is
a sub-level set of the function $\mathbf{g}$. By the strong convexity of $\mathbf{g}$, we know that the sub-level set $\mathcal{C}_0(\mathbf{g})$ is compact \cite{boyd2004convex}. By
\eqref{fgbdd} and \eqref{ss}, we know both $\check{\y}^*(\x)$ and $\check{\y}^*_1(\x)$ belong to a compact
set that is independent of $\beta$. Note that a finite convex function
is Lipschitz on any compact set. Consequently, there exists a
constant $\hat{L}_g>0$, which is independent of $\beta$, such that \eqref{177}
holds. This completes the proof.
\end{proof}
\begin{lm}\label{cl}
Suppose Assumption~\ref{assu:netw} on the weight matrix $\mr W$ holds.
Further, assume  $\{g_i\}_{i=1}^n$ are strongly convex with parameter $\mu_{g} > 0$.
Then, there exists a constant $\hat{L}_g>0$, which is independent of $\beta$, such that the distance between the optimal solution $\check{\y}^*(\x)$ to
the penalized Problem \eqref{eqn:approximate:y} and the optimal solution ${\y}^*(\x)$ to the
constrained Problem \eqref{pr: BLO2} is bounded by
\begin{align}\label{as}
\|{\y}^*(\x)- \check{\y}^*(\x)\|\leq \frac{2\hat{L}_g\beta}{(1-\sigma)}+\frac{2\hat{L}_g\beta^{1/2}}{(1-\sigma)^{1/2}\mu_g^{1/2}}.
\end{align}
\end{lm}
\begin{proof}
According to the triangle inequality, we have $\|\check{\y}^*(\x)-{\y}^*(\x)\|\leq \|\check{\y}^*_1(\x)-{\y}^*(\x) \| + \|\check{\y}^*_2(\x) \|$. We will bound $\|\check{\y}^*_1(\x)-{\y}^*(\x) \|$ and $\|\check{\y}^*_2(\x)\|$ in the following, respectively.
Since ${\y}^*(\x)$ is the minimizer of the constrained Problem \eqref{pr: BLO3} and $(\textbf{I}_{nd_2}-{\W})^{1/2} \check{\y}^*_1(\x)=0$, we have
\begin{align}\label{jkgg}
\mathbf{g}({\x},\y^*({\x}))\leq \mathbf{g}({\x},\check{\y}^*_1({\x})).
\end{align}
Combining \eqref{611} with \eqref{jkgg} and then substituting \eqref{177} in Lemma \ref{cn}, we have
\begin{align}\label{655}
\nonumber \frac{1}{2\beta}{\check{\y}^*(\x)^{\top}} (\textbf{I}_{nd_2}-{\W}){\check{\y}^*(\x)}&\leq \mathbf{g}({\x},\check{\y}^*_1({\x}))-\mathbf{g}({\x},\check{\y}^*({\x}))
\\&\leq \hat{L}_g\|\check{\y}^*_2(\x) \|.
\end{align}
Further combining \eqref{611} with \eqref{655}, we have
\begin{align}\label{lhh}
 \|\check{\y}^*_2(\x)\|\leq \frac{2\beta \hat{L}_g}{(1-\sigma)}.
\end{align}
For $\|\check{\y}^*_1(\x)-{\y}^*(\x) \|$, with \eqref{177} in Lemma \ref{cn}, \eqref{611}, and \eqref{lhh},
we have
\begin{align}\label{nvv}
\nonumber \mathbf{g}({\x},\check{\y}^*_1({\x})) &\leq \mathbf{g}({\x},\check{\y}^*({\x}))+\hat{L}_g\|\check{\y}^*_1(\x)-\check{\y}^*(\x) \|
 \\ \nonumber &\leq \mathbf{g}({\x},\y^*({\x}))+\hat{L}_g\|\check{\y}^*_2(\x)\|
 \\&\leq \mathbf{g}(\x,\y^*(\x))+\frac{2\beta \hat{L}_g^2}{(1-\sigma)}.
\end{align}
By the strong convexity of $\mathbf{g}$, we have
\begin{align}\label{jhvv}
\nonumber \mathbf{g}({\x},\check{\y}^*_1({\x})) &\geq \mathbf{g}(\x,\y^*({\x}))
+\langle \nabla\mathbf{g}(\x,\y^*({\x})),\check{\y}^*_1(\x)-{\y}^*(\x)  \rangle
\\\nonumber&+\frac{\mu_g}{2}\| \check{\y}^*_1(\x)-{\y}^*(\x) \|^2
\\&= \mathbf{g}(\x,\y^*({\x}))+\frac{\mu_g}{2}\|\check{\y}^*_1(\x)-{\y}^*(\x) \|^2,
\end{align}
where the equality holds because  $\nabla\mathbf{g}(\x,\y^*({\x}))=0$ due to the optimality condition of \eqref{pr: BLO2}.

Combining \eqref{jhvv} with \eqref{nvv}, we get
\begin{align}\label{jg}
 \|\check{\y}^*_1(\x)-{\y}^*(\x) \|\leq \frac{2\hat{L}_g\beta^{1/2}}{(1-\sigma)^{1/2}\mu_g^{1/2}}.
\end{align}
With \eqref{lhh} and \eqref{jg}, we have
\begin{align*}
 \nonumber \|\check{\y}^*(\x)-{\y}^*(\x) \|&\leq  \|\check{\y}^*_2(\x) \|+\|\check{\y}^*_1(\x)-{\y}^*(\x) \|
 \\&\leq \frac{2\beta \hat{L}_g}{(1-\sigma)}+\frac{2\hat{L}_g\beta^{1/2}}{(1-\sigma)^{1/2}\mu_g^{1/2}}.
\end{align*}
Consequently, if the network is not too poorly connected so
that $(1-\sigma)^{1/2}>>\beta^{1/2}$ , we have
\begin{align*}
\| \check{\y}^*(\x)-{\y}^*(\x)\|\leq \mathcal{O}(\beta^{1/2}),
\end{align*}
which completes the proof.
\end{proof}

\subsection{Proof of Lemma~\ref{lem:grad}}
\begin{proof}
Given $\x\in \mathbb R^{n d_1}$, the optimality condition of the inner problem in \eqref{eqn:approximate:y} at $\check{\y}^*(\x)$ is
\begin{equation*}
 \nabla_{\y}\mathbf{G}\left(\x,\check{\y}^*(\x)\right)=(\textbf{I}_{n d_2}-\W){\check{\y}^*(\x)}+\beta \nabla_{\y} \mathbf{g}\left(\x,\check{\y}^*(\x)\right)=0,
\end{equation*}
where
\begin{align*}
\nonumber &\quad\nabla_{\y} \mathbf{g}(\x,\check{\y}^*(\x))
\\&:=[\nabla_{\yy} g_1\left(\xx_1, \check{\yy}_{1}^*(\xx_1)\right);\ldots;\nabla_{\yy} g_n\left(\xx_n, \check{\yy}_{n}^*(\xx_n)\right)]\in\mathbb{R}^{nd_2}.
\end{align*}
Then, by using the implicit differentiation w.r.t. $\x$, we have
\begin{align}\label{fff}
\nonumber &\quad(\textbf{I}_{n d_2}-\W) \nabla \check{\y}^*(\x)+\beta\nabla^2_{\x \y} \mathbf{g}\left(\x,\check{\y}^*(\x)\right)
\\&+\beta\nabla \check{\y}^*(\x) \ \nabla_{\y}^2 \mathbf{g}\left(\x,\check{\y}^*(\x)\right) =0,
\end{align}
where the matrix $\nabla_{\y}^2 \mathbf{g}(\x,\check{\y}^*(\x))\in \mathbb{R}^{nd_2\times nd_2}$ is a block diagonal matrix formed by blocks containing the Hessian of the $i$-th local function $\nabla_{\yy}^2 g_i(\xx_{i}, \check{\yy}_{i}^*(\xx_{i}))\in \mathbb{R}^{d_2 \times d_2}$.
Moreover, the matrix $\nabla^2_{\x \y} \mathbf{g} \left(\x,\check{\y}^*(\x)\right) \in \mathbb{R}^{nd_1\times nd_2}$ is a block diagonal matrix formed by blocks containing the Jacobian of the $i$-th local function,
$
\nabla^2_{ \xx \yy}g_i(\xx_{i}, \check{\yy}_{i}^*(\xx_{i}))\in \mathbb{R}^{d_1 \times d_2}.
$
\\
Rearranging the terms of \eqref{fff} yields
\begin{align*}
   &\quad \nabla \check{\y}^*(\x)
   \\&=- \beta  \nabla^2_{\x \y} \mathbf{g}\left(\x,\check{\y}^*(\x)\right) \left((\textbf{I}_{n d_2}-\W)+\beta  \nabla_{\y}^2 \mathbf{g}(\x,\check{\y}^*(\x))\right)^{-1}\\&=- \beta  \nabla^2_{\x \y} \mathbf{g}\left(\x,\check{\y}^*(\x)\right) [\textbf{H}(\x,\check{\y}^*(\x))]^{-1},
\end{align*}
where $\mathbf{H}(\x,\check{\y}^*(\x)):=(\textbf{I}_{n d_2}-\W)+\beta \nabla_{\y}^2 \mathbf{g}(\x,\check{\y}^*(\x))$.
\\
Finally, by leveraging the strong convexity of $\mathbf{g}$ as per Assumption \ref{assu:g:strongly} and the positive definiteness of the matrix $\mathbf H$ as per Lemma~\ref{eigenvalue_bounds}, we attain the desired result.

To proceed with the proof of the second part, let us, for the sake of brevity, define
\begin{align*}
  &\quad\nabla_{\x}\mathbf{f}(\x,\check{\y}^*(\x))
  \\&:=[\nabla_{\mr x} f_1(\xx_1,\check{\yy}_1^*(\xx_1));\ldots;\nabla_{\mr x} f_n(\xx_n,\check{\yy}_n^*(\xx_n))] \in\mathbb{R}^{nd_1} ,\\
   &\quad\nabla_{\y}\mathbf{f}(\x,\check{\y}^*(\x))
   \\&:=[\nabla_{\mr y} f_1(\xx_1,\check{\yy}_1^*(\xx_1));\ldots;\nabla_{\mr y} f_n(\xx_n,\check{\yy}_n^*(\xx_n))] \in\mathbb{R}^{nd_1}.
\end{align*}
By considering \eqref{eqn:approximate:prob} and applying the chain rule, we have
\begin{align*}
\nabla \mathbf{F}  \left(\x,\check{\y}^*(\x)\right)
&=\frac{1}{\alpha}(\textbf{I}_{n d_1}-\acute{\W})\x
+ \nabla_{\x}\mathbf{f}\left(\x,\check{\y}^*(\x)\right)
\\&+\nabla \check{\y}^*(\x)\nabla_{\y}\mathbf{f}\left(\x,\check{\y}^*(\x)\right).
\end{align*}
This completes the proof.
\end{proof}

\subsection{Proof of Lemma~\ref{lem:lip}}

\begin{proof}
$\textbf{Part}$ \ref{L1}:

Let $\M(\x,\y):=\nabla^2_{\x \y} \mathbf{g} \left(\x, \y\right)[\mathbf{H}(\x,\y)]^{-1}$,
where $\mathbf{H}(\x,\y):=(\textbf{I}_{n d_2}-\W)+\beta \nabla_{\y}^2 \mathbf{g}(\x,\y)$. In addition, define
  \begin{align*}
    \Delta_k & = \nabla \mathbf{\F}(\x,\y)-\nabla \mathbf{\F}(\x,\check{\y}^*(\x)),\\
   \Delta_k^1  &= \nabla_{\x} \mathbf{f}(\x,\y)-\nabla_{\x} \mathbf{f}(\x,\check{\y}^*(\x)),\\
  \Delta_k^2   &=\beta\M(\x,\y)\nabla_{\y} \mathbf{f}(\x,\y)
  -\beta\M(\x,\check{\y}^*(\x))\nabla_{\y} \mathbf{f}(\x,\check{\y}^*(\x)),\\
  \Delta_k^3 &=\beta\M(\x,\y)\left\{ \nabla_{\y} \mathbf{f}(\x,\y)-\nabla_{\y} \mathbf{f}(\x,\check{\y}^*(\x))\right\},\\
  \Delta_k^4 &=\beta\left\{\M(\x,\y)-\M(\x,\check{\y}^*(\x))\right\}\nabla_{\y} \mathbf{f}(\x,\check{\y}^*(\x)),\\
  \Delta_k^5 &=\beta\left\{\nabla^2_{\x \y} \mathbf{g} \left(\x, \y\right)-\nabla^2_{\x \y} \mathbf{g} \left(\x, \check{\y}^*(\x)\right)\right \} [ \mathbf{H} \left(\x, \y\right)]^{-1},\\
  \Delta_k^6 &=\beta\nabla^2_{\x \y} \mathbf{g} \left(\x, \check{\y}^*(\x)\right)
\left \{ [ \mathbf{H} \left(\x, \y\right)]^{-1}-[ \mathbf{H} \left(\x, \check{\y}^*(\x)\right)]^{-1}\right \}.\\
  \end{align*}
Then, it follows from \eqref{eqn:grad:F} and \eqref{eqn:grad:tildF} that
  \begin{align}\label{jh}
 \nonumber  \Delta_k&= \Delta_k^1-\Delta_k^2  =\Delta_k^1-\Delta_k^3-\Delta_k^4
   \\&= \Delta_k^1-\Delta_k^3-(\Delta_k^5+\Delta_k^6)\nabla_{\y} \mathbf{f}(\x,\check{\y}^*(\x)).
  \end{align}
$\bullet$ \textbf{Bounding $\| \Delta_k^1 \|$}: Note that
  \begin{align}\label{mk}
 \nonumber   \| \Delta_k^1 \| & =  \|\nabla_{\x}\mathbf{f}(\x, \y)-\nabla_{\x}\mathbf{f}(\x,\check{\y}^*(\x)) \|\\
 \nonumber   & =  \sum_{i=1}^{n}\|\nabla_{\xx} f_i(\xx_i, \yy_i)-\nabla_{\xx} f_i(\xx_i,\check{\yy}_i^*(\xx_i)) \|\\
\nonumber    &\leq  L_{f_{\xx}}\sum_{i=1}^{n}\| \check{\yy}_i^*(\xx_i)- \yy_i\|\\
    &= L_{f_{\xx}}\|\check{\y}^*(\x)- \y \|,
  \end{align}
  where the inequality is due to Assumption \ref{assu:f:L}.
  \\
$\bullet$ \textbf{Bounding  $\|\Delta_k^3 \|$}: From Assumption \ref{assu:g:strongly} and $\beta<1$, observe that
  \begin{align*}
    \mathbf{H}(\x,\y)&= (\textbf{I}_{n d_2}-\W)+\beta \nabla_{\y}^2 \mathbf{g}(\x,\y)
    \\&\succeq \hat{\lambda}_{\min}(\textbf{I}_{nd_2}-\W)+\beta \mu_g
    \\&\geq \beta\hat{\lambda}_{\min}(\textbf{I}_{nd_2}-\W)+\beta \mu_g,
  \end{align*}
Thus, by setting $\mu_{\mathbf{G}}:=\hat{\lambda}_{\min}(\textbf{I}_{nd_2}-\W)+ \mu_g$, we get
    \begin{align}\label{llm}
  \|[\mathbf{H}(\x,\y)]^{-1} \| &\leq \frac{1}{\beta\mu_{\mathbf{G}}}.
  \end{align}
On the other hand, since matrix $\nabla^2_{\x \y} \mathbf{g} \left(\x, \y \right)$ is block diagonal and the eigenvalues of each diagonal block
$\nabla^2_{\xx\yy} g_i \left(\xx_i, \yy_i\right)$ are bounded by constant $0<C_{g_{\xx\yy}}<\infty$ due to Assumption \ref{assu:g:Cgxy}, we get
\begin{align}\label{samm}
\nabla^2_{\x \y} \mathbf{g} \left(\x, \y \right)\preceq C_{g_{\xx\yy}} \mr I.
\end{align}
Hence, from \eqref{llm} and \eqref{samm}, we obtain
  \begin{subequations}
  \begin{align}\label{mka}
  \nonumber  \|\M(\x,\y)\| & =\|\nabla^2_{\x \y} \mathbf{g} \left(\x, \y\right)[\mathbf{H}(\x,\y)]^{-1}\|\\
 \nonumber   &\leq \|\nabla^2_{\x \y} \mathbf{g} \left(\x, \y\right)\|\ \|[\mathbf{H}(\x,\y)]^{-1}\|\\
    &\leq  \frac{C_{g_{\xx\yy}}}{\beta\mu_{\mathbf{G}}}.
  \end{align}
From Assumption \ref{assu:f:L}, we also have
  \begin{align}\label{mkg}
 \nonumber  &\quad\| \nabla_{\y} \mathbf{f}(\x,\y)-\nabla_{\y} \mathbf{f}(\x,\check{\y}^*(\x)) \|
   \\\nonumber &= \sum_{i=1}^{n} \|\nabla_{\yy}f_i(\xx_i,\yy_i) - \nabla_{\yy}f_i(\xx_i,\check{\yy}_i^*({\xx}_i))\|\\\nonumber
   &\leq  L_{f_{\yy}}\sum_{i=1}^{n} \| \check{\yy}_i^*(\xx_i)-\yy_i\|\\
   &= L_{f_{\yy}}\|\check{\y}^*(\x)-\y \|.
  \end{align}
    \end{subequations}
Combining \eqref{mka} with \eqref{mkg}, we obtain
    \begin{align}\label{jkl}
\nonumber    \|\Delta_k^3 \| & =\beta\|\M(\x,\y)\big\{ \nabla_{\y} \mathbf{f}(\x,\y)-\nabla_{\y} \mathbf{f}(\x,\check{\y}^*(\x))\big\} \|\\\nonumber
    &\leq \beta\|\M(\x,{\y})\|\ \| \nabla_{\y} \mathbf{f}(\x,\y)-\nabla_{\y} \mathbf{f}(\x,\check{\y}^*(\x)) \|\\
    &\leq \frac{ C_{g_{\xx\yy}}L_{f_{\yy}}\|\check{\y}^*(\x)-\y \|}{\mu_{\mathbf{G}}}.
  \end{align}
  \\
\textbf{$\bullet$ Bounding $\| \Delta_k^5\|$}: Note that the Jacobian matrix $\nabla^2_{\x \y} \mathbf{g} (\x,\y)$ can be expressed as:
\begin{small}
\begin{eqnarray*}
	\nabla^2_{\x \y} \mathbf{g} (\x,\y) &=&
	\left(
     \begin{array}{ccc}
		\frac{\partial \nabla \mathbf{g} (\xx_1,\y)}{\partial \xx_1 }  &  \cdots & \frac{\partial \nabla \mathbf{g} (\xx_n,\y)}{\partial \xx_n } \\
		\vdots & \ddots & \vdots\\
		\frac{\partial \nabla \mathbf{g} (\xx_1,\y)}{\partial \xx_1 } & \cdots & \frac{\partial \nabla \mathbf{g} (\xx_n,\y)}{\partial \xx_n }
	\end{array}
   \right)
   \\&=&
   \left(
     \begin{array}{ccc}
		\frac{\partial \nabla g_1 (\xx_1,\yy_1)}{\partial \xx_1 \partial \yy_1}  &  \cdots & \frac{\partial \nabla g_1 (\xx_n,\yy_1)}{\partial \xx_n \partial \yy_1} \\
		\vdots & \ddots & \vdots\\
		\frac{\partial \nabla g_n (\xx_1,\yy_n)}{\partial \xx_1 \partial \yy_n} & \cdots & \frac{\partial \nabla g_n (\xx_n,\yy_n)}{\partial \xx_n \partial \yy_n}
	\end{array}
   \right)
   \\&=&
	\left(
     \begin{array}{c}
		\nabla^2_{\x  \yy_1}  g_1 (\x,\yy_1) \\
		\vdots  \\
		\nabla^2_{\x \yy_n} g_n (\x, \yy_n)
	\end{array}
   \right)
   .
\end{eqnarray*}
   \end{small}

Thus, we have
 \begin{align}\label{dfs}
 \nonumber &\quad\|\nabla^2_{\x \y} \mathbf{g} \left(\x, \y\right)-\nabla^2_{\x \y} \mathbf{g} \left(\x, \check{\y}^*(\x)\right) \|
\\\nonumber &=\sum_{i=1}^{n}\|\nabla^2_{\x  \yy}  g_i \left(\x, \yy_i \right)-\nabla^2_{\x \yy} g_i \left(\x, \check{\yy}_i^*(\x)\right) \|\\ \nonumber
 & =\sum_{i=1}^{n}\sum_{i=1}^{n}\|\nabla^2_{\xx  \yy} g_i \left(\xx_i, \yy_i \right)-\nabla^2_{\xx   \yy} g_i \left(\xx_i, \check{\yy}_i^*(\xx_i)\right) \|
 \\\nonumber
 &\leq L_{g_{\xx \yy}} \sum_{i=1}^{n}  \sum_{i=1}^{n} \|\check{\yy}_i^*(\xx_i)-\yy_i \|\\
 &=  L_{g_{\xx\yy}}  \|\check{\y}^*(\x)-\y \| ,
 \end{align}
where the inequality follows from  Assumption \ref{assu:f:L}.

Then, from \eqref{llm} and \eqref{dfs}, we have that
     \begin{align}\label{nk}
 \nonumber   \| \Delta_k^5\| & =  \beta\| \big\{\nabla^2_{\x \y} \mathbf{g} \left(\x, \y\right)-\nabla^2_{\x \y} \mathbf{g} \left(\x, \check{\y}^*(\x)\right)\big \} [ \mathbf{H} \left(\x, \y\right)]^{-1}\|\\\nonumber
    &\leq   \beta \|\nabla^2_{\x \y} \mathbf{g} \left(\x, \y\right)-\nabla^2_{\x \y} \mathbf{g} \left(\x, \check{\y}^*(\x)\right) \| \ \|[ \mathbf{H} \left(\x, \y\right)]^{-1}\|\\
    &\leq   \frac{  L_{g_{\xx\yy}}\|\check{\y}^*(\x)- \y\|}{\mu_{\mathbf{G}}}.
  \end{align}
\textbf{$\bullet$ Bounding $\|\Delta_k^6 \|$}: It follows from \eqref{sam} that
    \begin{align}\label{jkp}
\nonumber    \|\Delta_k^6 \|& \leq   \beta \|\nabla^2_{\x \y} \mathbf{g} \left(\x, \check{\y}^*(\x)\right)\|
\\\nonumber & \qquad \quad
\cdot\| [ \mathbf{H} \left(\x, \y\right)]^{-1}-[ \mathbf{H} \left(\x, \check{\y}^*(\x)\right)]^{-1}\|\\\nonumber
&\leq   \beta  C_{g_{\xx\yy}}
\| [ \mathbf{H} \left(\x, \y\right)]^{-1}-[ \mathbf{H} \left(\x, \check{\y}^*(\x)\right)]^{-1}\|\\\nonumber
&\leq  \beta  C_{g_{\xx\yy}}
\| [ \mathbf{H} \left(\x, \check{\y}^*(\x)\right)]^{-1}\| \ \| [ \mathbf{H} \left(\x, \y\right)]^{-1}\|
\\&\qquad \quad\cdot\| \mathbf{H} \left(\x, \check{\y}^*(\x)\right)- \mathbf{H} \left(\x, \y\right)\| ,
  \end{align}
    where the third inequality follows from the fact that
  \begin{align*}
    \|A_2^{-1}-A_1^{-1} \| & =\|A_1^{-1}(A_1-A_2)A_2^{-1} \|
    \\&\leq \|A_1^{-1} \| \|A_2^{-1} \| \|A_1-A_2 \|,
  \end{align*}
  for any invertible matrices $A_1$ and $A_2$.

  Thus, from \eqref{llm} and \eqref{jkp}, we get
  \begin{align}\label{wr}
 \nonumber   \|\Delta_k^6 \| & \leq   \frac{  \beta  C_{g_{\xx\yy}}\| \mathbf{H} \left(\x, \check{\y}^*(\x)\right)- \mathbf{H} \left(\x, \y\right)\|}{\beta^2\mu_{\mathbf{G}}^2} \\
    &=  \frac{  C_{g_{\xx\yy}}\| \nabla_{\y}^2 \mathbf{g}(\x, \check{\y}^*(\x))- \nabla_{\y}^2 \mathbf{g}(\x,\y)\|}{\mu_{\mathbf{G}}^2}
    ,
  \end{align}
  where the equality is by the fact that the matrix $\textbf{I}_{n d_2}-\W$ does not depend on the argument $(\x, \check{\y}^*(\x))$ of $\nabla_{\y}^2 \mathbf{g}(\x, \check{\y}^*(\x))$.

  Now, for any vector $\uu:=[\mathrm{u}_1;\ldots ;\mathrm{u}_n] \in \mathbb{R}^{nd_2}$ with
  $\mathrm{u}_i \in \mathbb{R}^{d_2}$, we get
  \begin{align}\label{njm}
\nonumber  &\quad\| \nabla_{\y}^2 \mathbf{g}(\x, \check{\y}^*(\x))- \nabla_{\y}^2 \mathbf{g}(\x,\y)\|
\\\nonumber &
  =\sqrt{\max_{\uu}\frac{\uu^{\top}[\nabla_{\y}^2 \mathbf{g}(\x, \check{\y}^*(\x))- \nabla_{\y}^2 \mathbf{g}(\x,\y)]^2\uu}{\|\uu\|^2}}\\ &=\sqrt{\max_{\uu}\frac{\sum_{i=1}^{n}\mathrm{u}_i^{\top}[\nabla_{\yy}^2 g_i(\xx_i, \check{\yy}_i^*(\xx_i))- \nabla_{\yy}^2 g_i(\xx_i,\yy_i)]^2\mathrm{u}_i}{\|\uu\|^2}}.
  \end{align}
  By using the Cauchy-Schwarz inequality,
  each summand in \eqref{njm} can be upper bounded as
\begin{align}\label{mkz}
\nonumber  &\quad\mathrm{u}_i^{\top}[\nabla_{\yy}^2 g_i(\xx_i, \check{\yy}_i^*(\xx_i))- \nabla_{\yy}^2 g_i(\xx_i,\yy_i)]^2\mathrm{u}_i
\\\nonumber& \leq \| \nabla_{\yy}^2 g_i(\xx_i, \check{\yy}_i^*(\xx_i))- \nabla_{\yy}^2 g_i(\xx_i,\yy_i)\|^2\ \| \mathrm{u}_i\|^2\\
  &\leq L^2_{g_{\yy\yy}}  \|\check{\yy}_i^*(\xx_i)-\yy_i \|^2 \ \| \mathrm{u}_i\|^2,
\end{align}
where the second inequality holds because Assumption \ref{assu:f:L}.

We next combine relations \eqref{njm} and \eqref{mkz} to obtain
\begin{align}\label{df}
 \nonumber &\quad\| \nabla_{\y}^2 \mathbf{g}(\x, \check{\y}^*(\x))- \nabla_{\y}^2 \mathbf{g}(\x,\y)\|
 \\\nonumber&
  \leq \sqrt{\max_{\uu}\frac{\sum_{i=1}^{n}L^2_{g_{\yy\yy}}  \|\check{\yy}_i^*(\xx_i)-\yy_i \|^2 \ \| \mathrm{u}_i\|^2}{\sum_{i=1}^{n}\|\mathrm{u}_i\|^2}}\\\nonumber
  &\leq L_{g_{\yy\yy}}\sqrt{\sum_{i=1}^{n}\|\check{\yy}_i^*(\xx_i)-\yy_i \|^2}\\
  &=  L_{g_{\yy\yy}}\sqrt{\|\check{\y}^*(\x)-\y\|^2},
\end{align}
where the second inequality holds since dividing both sides of $\sum_{i=1}^{n}a_i^2 b_i^2\leq (\sum_{i=1}^{n}a_i^2 )(\sum_{i=1}^{n}b_i^2 )$ by $\sum_{i=1}^{n}b_i^2$ with $a_i=\|\check{\yy}_i^*(\xx_i)-\yy_i \|$ and $b_i= \|\mathrm{u}_i \|$ yields
\begin{align*}
 \frac{ \sum_{i=1}^{n}\|\check{\yy}_i^*(\xx_i)-\yy_i \|^2\|\mathrm{u}_i\|^2}{\sum_{i=1}^{n}\|\mathrm{u}_i\|^2}\leq \sum_{i=1}^{n}\|\check{\yy}_i^*(\xx_i)-\yy_i \|^2.
\end{align*}

Consequently, plugging \eqref{df} into \eqref{wr} yields
\begin{align}\label{dffa}
   \|\Delta_k^6 \|
    \leq \frac{  C_{g_{\xx\yy}}L_{g_{\yy\yy}}}{\mu_{\mathbf{G}}^2}\|\check{\y}^*(\x)-\y \|.
  \end{align}
\\
\textbf{Bounding $\| \nabla_{\y} \mathbf{f}\|$}: Using Assumption \ref{assu:f:grad}, we have
  \begin{align}\label{njks}
    \|\nabla_{\y}\mathbf{f}(\x,\check{\y}^*(\x))\|
    = \sum_{i=1}^{n} \| \nabla_{\yy} f_i(\xx_i,\check{\yy}_i^*(\xx_i))\|\leq   C_{f_{\yy}}.
  \end{align}
Therefore, by substituting \eqref{mk}, \eqref{jkl}, \eqref{nk}, \eqref{dffa} and \eqref{njks} in \eqref{jh} and employing the Cauchy-Schwarz inequality,
we obtain
 \begin{equation} \label{eqn:ff}
    \left\|
\tilde{\nabla} \mathbf{F}(\x,\y)-
\nabla \mathbf{F}(\x,\check{\y}^*(\x)) \right\|\le C\|\check{\y}^*(\x)- \y\|,
\end{equation}
where
\begin{align}\label{eqn:cf}
\nonumber  C &:= L_{f_{\xx}}+\frac{ C_{g_{\xx\yy}}L_{f_{\yy}}}{\mu_{\mathbf{G}}}
\\&+  C_{f_{\yy}}\left(\frac{ L_{g_{\xx\yy}}}{\mu_{\mathbf{G}}}+\frac{  C_{g_{\xx\yy}}L_{g_{\yy\yy}}}{\mu_{\mathbf{G}}^2}\right).
\end{align}
This completes the proof of \ref{L1}.

$\textbf{Part}$ \ref{L2}:

By utilizing Eq. \eqref{eqn:grad:y}, we have
\begin{align}\label{yyy}
\nonumber\|\nabla \check{\y}^*(\x) \|&=\|\beta \nabla^2_{\x \y} \mathbf{g}\left(\x,\check{\y}^*(\x)\right) [\textbf{H}(\x,\check{\y}^*(\x))]^{-1} \|\\\nonumber
&\leq \beta \| \nabla^2_{\x \y} \mathbf{g}\left(\x,\check{\y}^*(\x)\right) \| \ \| [\textbf{H}(\x,\check{\y}^*(\x))]^{-1} \|\\
&\leq \frac{  C_{g_{\xx\yy}}}{\mu_{\mathbf{G}}},
\end{align}
where the second inequality is due to \eqref{llm} and \eqref{samm}.

$\textbf{Part}$ \ref{L3}:

From \eqref{eqn:grad:tildF}, we have
\begin{align}\label{km}
\nonumber  &\quad \|\nabla \mathbf{F}(\x_2,\check{\y}^*(\x_2))-\nabla \mathbf{F}(\x_1,\check{\y}^*(\x_1))\|
\\\nonumber & \leq \underbrace{\| \nabla \mathbf{F}(\x_2,\check{\y}^*(\x_2))-\tilde{\nabla} \mathbf{F}(\x_2,\check{\y}^*(\x_1))\|}_{=:I_1}
\\ &+\underbrace{\|\tilde{\nabla }\mathbf{F}(\x_2,\check{\y}^*(\x_1))-\nabla \mathbf{F}(\x_1,\check{\y}^*(\x_1)) \|}_{=:I_2}.
\end{align}
We then study each terms separately. Using \eqref{eqn:ff}, we have
\begin{align}\label{eer}
 \nonumber I_1
  & \leq  C\|\check{\y}^*(\x_2)-\check{\y}^*(\x_1)\|
  \\&\leq  \frac{C  C_{g_{\xx\yy}}\|\x_2-\x_1 \|}{\mu_{\mathbf{G}}} ,
\end{align}
where the second inequality comes from \ref{L2}.

Further, similar to the proof of \eqref{eqn:ff} and using Assumption \ref{assu:lip}, we have
\begin{align}\label{hv}
 I_2 & \leq \vartheta \|\x_2-\x_1 \|,
\end{align}
where
\begin{align*}
 \vartheta &:= L_{f_{\xx}}+\frac{ C_{g_{\xx\yy}}\tilde{L}_{f_{\yy}}}{\mu_{\mathbf{G}}}
+  C_{f_{\yy}}\left(\frac{ \tilde{L}_{g_{\xx\yy}}C_{f_{\yy}}}{\mu_{\mathbf{G}}}
+\frac{  C_{g_{\xx\yy}}\tilde{L}_{g_{\yy\yy}}}{\mu_{\mathbf{G}}^2}\right).
\end{align*}
Putting \eqref{eer} and \eqref{hv}  into \eqref{km}, we obtain
\begin{equation}\label{eqn:gg}
  \|\nabla \mathbf{F}(\x_2,\check{\y}^*(\x_2))-\nabla \mathbf{F}(\x_1,\check{\y}^*(\x_1)) \|\leq L_{F}\|\x_2-\x_1 \|,
\end{equation}
where
\begin{align}\label{eqn:nn}
\nonumber L_{F}&:=\frac{(\tilde{L}_{f_{\yy}}+ C)\cdot C_{g_{\xx\yy}}}{\mu_{\mathbf{G}}}+ L_{f_{\xx}}
\\&+  C_{f_{\yy}}\left(\frac{ \tilde{L}_{g_{\xx\yy}}C_{f_{\yy}}}{\mu_{\mathbf{G}}}
+\frac{  C_{g_{\xx\yy}}\tilde{L}_{g_{\yy\yy}}}{\mu_{\mathbf{G}}^2}\right).
\end{align}
\end{proof}



\subsection{Proof of Lemma~\ref{lem:error}}
The following lemma establishes the convergence of the inner loop in Algorithm \ref{algo:D-BLGD}.
\begin{lm}\label{yy}
Suppose that Assumptions~\ref{assu:netw}, \ref{assu:f:L} and \ref{assu:g:strongly} hold. Let $\{\y_{k}^{t}\}_{t=0}^{M}$ denote the inner sequence defined in \eqref{qqq}. If
\begin{align}\label{gabb}
\beta\leq \bar{\beta}:=\min\left\{\frac{b_g}{\lambda_{\max}(\textbf{I}_{nd_2}-\W) L_{g}},\frac{2}{\mu_{g}+L_{g}},\frac{1}{b_g}\right\},
\end{align}
where $b_g:=\hat{\lambda}_{\min}(\textbf{I}_{nd_2}-\W)+\frac{\mu_{g}L_{g}}{\mu_{g}+L_{g}}$,
then we have
\begin{equation*}
 \|\y_{k}^{M}-\check{\y}^*(\x_k) \|\leq \left(1- \beta b_g\right)^{M/2} \|\y_{k}^0-\check{\y}^*(\x_k) \|.
\end{equation*}
Here, $\check{\y}^*(\x)$ is the minimizer of $\mathbf{G}(\x,\y)$ defined in \eqref{eqn:approximate:y}.
\end{lm}
\begin{proof}
By Eq. \eqref{qqq}, we have
\begin{align}\label{dd2}
\nonumber &\quad\|\y_{k}^{M}-\check{\y}^*(\x_k) \|^2
\\\nonumber&= \|\y_k^{M-1}-\beta \nabla_{\y} \mathbf{G}(\x_{k},\y_k^{M-1})-\check{\y}^*(\x_k) \|^2 \\\nonumber
&=\|\y_k^{M-1}-\check{\y}^* (\x_k)\|^2
\\\nonumber&-2 \beta \langle  \nabla_{\y} \mathbf{G}(\x_{k},\y_k^{M-1}),\y_k^{M-1}-\check{\y}^*(\x_k) \rangle \\\nonumber&+\beta^2\| \nabla_{\y} \mathbf{G}(\x_{k},\y_k^{M-1})\|^2\\\nonumber
&=\|\y_k^{M-1}-\check{\y}^*(\x_k) \|^2
\\\nonumber&-2 \beta\langle  \nabla_{\y} \mathbf{G}(\x_{k},\y_k^{M-1})-\nabla_{\y}\mathbf{G}\left(\x_k,\check{\y}^*(\x_k)\right),\y_k^{M-1}-\check{\y}^*(\x_k) \rangle \\&+\beta^2\| \nabla_{\y} \mathbf{G}(\x_{k},\y_k^{M-1})-\nabla_{\y}\mathbf{G}\left(\x_k,\check{\y}^*(\x_k)\right)\|^2,
\end{align}
where the last equality holds because $ \nabla_{\y}\mathbf{G}\left(\x_k,\check{\y}^*(\x_k)\right)=0$ due to the optimality of the inner problem in \eqref{eqn:obj:cbo}.
\\
In the following, we will bound the second and third terms of Eq. \eqref{dd2}.

Note that, based on Assumptions \ref{assu:f:L} and \ref{assu:g:strongly}, the objective function $\g$ is strongly convex w.r.t $\y$ with constant $\mu_{g}$ and its gradients $\nabla_{\y} \g$ are Lipschitz continuous (w.r.t $\y$) with constant $L_{g}$. Hence, from Lemma \ref{vv}, we get
\begin{align*}
\nonumber &\quad (\y_k^{M-1}-\check{\y}^*(\x_k))^{\top}(\nabla_{\y} \g(\x_{k},\y_k^{M-1})-\nabla_{\y} \g(\x_{k},\check{\y}^*(\x_k)))
 \\ & \geq
 \frac{1}{\mu_{g}+L_{g}}\|\nabla_{\y} \g(\x_{k},\y_k^{M-1})-\nabla_{\y} \g(\x_{k},\check{\y}^*(\x_k)) \|^2\\\nonumber&+\frac{\mu_{g}L_{g}}{\mu_{g}+L_{g}}\|\y_k^{M-1}-\check{\y}^*(\x_k) \|^2.
\end{align*}
The above inequality together with \eqref{eqn:approximate:y} gives
\begin{subequations}\begin{align}\label{vb}
\nonumber &\quad \langle  \nabla_{\y} \mathbf{G}(\x_{k},\y_k^{M-1})-\nabla_{\y}\mathbf{G}\left(\x_k,\check{\y}^*(\x_k)\right),\y_k^{M-1}-\check{\y}^*(\x_k) \rangle
  \\\nonumber &= \langle (\textbf{I}_{nd_2}-\W)({\y}_k^{M-1}-\check{\y}^*(\x_k)),\y_k^{M-1}-\check{\y}^*(\x_k) \rangle
 \\\nonumber& +
 \beta \langle  \nabla_{\y} \mathbf{g}(\x_{k},\y_k^{M-1})-\nabla_{\y} \mathbf{g}(\x_{k},\check{\y}^*(\x_k)),\y_k^{M-1}-\check{\y}^*(\x_k) \rangle\\\nonumber
 &\geq \hat{\lambda}_{\min}(\textbf{I}_{nd_2}-\W)\|\y_k^{M-1}-\check{\y}^*(\x_k) \|^2\\\nonumber
 &+\frac{\beta^2}{\mu_{g}+L_{g}}\|\nabla_{\y} \g(\x_{k},\y_k^{M-1})-\nabla_{\y} \g(\x_{k},\check{\y}^*(\x_k)) \|^2\\&+\frac{\mu_{g}L_{g}}{\mu_{g}+L_{g}}\|\y_k^{M-1}-\check{\y}^*(\x_k) \|^2.
\end{align}
Next, we bound the last term in RHS of \eqref{dd}. Note that
\begin{align}\label{vbbn}
\nonumber  &\quad \| \nabla_{\y} \mathbf{G}(\x_{k},\y_k^{M-1})-\nabla_{\y}\mathbf{G}\left(\x_k,\check{\y}^*(\x_k)\right)\|^2
\\\nonumber & =
 \|(\textbf{I}_{nd_2}-\W) \big(\y_k^{M-1}-\check{\y}^*(\x_k)\big)
 \\\nonumber&+\beta \big( \nabla_{\y} \g(\x_{k},\y_k^{M-1})-\nabla_{\y} \g(\x_{k},\check{\y}^*(\x_k))\big) \|^2\\\nonumber
 &= \|(\textbf{I}_{nd_2}-\W) \big(\y_k^{M-1}-\check{\y}^*(\x_k)\big) \|^2
 \\\nonumber&+\beta ^2\| \nabla_{\y} \g(\x_{k},\y_k^{M-1})-\nabla_{\y} \g(\x_{k},\check{\y}^*(\x_k)) \|^2
 \\&+2\beta J.
\end{align}
\end{subequations}
where $J:=\langle (\textbf{I}_{nd_2}-\W) \big(\y_k^{M-1}-\check{\y}^*(\x_k)\big) ,\nabla_{\y} \g(\x_{k},\y_k^{M-1})-\nabla_{\y} \g(\x_{k},\check{\y}^*(\x_k)) \rangle$.

Now, using Assumption \ref{assu:f:L}, we get
\begin{align*}
 \nonumber  J &\leq \|(\textbf{I}_{nd_2}-\W) \big(\y_k^{M-1}-\check{\y}^*(\x_k)\big) \| \\&\qquad \cdot\|\nabla_{\y} \g(\x_{k},\y_k^{M-1})-\nabla_{\y} \g(\x_{k},\check{\y}^*(\x_k))  \|\\\nonumber
 &=\|(\textbf{I}_{nd_2}-\W) \big(\y_k^{M-1}-\check{\y}^*(\x_k)\big) \| \ \\&\qquad \cdot\sum_{i=1}^{n}\|\nabla_{\yy} g_i(\xx_{i,k},\yy_{i,k}^{M-1})-\nabla_{\yy} g_i(\xx_{i,k},\check{\yy}_i^*(\xx_{i,k}))  \|\\\nonumber
  & \leq \lambda_{\max}(\textbf{I}_{nd_2}-\W)\|\y_k^{M-1}-\check{\y}^*(\x_k) \|\ \\&\qquad \cdot L_{g}\sum_{i=1}^{n}\|\yy_{i,k}^{M-1}-\check{\yy}_i^*(\xx_{i,k}) \|\\
  &=\lambda_{\max}(\textbf{I}_{nd_2}-\W)L_{g}\|\y_k^{M-1}-\check{\y}^*(\x_k) \|^2.
\end{align*}
Let $b_g:=\big(\hat{\lambda}_{\min}(\textbf{I}_{nd_2}-\W)+\frac{\mu_{g}L_{g}}{\mu_{g}+L_{g}}\big)$.
Plugging \eqref{vb} and \eqref{vbbn} into \eqref{dd} gives
\begin{align*}
\nonumber &\quad \|\y_k^{M}-\check{\y}^*(\x_k) \|^2 \\\nonumber
&\leq\left(1-2 \beta b_g
+\beta^2\lambda_{\max}(\textbf{I}_{nd_2}-\W) L_{g}\right)\|\y_k^{M-1}-\check{\y}^*(\x_k) \|^2
\\\nonumber &
 +\big(\beta^4-\frac{2 \beta^3}{\mu_{g}+L_{g}}\big)\| \nabla_{\y} \g(\x_{k},\y_k^{M-1})-\nabla_{\y} \g(\x_{k},\check{\y}^*(\x_k)) \|^2\\
 &\leq \left(1- \beta b_g\right)\
  \|\y_k^{M-1}-\check{\y}^*(\x_k) \|^2,
\end{align*}
where the second inequality uses \eqref{gabb}.

Then, we have
\begin{align*}
 \|\y_k^{M}-\check{\y}^*(\x_k) \| \leq \left(1- \beta b_g\right)^{1/2} \|\y_k^{M-1}-\check{\y}^*(\x_k) \|.
\end{align*}
This completes the proof.
\end{proof}

We are now ready to prove Lemma \ref{lem:error}.

\begin{proof}
Let
\begin{align}\label{eq:hatdelta}
   \widehat{\Delta}_k:=\widehat{\nabla} \mathbf{F}(\x_k,\y^M_k)-\nabla \mathbf{F}(\x_k,\check{\y}^*(\x_k))=\iota_k+\Delta_k,
\end{align}
 where
 \begin{align*}
   \iota_k&:=\widehat{\nabla} \mathbf{F}(\x_k,\y^M_k)-\tilde{\nabla} \mathbf{F}(\x_k,\y^M_k), \\
\Delta_k&:= \tilde{\nabla} \mathbf{F}(\x_k,\y^M_k)- \nabla \mathbf{F}(\x_k,\check{\y}^*(\x_k)).
 \end{align*}
\textbf{Bounding $\|\Delta_k\|^2$}:
From Eq. \eqref{eq:lipschitz} and Lemma \ref{yy}, we have
\begin{align}\label{mmma}
\nonumber&\quad\|\tilde{\nabla} \mathbf{F}(\x_k,\y^M_k)- \nabla \mathbf{F}(\x_k,\check{\y}^*(\x_k)) \|^2
\\\nonumber&\leq C^2\| \y_{k}^{M}-\check{\y}^*(\x_k)\|^2
\\&\leq C^2 \left(1- \beta b_g\right)^{M} \|\y_{k}^{0}-\check{\y}^*(\x_k) \|^2,
\end{align}
where $\tilde{\nabla} \mathbf{F}(\x_k,\y^M_k)$ is defined as in \eqref{eqn:grad:tildF}.
Next, we bound $\|\y_{k}^{0}-\check{\y}^*(\x_k) \|^2$. Notice that $\y_{k}^{0}=\y_{k-1}^{M}$ (as defined in Algorithm \ref{algo:D-BLGD}), which leads to
\begin{align}\label{dccz}
 \nonumber  &\quad\|\y_{k}^{0}-\check{\y}^*(\x_k) \|^2  \\\nonumber&\leq 2\|\y_{k-1}^{M}-\check{\y}^*(\x_{k-1}) \|^2+2\|\check{\y}^*(\x_{k-1})-\check{\y}^*(\x_k) \|^2
  \\ \nonumber & \leq 2\left(1- \beta b_g\right)^{M}\|\y_{k-1}^{0}-\check{\y}^*(\x_{k-1}) \|^2
 \\& +2\varrho^2 \|\x_{k-1}-\x_{k} \|^2,
\end{align}
where the second inequality uses Lemma \ref{yy} and Lemma \ref{lem:lip}--(\ref{L2}).

Choose $M$ such that
$$M\geq \log (\frac{1}{4})/\log \left(1- \beta b_g\right)=\mathcal{O}(\frac{1}{\beta}).$$
Then, \eqref{dccz} gives
\begin{align}\label{cc}
 \nonumber  &\quad \|\y_{k}^{0}-\check{\y}^*(\x_k) \|^2
 \\\nonumber &\leq \frac{1}{2}\|\y_{k-1}^{0}-\check{\y}^*(\x_{k-1}) \|^2
  +2\varrho^2 \|\x_{k-1}-\x_{k} \|^2
  \\\nonumber &\leq \big(\frac{1}{2}\big)^k \|\y_{0}-\check{\y}^*(\x_{0}) \|^2
  \\&+2\varrho^2 \sum_{j=0}^{k-1}\big(\frac{1}{2}\big)^{k-1-j}\|\x_{j}-\x_{j+1} \|^2.
\end{align}
Thus, combining \eqref{cc} and \eqref{mmma}, we get
\begin{align} \label{L11}
\nonumber  &\quad  \|\tilde{\nabla} \mathbf{F}(\x_k,\y^M_k)- \nabla \mathbf{F}(\x_k,\check{\y}^*(\x_k)) \|^2
\\\nonumber &\leq C^2 \left(1- \beta b_g\right)^{M} \\&\cdot \Big(\big(\frac{1}{2}\big)^k P_0+2\varrho^2 \sum_{j=0}^{k-1}\big(\frac{1}{2}\big)^{k-1-j}\|\x_{j}-\x_{j+1} \|^2\Big).
\end{align}
\textbf{Bounding $\|\iota_k \|^2$}:
It follows from \eqref{olhij} that
    \begin{align*}
\nonumber    \|\mathbf{H}^{-1} \| & =
    \|\textbf{D}^{-1/2}\sum_{u=0}^{\infty}(\textbf{D}^{-1/2} \textbf{B} \textbf{D}^{-1/2})^u\textbf{D}^{-1/2} \| \\
\nonumber    &\leq \|\textbf{D}^{-1/2}  \|^2\ \|\sum_{u=0}^{\infty}(\textbf{D}^{-1/2} \textbf{B} \textbf{D}^{-1/2})^u \|\\
\nonumber    &\leq \|\textbf{D}^{-1/2}  \|^2\ \sum_{u=0}^{\infty}\|\textbf{D}^{-1/2} \textbf{B} \textbf{D}^{-1/2}\|^u,
  \end{align*}
  where the second inequality is due to the triangle inequality. Then from Lemma \ref{eigenvalue_bounds}, we have
  \begin{subequations}

    \begin{align}\label{ll}
\nonumber    \|\mathbf{H}^{-1} \| &\leq \frac{1}{(2(1-\Theta)+{\beta \mu_g} )}\ \sum_{u=0}^{\infty}\|\textbf{D}^{-1/2} \textbf{B} \textbf{D}^{-1/2}\|^u\\
 \nonumber   & \leq \frac{1}{(2(1-\Theta)+{\beta \mu_g} )}\ \sum_{u=0}^{\infty}\rho^u
 \\
             &\leq \frac{1}{(2(1-\Theta)+{\beta \mu_g} )(1-\rho)}
             ,
  \end{align}
where the second inequality uses Lemma \ref{symmetric_term_bounds11}.
On the other hand, since the matrix $\nabla^2_{\x \y} \mathbf{g} \left(\x, \y \right)$ is block diagonal and the eigenvalues of each diagonal block
$\nabla^2_{\xx\yy} g_i \left(\xx_i, \yy_i\right)$ are bounded by a constant $0<C_{g_{\xx\yy}}<\infty$ due to Assumption \ref{assu:g:Cgxy}, this implies that
\begin{align}\label{sam}
\nabla^2_{\x \y} \mathbf{g} \left(\x, \y \right)\preceq C_{g_{\xx\yy}} \mr I.
\end{align}
  \end{subequations}
Now, we make use of \eqref{eqn:grad:tildF} and \eqref{eqn:d-k-U} as
\begin{align}\label{cxc}
 \nonumber   \|\iota_k \|&\leq \beta\|\nabla^2_{\x \y} \mathbf{g}\left(\x_k,\y^M_k\right) \|\,
    \|\hat{\textbf{H}}_{k,(U)}^{-1}-\textbf{H}_k^{-1} \| \, \|\nabla_{\y} \mathbf{f}(\x_k,\y^M_k) \|
    \\\nonumber &\leq \beta\|\nabla^2_{\x \y} \mathbf{g}\left(\x_k,\y^M_k\right) \|\,
    \|\textbf{I}_{nd_2}-\hat{\textbf{H}}_{k,(U)}^{-1}\textbf{H}_k \| \\&\qquad \|\textbf{H}_k^{-1} \|\,\|\nabla_{\y} \mathbf{f}(\x_k,\y^M_k) \|
\leq \eta \rho^{U+1},
\end{align}
where $\eta$ is defined as in Eq. \eqref{var} and the second inequality follows from Lemma \ref{Hessian_inverse_eigenvalue_bounds_lemma}, and Eqs.
\eqref{ll} and \eqref{sam}.

Finally, it follows from the fact that $\|a+b\|^2\leq 2\|a\|^2+2\|b\|^2$, Eqs. \eqref{cxc} and \eqref{L11} that
\begin{align*}
\nonumber \|\widehat{\Delta}_k \|^2 &\leq 2\|\iota_k \|^2+2\| \Delta_k \|^2
\\&\leq 2 \eta^2 \rho^{2(U+1)} +2C^2 \left(1- \beta b_g\right)^{M} \\&\quad\Big(\big(\frac{1}{2}\big)^k P_0+2\varrho^2 \sum_{j=0}^{k-1}\big(\frac{1}{2}\big)^{k-1-j}\|\x_{j}-\x_{j+1} \|^2\Big).
\end{align*}
This completes the proof.
\end{proof}

\subsection{Proof of Lemma~\ref{lem:up}}
First, we present the following lemma, which provides an upper bound on the deviation of each local copy at each iteration from the mean of all local copies.
\begin{lm}\label{lem:consensus}
Suppose Assumptions \ref{assu:netw} and \ref{assu:lip} hold,
and the step size $\beta$ is defined in \eqref{eqn:beta}.
Then, the iterates $\{\x_{k}\}_{k=0}^{K-1}$ generated by  \eqref{eqn:update:DAGM} satisfy
\begin{align*}
    \|\x_{k} -\bar{\x}_k\|\leq \frac{\alpha  \tilde{C}}{1-\sigma},
\end{align*}
where $\tilde{C}:=C_{f_{\xx}}+\frac{2C_{g_{\xx\yy}} \Lambda C_{f_{\yy}}}{\mu_{g}+L_{g}} $ and $\Lambda $ is defined in Lemma \ref{Hessian_inverse_eigenvalue_bounds_lemma}.
\end{lm}
\begin{proof}
From the update rule of Algorithm \ref{algo:D-BLGD}, we have
\begin{align*}
 \x_{k}&=\x_{k-1}-   \big((\textbf{I}_{n d_1}-\acute{\W}) \x_{k-1}+\alpha\nabla_{\x} \mathbf{f}(\x_{k-1},\tilde{\y}_{k-1})
 \\&-\alpha\beta
  \nabla_{\x \y}^2 \mathbf{g}(\x_{k-1},\tilde{\y}_{k-1}) \hat{\textbf{H}}_{k-1,(U)}^{-1}\p_{k-1}\big)\\
  &=\acute{\W} \x_{k-1}-  \alpha \mb z_{k-1}=-\alpha \sum_{r=0}^{k-1}\acute{\W}^{k-1-r} \mb z_r,
\end{align*}
where
\begin{align*}
 \mb z_{k-1}&:=\nabla_{\x} \mathbf{f}(\x_{k-1},\tilde{\y}_{k-1})
 \\&-\beta
  \nabla_{\x \y}^2 \mathbf{g}\left(\x_{k-1},\tilde{\y}_{k-1}\right) \hat{\textbf{H}}_{k-1,(U)}^{-1}\p_{k-1}.
\end{align*}
Note that by our definition,
\begin{align*}
 \x  =[\mr x_1;\ldots;\mr x_n],~~~
\bar \x_{k}  = (\frac{1}{n}1_n1_n^{\top}\otimes \mr I_{d_1}) \x_{k} \in \mathbb{R}^{nd_1\times 1}.
\end{align*}
This together with Assumption \ref{assu:netw} and $\xx_{i,0}=0$ give
  \begin{align}\label{dsgn}
\nonumber  \|\x_{k} -\bar{\x}_k\|
&=\|\x_{k} -(\frac{1}{n}1_n1_n^{\top}\otimes \mr I_{d_1}) \x_{k}\|
\\\nonumber&=\|-\alpha \sum_{r=0}^{k-1} ({\mr W}^{k-1-r} \otimes \mr I_{d_1})\mb z_{r}
\\\nonumber &+\alpha \sum_{r=0}^{k-1} \frac{1}{n}((1_n1_n^{\top}{\mr W}^{k-1-r})\otimes \mr I_{d_1})\mb z_r\|
\\\nonumber&=\|-\alpha \sum_{r=0}^{k-1} ({\mr W}^{k-1-r} \otimes \mr I_{d_1})\mb z_{r}
\\\nonumber &+\alpha \sum_{r=0}^{k-1} \frac{1}{n}((1_n1_n^{\top})\otimes \mr I_{d_1})\mb z_r \|
\\\nonumber&\leq \alpha \sum_{r=0}^{k-1} \|{\mr W}^{k-1-r}-\frac{1}{n}1_n1_n^{\top} \| \|\mb z_r \|
\\&\leq \alpha \sum_{r=0}^{k-1} \sigma^{k-1-r}\|\mb z_r \|,
  \end{align}
  where the last inequality uses Lemma \ref{ww}.

We proceed to bound $\| \mb z_r \|$ . To do so, we use the triangular inequality as follows:
\begin{align}\label{nkk}
\nonumber\| \mb z_r \|&\leq \| \nabla_{\x} \mathbf{f}(\x_{r},\tilde{\y}_{r})\|+\beta\|
  \nabla^2_{\x \y} \mathbf{g}\left(\x_{r},\tilde{\y}_{r}\right) \hat{\textbf{H}}_{r,(U)}^{-1}\p_{r} \|
  \\\nonumber &\leq C_{f_{\x}}+\beta C_{g_{\x\y}}\|
\hat{\textbf{H}}_{r,(U)}^{-1}\| C_{f_{\y}}
\\&\leq C_{f_{\x}}+\frac{2}{\mu_{g}+L_{g}} C_{g_{\x\y}} \Lambda C_{f_{\y}}=:\tilde{C},
\end{align}
where the second inequality uses Assumptions \ref{assu:f:grad} and \ref{assu:g:Cgxy}, and the last inequality follows from Lemma \ref{Hessian_inverse_eigenvalue_bounds_lemma}.

Next, by substituting \eqref{nkk} into \eqref{dsgn}, we get
\begin{align*}
  \|\x_{k} -\bar{\x}_k\|&\leq \alpha \sum_{r=0}^{k-1} \sigma^{k-1-r}\|\mb z_r \|
  \\&\leq \alpha \sum_{r=0}^{k-1} \sigma^{k-1-r}\tilde{C}
  \\&\leq \frac{\alpha \tilde{C}}{1-\sigma},
\end{align*}
as claimed.
\end{proof}

We are now ready to prove Lemma \ref{lem:up}.

\begin{proof}
From Assumption \ref{assu:netw:item2}, the matrix $\mr W$ has an eigenvalue decomposition of the form
$\mr W = \mr R \mr S \mr R^{\top}$, where $\mr R$ is an orthogonal matrix, and
the diagonal matrix $\mr S$ has values within the
range $(0,1]$,  containing exactly one entry with a value of $1$. Hence
\begin{subequations}
\begin{align}\label{ujg}
 \nonumber \|(\mathbf{I}_{nd_1}-\acute{\W})\x \|^2 &= \|(\mathbf{I}_{nd_1}-\acute{\W})(\x-\bar{\x}) \|^2\\\nonumber
  &=\|\big( \mr R(\mathbf{I}_{nd_1}-\mr S)\mr R^{\top}\otimes \mathbf{I}_{nd_1}\big)(\x-\bar{\x}) \|^2\\
  &\geq (1-\sigma)^2\| \x-\bar{\x}\|^2,
\end{align}
where the first equality holds because ${\mathbf 1}_{nd_1}^{\top}(\mathbf{I}_{nd_1}-\acute{\W})=0$ and the inequality holds due to Eq. \eqref{eqn:mixing}.

Moreover, we have
\begin{align}\label{eq:jp}
\nonumber 1^{\top}\mathbf{f}(\x,\y^*(\x))&=1^{\top}\mathbf{f}(\bar\x,\y^*(\bar\x))
\\\nonumber&+\big(1^{\top}\mathbf{f}(\x,\y^*(\x))-1^{\top}\mathbf{f}(\bar\x,\y^*(\bar\x)) \big)
\\&\geq 1^{\top}\mathbf{f}(\bar\x,\y^*(\bar\x))- \hat{C}  \| \x-\bar{\x}\|,
\end{align}
\end{subequations}
where the first inequality  holds since from Lemma \ref{lem:grad}, \eqref{yyy}, \eqref{njks}, and Assumption \ref{assu:f:grad}, we have
\begin{align}\label{eqn:om}
 \nonumber & \quad \|\nabla f_i(\bar \xx ,\yy_i^*(\bar \xx)) \|
 \\\nonumber &\leq  \|\nabla_{\xx}f_i(\bar \xx ,\yy_i^*(\bar \xx))\|
 + \|\nabla_{\yy} f_i(\bar \xx ,\yy_i^*(\bar \xx)) \|\\
  &\leq  C_{f_{\xx}}+C_{f_{\yy}}:=\hat{C},
\end{align}
which implies
\begin{align*}
f_i(\xx_i,\yy^*_i(\xx_i))&\geq f_i(\bar \xx,\yy_i^*(\bar \xx))+\nabla f_i(\bar \xx ,\yy_i^*(\bar \xx))(\xx_i-\bar{\mr \xx})
 \\&\geq f_i(\bar \xx,\yy_i^*(\bar \xx))-\hat{C}   |\xx_i-\bar{\mr \xx} |.
\end{align*}
By combining \eqref{ujg} and \eqref{eq:jp} and using \eqref{eqn:approximate:prob}, we conclude that
\begin{align*}
 \nonumber \F(\x,\y^*(\x))-\F^*
  &\geq  1^{\top}\mathbf{f}(\bar\x,\y^*(\bar\x)) -\F^*- \hat{C}  \| \x-\bar{\x}\|
  \\&+\frac{1}{2 \alpha} (1-\sigma)^2\| \x-\bar{\x}\|^2
  \\\nonumber &\geq  1^{\top}\mathbf{f}(\bar\x,\y^*(\bar\x)) -\F^*- \hat{C}  \| \x-\bar{\x}\|
  \\\nonumber &\geq  1^{\top}\mathbf{f}(\bar\x,\y^*(\bar\x)) -\F^* -\frac{   \tilde{C} \hat{C}\alpha}{1-\sigma}
  \\&\geq  1^{\top}\mathbf{f}(\bar\x,\y^*(\bar\x)) -1^{\top}\mathbf{f}^* -\frac{    \tilde{C}\hat{C}\alpha}{1-\sigma},
\end{align*}
where the third inequality follows from Lemma \ref{lem:consensus}.
\end{proof}
\subsection{Proof of Lemma~\ref{lem:dif:convex}}

\begin{lm}\label{lem:first:conv}
Suppose that Assumptions \ref{assu:netw} and \ref{assu:lip} hold. Let $\alpha\leq 1/L_F$. Then, for the objective function $\F$ that is defined in \eqref{eqn:reform:dblo}, we have
\begin{align*}
\nonumber   \sqrt{\F(\x_{k+1},\check{\y}^*(\x_{k+1}))-\F^*}&\leq \sqrt{\F(\x_{0},\check{\y}^*(\x_{0}))-\F^*}
\\&+\frac{1}{\sqrt{2L_F}}\sum_{r=0}^{k}\|\widehat{\Delta}_r \|,\quad \forall k\geq 0,
  \end{align*}
  where $\widehat{\Delta}_r := \widehat{\nabla} \mathbf{F}(\x_r,\y_r^M)- \nabla \mathbf{F}(\x_r,\check{\y}^*(\x_r))$ and $\mathbf{F}^*:= \mathbf{F}(\check{\x}^*,\check{\y}^*(\check{\x}^*))$.
\end{lm}
\begin{proof}
From the update $\x_{k+1}=\x_{k}-  \alpha \widehat{\nabla} \mathbf{F}(\x_k,{\y}^M_k)$ and the smoothness of $\mathbf{F}$ as stated in Lemma \ref{lem:lip}, we have
  \begin{align*}
  \nonumber  &\quad \F(\x_{k+1},\check{\y}^*(\x_{k+1}))
  \\\nonumber &\leq \F(\x_{k},\check{\y}^*(\x_{k}))+\langle \nabla \F(\x_{k},\check{\y}^*(\x_{k})),\x_{k+1}-\x_k  \rangle\\\nonumber &+\frac{L_{F}}{2}\|\x_{k+1}-\x_k \|^2
    \\\nonumber &= \F(\x_{k},\check{\y}^*(\x_{k}))-\alpha \langle \nabla \F(\x_{k},\check{\y}^*(\x_{k})),\widehat{\nabla} \mathbf{F}(\x_k,{\y}^M_k)  \rangle\\\nonumber &+\frac{L_{F}\alpha^2}{2}\|\widehat{\nabla} \mathbf{F}(\x_k,{\y}^M_k) \|^2
    \\\nonumber &=\F(\x_{k},\check{\y}^*(\x_{k}))+\alpha \langle \widehat{\Delta}_k ,\widehat{\nabla} \mathbf{F}(\x_k,{\y}^M_k)  \rangle
    \\\nonumber &+\left(\frac{L_{F}\alpha^2}{2}-\alpha\right)\|\widehat{\nabla} \mathbf{F}(\x_k,{\y}^M_k) \|^2
    \\\nonumber&\leq \F(\x_{k},\check{\y}^*(\x_{k}))+\left(\frac{L_{F}\alpha^2-\alpha}{2}\right)\|\widehat{\nabla} \mathbf{F}(\x_k,{\y}^M_k) \|^2
    \\&+\frac{\alpha}{2}\|\widehat{\Delta}_k \|^2,
  \end{align*}
  where $\widehat{\Delta}_r$ is defined in the statement of Lemma \ref{lem:first:conv}.

Then, by using $\alpha\leq 1/L_F$, we get
  \begin{align*}
\nonumber   \F(\x_{k+1},\check{\y}^*(\x_{k+1}))&\leq \F(\x_{k},\check{\y}^*(\x_{k}))+\frac{1}{2L_F}\|\widehat{\Delta}_k \|^2
\\&\leq \F(\x_{0},\check{\y}^*(\x_{0}))+\frac{1}{2L_F}\sum_{r=0}^{k}\|\widehat{\Delta}_r \|^2.
  \end{align*}
By subtracting $\mathbf{F}^*:= \mathbf{F}(\check{\x}^*,\check{\y}^*(\check{\x}^*))$ from both sides of the above
inequality and using the fact that $\sqrt{a+b}\leq \sqrt{a}+\sqrt{b}$ for $a,b \geq 0$, we obtain:
\begin{align*}
\nonumber   \sqrt{\F(\x_{k+1},\check{\y}^*(\x_{k+1}))-\F^*}&\leq \sqrt{\F(\x_{0},\check{\y}^*(\x_{0}))-\F^*}
\\&+\frac{1}{\sqrt{2L_F}}\sum_{r=0}^{k}\|\widehat{\Delta}_r \|,
  \end{align*}
  which completes the proof.
\end{proof}

We are now ready to prove Lemma \ref{lem:dif:convex}.

\begin{proof}
The proof is similar to that of \eqref{L11}. From  Lemmas \ref{lem:lip}--(\ref{L1}) and \ref{yy}, we obtain: 
\begin{align}\label{mmmb}
\nonumber&\quad\|\tilde{\nabla} \mathbf{F}(\x_k,\y^M_k)- \nabla \mathbf{F}(\x_k,\check{\y}^*(\x_k)) \|  
\\\nonumber &\leq C\| \y_{k}^{M}-\check{\y}^*(\x_k)\|
\\&\leq C \left(1- \beta b_g\right)^{M/2} \|\y_{k}^{0}-\check{\y}^*(\x_k) \|,
\end{align}
where $\tilde{\nabla} \mathbf{F}(\x_k,\y^M_k)$ is defined as in \eqref{eqn:grad:tildF}.

We now bound $\|\y_{k}^{0}-\check{\y}^*(\x_k) \|$. By the warm start strategy, i..e. setting $\y_{k}^{0}=\y_{k-1}^{M}$ in Algorithm~\ref{algo:D-BLGD}, we have
\begin{align*}
 \nonumber  &\quad \|\y_{k}^{0}-\check{\y}^*(\x_k) \| 
 \\\nonumber & \leq \|\y_{k-1}^{M}-\check{\y}^*(\x_{k-1}) \|+\|\check{\y}^*(\x_{k-1})-\check{\y}^*(\x_k) \|
  \\ \nonumber & \leq \left(1- \beta b_g\right)^{M/2}\|\y_{k-1}^{0}-\check{\y}^*(\x_{k-1}) \|
  +\varrho \|\x_{k-1}-\x_{k} \|
  \\\nonumber & = \left(1- \beta b_g\right)^{M/2}\|\y_{k-1}^{0}-\check{\y}^*(\x_{k-1}) \|
  \\ &+\varrho \alpha \|\widehat{\nabla} \mathbf{F}(\x_{k-1},\y_{k-1}^M) \|,
\end{align*}
where the second inequality follows from Lemmas \ref{yy} and \ref{lem:lip}--(\ref{L2}); and the last line uses Eq. \eqref{eqn:update:DAGM}.

Consequently, by applying the triangle inequality, we have
\begin{subequations}
\begin{align}
\nonumber  &\quad \|\y_{k}^{0}-\check{\y}^*(\x_k) \| 
\\\nonumber &\leq \left(1- \beta b_g\right)^{M/2}\|\y_{k-1}^{0}-\check{\y}^*(\x_{k-1}) \|
 \\\nonumber& +\varrho \alpha \|\nabla \mathbf{F}(\x_{k-1},\check{\y}^*(\x_{k-1})) \| 
  \\\label{eqn:lem1:convex}&+\varrho \alpha \|\tilde{\nabla} \mathbf{F}(\x_{k-1},\y_{k-1}^M)-\nabla \mathbf{F}(\x_{k-1},\check{\y}^*(\x_{k-1})) \|
  \\\label{dcc1z}& +\varrho \alpha \|\widehat{\nabla} \mathbf{F}(\x_{k-1},\y_{k-1}^M)-\tilde{\nabla} \mathbf{F}(\x_{k-1},\y_{k-1}^M) \|.
\end{align}
\end{subequations}
We proceed by separately upper bounding terms \eqref{eqn:lem1:convex} and \eqref{dcc1z}.
To this end, from \eqref{mmmb}, we have
\begin{align*}
   \eqref{eqn:lem1:convex} \leq
  \varrho \alpha C\left(1- \beta b_g\right)^{M/2} \|\y_{k-1}^{0}-\check{\y}^*(\x_{k-1}) \|.
\end{align*}
Further, using arguments similar to those in Eq. \eqref{cxc}, we  have
\begin{align*}
 \eqref{dcc1z} \leq  \varrho \alpha  \eta  \rho^{U+1}.
\end{align*}
Hence,
\begin{align*}
\nonumber &\quad \|\y_{k}^{0}-\check{\y}^*(\x_k) \| 
\\\nonumber &\leq (1+\varrho \alpha    C) \left(1- \beta b_g\right)^{M/2}\|\y_{k-1}^{0}-\check{\y}^*(\x_{k-1}) \|
 \\& +\varrho \alpha \big(\|\nabla \mathbf{F}(\x_{k-1},\check{\y}^*(\x_{k-1})) \|
  + \eta \rho^{U+1}\big).
\end{align*}
Now, by choosing
$$M\geq \frac{2\log (1/(2+2\varrho \alpha C))}{\log \left(1- \beta b_g\right)}=\mathcal{O}\left(\frac{\alpha}{\beta}\right),$$ we get
\begin{align}\label{jkv}
 \nonumber  &\quad\|\y_{k}^{0}-\check{\y}^*(\x_k) \|
\\\nonumber  &\leq \frac{1}{2}\|\y_{k-1}^{0}-\check{\y}^*(\x_{k-1}) \|\\\nonumber &+ \varrho \alpha  \big(\|\nabla \mathbf{F}(\x_{k-1},\check{\y}^*(\x_{k-1})) \|+ \eta \rho^{U+1}\big)
  \\\nonumber  &\leq \big(\frac{1}{2}\big)^{k} \|\y_{0}-\check{\y}^*(\x_{0}) \|\\\nonumber &+\varrho \alpha \sum_{j=0}^{k-1}\big(\frac{1}{2}\big)^{k-1-j}\left(\|\nabla \mathbf{F}(\x_{j},\check{\y}^*(\x_{j}))\|+ \eta \rho^{U+1}\right)
  \\\nonumber &\leq \big(\frac{1}{2}\big)^{k} \|\y_{0}-\check{\y}^*(\x_{0}) \|
  +\varrho \alpha \sum_{j=0}^{k-1}\big(\frac{1}{2}\big)^{k-1-j}\\ &\left(\sqrt{2L_F\big( \F(\x_{j},\check{\y}^*(\x_{j}))-\F^*\big)}+ \eta \rho^{U+1}\right),
\end{align}
where the last inequality follows from Lemmas \ref{lem:smooth} and \ref{lem:lip}--(\ref{L3}).

Substituting \eqref{jkv} into \eqref{mmmb} gives
\begin{align}\label{L22v}
\nonumber &\quad\|\tilde{\nabla} \mathbf{F}(\x_k,\y^M_k)- \nabla \mathbf{F}(\x_k,\check{\y}^*(\x_k)) \|
\\\nonumber&\leq C \left(1- \beta b_g\right)^{M/2} \Big( \big(\frac{1}{2}\big)^{k} \|\y_{0}-\check{\y}^*(\x_{0}) \|
  \\\nonumber&+\varrho \alpha \sum_{j=0}^{k-1}\big(\frac{1}{2}\big)^{k-1-j}
  \\&\cdot\big(\sqrt{2L_F\big( \F(\x_{j},\check{\y}^*(\x_{j}))-\F^*\big)}+ \eta \rho^{U+1}\big)\Big).
\end{align}
Note that, we have
\begin{align*}
 &\quad\|\widehat{\nabla} \mathbf{F}(\x_k,\y^M_k)-\nabla \mathbf{F}(\x_k,\check{\y}^*(\x_k))\| 
 \\&\leq \|\widehat{\nabla} \mathbf{F}(\x_k,\y^M_k)-\tilde{\nabla} \mathbf{F}(\x_k,\y^M_k)\|
 \\&+\| \tilde{\nabla} \mathbf{F}(\x_k,\y^M_k)- \nabla \mathbf{F}(\x_k,\check{\y}^*(\x_k)) \|.
\end{align*}
Thus, using \eqref{L22v} and \eqref{cxc}, we get
\begin{align}\label{eqn:fr}
\nonumber &\quad \sum_{k=0}^{K-1}\|\widehat{\nabla} \mathbf{F}(\x_k,\y^M_k)- \nabla \mathbf{F}(\x_k,\check{\y}^*(\x_k)) \|
\\\nonumber &\leq \sum_{k=0}^{K-1}\eta\rho^{U+1}
\\\nonumber&+C \left(1- \beta b_g\right)^{M/2}\sum_{k=0}^{K-1} \Big( \big(\frac{1}{2}\big)^{k} \|\y_{0}-\check{\y}^*(\x_{0}) \|
  \\\nonumber&+\varrho \alpha \sum_{j=0}^{k-1}\big(\frac{1}{2}\big)^{k-1-j}
  \\&\cdot\big(\sqrt{2L_F\big( \F(\x_{j},\check{\y}^*(\x_{j}))-\F^*\big)}+ \eta \rho^{U+1}\big)\Big).
\end{align}
Since, by our assumption, $U=\left|\left\lceil\log_{1/\rho} (\eta K) \right\rceil\right|$, we obtain:
\begin{subequations}

Since $\rho<1$ in Lemma \ref{symmetric_term_bounds11}, we have
\begin{align}\label{eqn:h}
\eta\rho^{(U+1)}\leq \eta\rho^{U}= \frac{1}{K}.
\end{align}
Then, we get
\begin{align}\label{eqn:fd}
\nonumber\sum_{k=1}^{K-1}\sum_{j=0}^{k-1}\big(\frac{1}{2}\big)^{k-1-j}\eta\rho^{U+1}
&\leq  \sum_{k=0}^{K-1}\frac{1}{2^{k}}\sum_{k=0}^{K-1}\eta \rho^{U+1}
\\& \leq 2\sum_{k=0}^{K-1} \eta\rho^{U+1}\leq 2.
\end{align}
Moreover, we have
\begin{align}\label{eqn:gff}
\nonumber &\quad \sum_{k=1}^{K-1}\sum_{j=0}^{k-1}\big(\frac{1}{2}\big)^{k-1-j}\sqrt{2L_F\big( \F(\x_{j},\check{\y}^*(\x_{j}))-\F^*\big)}
\\\nonumber&\leq \sum_{k=0}^{K-1}\frac{1}{2^{k}}\sum_{k=0}^{K-1}\sqrt{2L_F\big( \F(\x_{k},\check{\y}^*(\x_{k}))-\F^*\big)}
\\\nonumber&\leq 2\sqrt{2L_F}\sum_{k=0}^{K-1}\sqrt{ \F(\x_{k},\check{\y}^*(\x_{k}))-\F^*}
\\\nonumber&\leq 2\sqrt{2L_F}\sum_{k=0}^{K-1}\big(\sqrt{\F(\x_{0},\check{\y}^*(\x_{0}))-\F^*}\\\nonumber&+\frac{1}{\sqrt{2L_F}}\sum_{r=0}^{k-1}\|\widehat{\Delta}_r \|\big)
\\\nonumber&\leq 2\sqrt{2L_F}\sum_{k=0}^{K-1}\big(\sqrt{\F(\x_{0},\check{\y}^*(\x_{0}))-\F^*}\\&+\frac{1}{\sqrt{2L_F}}\sum_{r=0}^{K-1}\|\widehat{\Delta}_r \|\big),
\end{align}
where $\widehat{\Delta}_r := \widehat{\nabla} \mathbf{F}(\x_r,\y_r^M)- \nabla \mathbf{F}(\x_r,\check{\y}^*(\x_r))$ and the third inequality follows from Lemma \ref{lem:first:conv}.
\end{subequations}

Putting \eqref{eqn:h}-\eqref{eqn:gff} back into \eqref{eqn:fr}, we obtain that
\begin{align*}
\nonumber &\quad \sum_{k=0}^{K-1}\|\widehat{\nabla} \mathbf{F}(\x_k,\y^M_k)- \nabla \mathbf{F}(\x_k,\check{\y}^*(\x_k)) \|
\\&\leq 1
+2C \left(1- \beta b_g\right)^{M/2} \left( \|\y_{0}-\check{\y}^*(\x_{0}) \|+\varrho \alpha \right)
  \\ &+C \left(1- \beta b_g\right)^{M/2} \varrho \alpha2\sqrt{2L_F} K \Big(  \sqrt{ \F(\x_{0},\check{\y}^*(\x_{0}))-\F^*}
  \\&+\frac{1}{\sqrt{2L_F}} \sum_{r=0}^{K-1}\|\widehat{\Delta}_r \|\Big).
\end{align*}
Choose $M$ such that
\begin{subequations}
\begin{align}
\label{eqn:lem:conv} & 2C\left(1- \beta b_g\right)^{M/2}\leq 1, \quad \textnormal{and}
  \\ \label{eqn:lem:conv2}&  2C\varrho\left(1- \beta b_g\right)^{M/2} \alpha  \leq \frac{1}{2K}.
\end{align}
\end{subequations}
It is easy to verify that both \eqref{eqn:lem:conv} and \eqref{eqn:lem:conv2} hold when
\begin{align*}
M&\geq \frac{2\log \left(   1/2C\min\{1,2\varrho K\alpha\} \right)}{\log\left(1- \beta b_g\right)}=\mathcal{O}\left(\frac{K\alpha}{\beta}\right).
\end{align*}
By combining the inequalities above, we obtain
\begin{align*}
\nonumber &\quad \sum_{k=0}^{K-1}\|\widehat{\nabla} \mathbf{F}(\x_k,\y^M_k)- \nabla \mathbf{F}(\x_k,\check{\y}^*(\x_k)) \|
\\&\leq 1
+  \|\y_{0}-\check{\y}^*(\x_{0}) \|+\varrho
 \\ & +    \frac{\sqrt{2L_F}}{2}  \left(\sqrt{ \F(\x_{0},\check{\y}^*(\x_{0}))-\F^*}+\frac{1}{\sqrt{2L_F}} \sum_{r=0}^{K-1}\|\widehat{\Delta}_r \|\right).
\end{align*}
Rearranging the terms above yields the desired result.
\end{proof}
\subsection{Proof of Lemma~\ref{thm33}}

\begin{lm}\label{lem:dif:non-convex}
Suppose  Assumptions \ref{assu:netw} and \ref{assu:lip} hold.
Set parameter $\beta$ as in Eq. \eqref{eqn:beta}.
If $M\geq \mathcal{O}((1+\log\left(\alpha\right))/\beta)$, then we have
\begin{align*}
\nonumber\quad &\quad  \sum_{k=0}^{K-1}  \|\widehat{\nabla} \mathbf{F}(\x_k,\y^M_k)-\nabla \mathbf{F}(\x_k,\check{\y}^*(\x_k))\|^2
   \\&\leq
2\eta^2\sum_{k=0}^{K-1}\rho^{2(U+1)}
+2C^2 \left(1- \beta b_g\right)^{M} \\&\Big( P_0
+12\varrho^2\alpha^2 \sum_{k=0}^{K-1}\big(\|\nabla \mathbf{F}(\x_{k},\check{\y}^*(\x_{k}))\|^2+ \eta^2 \rho^{2(U+1)}\big)\Big).
\end{align*}
\end{lm}
\begin{proof}
The proof is similar to that of \eqref{L11}. From Lemma \ref{lem:lip}--(\ref{L1}) and Lemma \ref{yy}, we get
\begin{align}\label{mmm}
\nonumber&\quad \|\tilde{\nabla} \mathbf{F}(\x_k,\y^M_k)- \nabla \mathbf{F}(\x_k,\check{\y}^*(\x_k)) \|^2
\\\nonumber &\leq C^2\| \y_{k}^{M}-\check{\y}^*(\x_k)\|^2
\\&\leq C^2 \left(1- \beta b_g\right)^{M} \|\y_{k}^{0}-\check{\y}^*(\x_k) \|^2,
\end{align}
where $\tilde{\nabla} \mathbf{F}(\x_k,\y^M_k)$ is defined as in Eq. \eqref{eqn:grad:tildF}.
Next, we bound $\|\y_{k}^{0}-\check{\y}^*(\x_k) \|^2$. By the warm start strategy $\y_{k}^{0}=\y_{k-1}^{M}$, we have
\begin{align*}
 \nonumber  &\quad \|\y_{k}^{0}-\check{\y}^*(\x_k) \|^2
 \\\nonumber&\leq 2\|\y_{k-1}^{M}-\check{\y}^*(\x_{k-1}) \|^2
 \\\nonumber&+2\|\check{\y}^*(\x_{k-1})-\check{\y}^*(\x_k) \|^2
  \\ \nonumber & \leq 2\left(1- \beta b_g\right)^{M}\|\y_{k-1}^{0}-\check{\y}^*(\x_{k-1}) \|^2
  +2\varrho^2 \|\x_{k-1}-\x_{k} \|^2
  \\& =2 \left(1- \beta b_g\right)^{M}\|\y_{k-1}^{0}-\check{\y}^*(\x_{k-1}) \|^2
\\&  +2\varrho^2\alpha^2 \|\widehat{\nabla} \mathbf{F}(\x_{k-1},\y_{k-1}^M) \|^2,
\end{align*}
where the second inequality is by Lemmas \ref{yy} and \ref{lem:lip}--(\ref{L2}), the last line uses Eq. \eqref{eqn:update:DAGM}.

Consequently, using the fact that $\|a+b+c\|^2\leq 3\|a\|^2+3\|b\|^2+3\|c\|^2$, we find
\begin{subequations}

\begin{align}
\nonumber  &\quad \|\y_{k}^{0}-\check{\y}^*(\x_k) \|^2
\\\nonumber&\leq 2 \left(1- \beta b_g\right)^{M}\|\y_{k-1}^{0}-\check{\y}^*(\x_{k-1}) \|^2
  \\\nonumber&+6\varrho^2\alpha^2 \|\nabla \mathbf{F}(\x_{k-1},\check{\y}^*(\x_{k-1})) \|^2
  \\\label{dcc}&+6\varrho^2\alpha^2 \|\tilde{\nabla} \mathbf{F}(\x_{k-1},\y_{k-1}^M)-\nabla \mathbf{F}(\x_{k-1},\check{\y}^*(\x_{k-1})) \|^2
  \\\label{dcc1}& +6\varrho^2\alpha^2 \|\widehat{\nabla} \mathbf{F}(\x_{k-1},\y_{k-1}^M)-\tilde{\nabla} \mathbf{F}(\x_{k-1},\y_{k-1}^M) \|^2.
\end{align}

\end{subequations}
We proceed by separately upper-bounding terms \eqref{dcc} and \eqref{dcc1}.
To this end, based on \eqref{mmm}, we have
\begin{align*}
   \eqref{dcc} \leq
  6\varrho^2\alpha^2 C^2 \left(1- \beta b_g\right)^{M} \|\y_{k-1}^{0}-\check{\y}^*(\x_{k-1}) \|^2.
\end{align*}
Further, using the arguments similar to Eq. \eqref{cxc}, we also have
\begin{align*}
 \eqref{dcc1}\leq   6\varrho^2\alpha^2 \eta^2 \rho^{2(U+1)}.
\end{align*}
By combining these inequalities one obtains:
\begin{align*}
\nonumber  &\quad\|\y_{k}^{0}-\check{\y}^*(\x_k) \|^2 \\&\leq \left(2+6\varrho^2\alpha^2    C^2\right)\left(1- \beta b_g\right)^{M}\|\y_{k-1}^{0}-\check{\y}^*(\x_{k-1}) \|^2
 \\& +6\varrho^2\alpha^2 \big(\|\nabla \mathbf{F}(\x_{k-1},\check{\y}^*(\x_{k-1})) \|^2
  + \eta^2 \rho^{2(U+1)}\big).
\end{align*}
Next, by setting
$$M\geq \frac{\log \left(1/\left(4+12\varrho^2\alpha^2    C^2\right)\right)}{\log \left(1- \beta b_g\right)}=\mathcal{O}\left(\frac{1}{\beta}+\frac{\log\left(\alpha\right)}{\beta}\right),$$ we get
\begin{align}\label{jk}
 \nonumber  &\quad \|\y_{k}^{0}-\check{\y}^*(\x_k) \|^2
\\\nonumber &\leq \frac{1}{2}\|\y_{k-1}^{0}-\check{\y}^*(\x_{k-1}) \|^2
\\\nonumber &+6\varrho^2\alpha^2 \big(\|\nabla \mathbf{F}(\x_{k-1},\check{\y}^*(\x_{k-1})) \|^2+ \eta^2 \rho^{2(U+1)}\big)
  \\\nonumber &\leq \big(\frac{1}{2}\big)^{k} \|\y_{0}-\check{\y}^*(\x_{0}) \|^2+6\varrho^2\alpha^2 \sum_{j=0}^{k-1}\big(\frac{1}{2}\big)^{k-1-j}
  \\&\big(\|\nabla \mathbf{F}(\x_{j},\check{\y}^*(\x_{j}))\|^2+ \eta^2 \rho^{2(U+1)}\big).
\end{align}
Inserting \eqref{jk} into \eqref{mmm} gives
\begin{align}\label{L22}
\nonumber &\quad\sum_{k=0}^{K-1}\|\tilde{\nabla} \mathbf{F}(\x_k,\y^M_k)- \nabla \mathbf{F}(\x_k,\check{\y}^*(\x_k)) \|^2
\\\nonumber &\leq C^2 \left(1- \beta b_g\right)^{M} \Big(\sum_{k=0}^{K-1}\big(\frac{1}{2}\big)^k P_0
\\\nonumber&+6\varrho^2\alpha^2 \sum_{k=1}^{K-1}\sum_{j=0}^{k-1}\big(\frac{1}{2}\big)^{k-1-j}
\big(\|\nabla \mathbf{F}(\x_{j},\check{\y}^*(\x_{j}))\|^2
\\\nonumber&+\eta^2 \rho^{2(U+1)}\big)\Big)
\\\nonumber & \leq C^2 \left(1- \beta b_g\right)^{M} \Big( P_0
+12\varrho^2\alpha^2 \sum_{k=0}^{K-1}\big(\|\nabla \mathbf{F}(\x_{k},\check{\y}^*(\x_{k}))\|^2
\\&+ \eta^2\rho^{2(U+1)}\big)\Big),
\end{align}
where the second inequality holds since
\begin{align*}
\nonumber &\quad\sum_{k=1}^{K-1}\sum_{j=0}^{k-1}\big(\frac{1}{2}\big)^{k-1-j}\|\nabla \mathbf{F}(\x_{j},\check{\y}^*(\x_{j}))\|^2
\\&\leq \sum_{k=0}^{K-1}\frac{1}{2^k}\sum_{k=0}^{K-1}\|\nabla \mathbf{F}(\x_{k},\check{\y}^*(\x_{k}))\|^2
\\&\leq 2\sum_{k=0}^{K-1}\|\nabla \mathbf{F}(\x_{k},\check{\y}^*(\x_{k}))\|^2.
\end{align*}
Recall that $\iota_k:=\widehat{\nabla} \mathbf{F}(\x_k,\y^M_k)-\tilde{\nabla} \mathbf{F}(\x_k,\y^M_k)$ and
$\Delta_k:= \tilde{\nabla} \mathbf{F}(\x_k,\y^M_k)- \nabla \mathbf{F}(\x_k,\check{\y}^*(\x_k))$.
Then, we have
\begin{align*}
 &\quad\sum_{k=0}^{K-1} \|\widehat{\nabla} \mathbf{F}(\x_k,\y^M_k)-\nabla \mathbf{F}(\x_k,\check{\y}^*(\x_k))\|^2
 \\&\leq 2\sum_{k=0}^{K-1}\|\iota_k \|^2+2\sum_{k=0}^{K-1}\| \Delta_k \|^2.
\end{align*}
Thus, according to the upper bounds in \eqref{cxc} and \eqref{L22} for $\sum_{k=0}^{K-1}\|\iota_k \|^2$ and $\sum_{k=0}^{K-1}\| \Delta_k \|^2$, respectively, we obtain the desired result.
\end{proof}

We are now ready to prove Lemma \ref{thm33}.

\begin{proof}
Utilizing the $L_{F}$-smooth property of $\F$ due to Lemma \ref{lem:lip}--(\ref{L3}), we have
\begin{align}\label{po}
 \nonumber &\quad\F(\x_{k+1},\check{\y}^*(\x_{k+1}))
 \\\nonumber& \leq \F(\x_{k},\check{\y}^*(\x_{k}))+\langle \nabla \F(\x_{k},\check{\y}^*(\x_{k})),\x_{k+1}-\x_k  \rangle
 \\\nonumber&+\frac{L_{F}}{2}\|\x_{k+1} -\x_k\|^2\\\nonumber
  &= \F(\x_{k},\check{\y}^*(\x_{k}))-\alpha\langle \nabla \F(\x_{k},\check{\y}^*(\x_{k})),  \widehat{\nabla} \mathbf{F}(\x_k,\y_k^M)  \rangle \\\nonumber&+\frac{L_{F}\alpha ^2}{2}\| \widehat{\nabla} \mathbf{F}(\x_k,\y_k^M)\|^2\\\nonumber
&\leq \F(\x_{k},\check{\y}^*(\x_{k}))-\alpha\|  \nabla \mathbf{F}(\x_k,\check{\y}^*(\x_{k}))\|^2
\\\nonumber&-\alpha\langle   \nabla \mathbf{F}(\x_k,\check{\y}^*(\x_{k})) ,\widehat{\Delta}_k \rangle
\\\nonumber &+L_{F}\alpha ^2\| \nabla \mathbf{F}(\x_k,\check{\y}^*(\x_k))\|^2+L_{F}\alpha ^2\| \widehat{\Delta}_k\|^2\\\nonumber
&\leq \F(\x_{k},\check{\y}^*(\x_{k}))-\frac{\alpha}{2}(1-2L_{F}\alpha)\| \nabla \mathbf{F}(\x_k,\check{\y}^*(\x_{k}))\|^2
\\&+(\frac{\alpha }{2}+L_{F}\alpha ^2)\|\widehat{\Delta}_k\|^2 ,
\end{align}
where $\widehat{\Delta}_k := \widehat{\nabla} \mathbf{F}(\x_k,\y_k^M)- \nabla \mathbf{F}(\x_k,\check{\y}^*(\x_k))$, the equality is the result of \eqref{eqn:update:DAGM}, the second inequality holds due to
$(a+b)^2\leq 2(\|a\|^2+\|b\|^2)$, and the last inequality follows from $ab< (1/2)(a^2+b^2)$.

Summing up both sides of \eqref{po}, using Lemma \ref{lem:dif:non-convex}, we get
\begin{equation}\label{ok}
\begin{split}
 & \quad\sum_{k=0}^{K-1} \left(\F(\x_{k+1},\check{\y}^*(\x_{k+1}))-\F(\x_{k},\check{\y}^*(\x_{k}))\right)
\\  & \leq
-\frac{\alpha}{2}(1-2L_{F}\alpha)\sum_{k=0}^{K-1}\| \nabla \F(\x_{k},\check{\y}^*(\x_{k}))\|^2
\\&+\left(\frac{\alpha }{2}+L_{F}\alpha ^2\right)2\eta^2\sum_{k=0}^{K-1}\rho^{2(U+1)}
\\  &+\left(\frac{\alpha }{2}+L_{F}\alpha ^2\right)2 C^2 \left(1- \beta b_g\right)^{M}
\\&\left( P_0
+12\varrho^2\alpha^2 \sum_{k=0}^{K-1}\left(\|\nabla \mathbf{F}(\x_{k},\check{\y}^*(\x_{k}))\|^2+ \eta^2 \rho^{2(U+1)}\right)\right),
\end{split}
\end{equation}
where $\rho$ is defined in Lemma \ref{symmetric_term_bounds11}.
Choose $M$ such that
\begin{align}\label{vvb}
\nonumber &\left(\frac{\alpha }{2}+L_{F}\alpha ^2\right) 2C^2 \left(1- \beta b_g\right)^{M}12\varrho^2\alpha \leq \frac{1}{4},
  \quad \quad \textnormal{and}
  \\& 2C^2\left(1- \beta b_g\right)^{M}\leq 1.
\end{align}
It is easy to verify that \eqref{vvb} holds when
\begin{align*}
M&\geq\frac{\max\left\{\log\left(\frac{1}{2C^2}\right),\log\left(\frac{1}{96\varrho^2\alpha C^2\left(\frac{\alpha }{2}+L_{F}\alpha ^2\right)}\right)\right\}}{\log\left(1- \beta b_g\right)}
\\&=\mathcal{O}\left(\frac{\log\left(\alpha\right)}{\beta} \right).
\end{align*}
Now applying Eq. \eqref{vvb} into Eq. \eqref{ok}, we have
\begin{align*}
&\quad \alpha(\frac{1}{4}-L_{F}\alpha) \sum_{k=0}^{K-1}\|\nabla \F(\x_k,\check{\y}^*(\x_k))\|^2
\\&\leq \F(\x_0,\check{\y}^*(\x_0))-\F(\check{\x}^*,\check{\y}^*(\check{\x}^*))
\\&
+(\frac{\alpha }{2}+L_{F}\alpha ^2)\left((2+12\varrho^2\alpha^2  )\eta^2\sum_{k=0}^{K-1}\rho^{2(U+1)}+ P_0\right).
\end{align*}
Let $D_{F}:=\F(\x_0,\check{\y}^*(\x_0))-\F(\check{\x}^*,\check{\y}^*(\check{\x}^*))$.
Rearranging the terms and using $\alpha \leq 1/8L_{F} $, we get
\begin{align*}
 &\quad\frac{\alpha}{8} \sum_{k=0}^{K-1}\|\nabla \F(\x_k,\check{\y}^*(\x_k))\|^2
\\&\leq D_{F}
+\frac{5}{64L_{F}}\left(\left(2+\frac{3\varrho^2}{16 L_F^2}\right )\eta^2\sum_{k=0}^{K-1}\rho^{2(U+1)}+ P_0\right).
\end{align*}
Let
$$U=\left|\left\lceil\frac{\log (\eta^2K)}{2\log (\frac{1}{\rho})} \right\rceil\right|.$$
Then, since $\rho<1$ in Lemma \ref{symmetric_term_bounds11}, we have
\begin{align*}
\eta^2\rho^{2(U+1)}\leq \eta^2\rho^{2U}= \frac{1}{K},
\end{align*}
which gives
\begin{align}\label{ffn}
\nonumber&\quad\frac{1}{K}\sum_{k=0}^{K-1}\|\nabla \F(\x_k,\check{\y}^*(\x_k))\|^2
\\&\leq \frac{8}{K\alpha} \left(D_{F}
+\frac{5}{64L_{F}}\left(\left(2+\frac{3\varrho^2}{16 L_F^2}\right )
+ P_0\right)\right).
\end{align}
Note that from Lemma \ref{lem:lip}, we have
\begin{align}\label{eqn:ph}
\nonumber   &\quad \|\nabla \F(\x_k,\check{\y}^*(\x_k))-\nabla \F(\x_k,\y^*(\x_k))\|^2
   \\\nonumber &\leq  C^2\|\check{\y}^*(\x_k)-\y^*(\x_k) \|^2
   \\\nonumber &\leq C^2\left(\frac{2\hat{L}_g\beta}{(1-\sigma)}+\frac{2\hat{L}_g\beta^{1/2}}{(1-\sigma)^{1/2}\mu_g^{1/2}}\right)^2
   \\&\leq C^2\left(\frac{4\hat{L}_g\beta}{(1-\sigma)}+\frac{4\hat{L}_g\beta^{1/2}}{(1-\sigma)^{1/2}\mu_g^{1/2}}\right),
\end{align}
where the second inequality is by Lemma \ref{cl}. Then, from $\|a+b\|^2\leq 2\|a\|^2+2\|b\|^2$ for all $a,b\in \mathbb{R}^d$ and \eqref{eqn:ph}, we have
\begin{align*}
\|\nabla \F(\x_k,\y^*(\x_k))\|^2&\leq 2\|\nabla \F(\x_k,\check{\y}^*(\x_k))\|^2
\\&+2\|\nabla \F(\x_k,\check{\y}^*(\x_k))-\nabla \F(\x_k,\y^*(\x_k))\|^2
\\&\leq  2\|\nabla \F(\x_k,\check{\y}^*(\x_k))\|^2
\\&+2C^2\left(\frac{4\hat{L}_g\beta}{(1-\sigma)}+\frac{4\hat{L}_g\beta^{1/2}}{(1-\sigma)^{1/2}\mu_g^{1/2}}\right).
\end{align*}
Then, from \eqref{ffn}, we get
\begin{align*}
\nonumber&\quad\frac{1}{K}\sum_{k=0}^{K-1}\|\nabla \F(\x_k,{\y}^*(\x_k))\|^2
\\&\leq
\frac{16}{K\alpha} \left(D_{F}
+\frac{5}{64L_{F}}\left(\left(2+\frac{3\varrho^2}{16 L_F^2} \right)
+ P_0\right)\right)
\\\nonumber &+2C^2\left(\frac{4\hat{L}_g\beta}{(1-\sigma)}+\frac{4\hat{L}_g\beta^{1/2}}{(1-\sigma)^{1/2}\mu_g^{1/2}}\right).
\end{align*}
Then, from
\begin{align*}
&\quad\frac{1}{K}\sum_{k=0}^{K-1}\|\frac{1}{n} 1^{\top}\nabla \F(\x_k,\y^*(\x_k))\|^2
\\&\leq \frac{1}{nK}\sum_{k=0}^{K-1}\|\nabla \F(\x_k,{\y}^*(\x_k))\|^2,
\end{align*}
we get the desired result.
\end{proof}

\subsection{Proof of Lemma~\ref{vbb}}
\begin{proof}
From Eq. \eqref{ujg}, we have
  \begin{align}\label{dsg}
\nonumber &\quad \frac{1}{K}\sum_{k=0}^{K-1}\|\frac{1}{n} 1^{\top}(\x_{k} -\bar{\x}_k)\|^2
\\\nonumber&\leq \frac{\alpha^2}{(1-\sigma)^2K} \sum_{k=0}^{K-1}\|\frac{1^{\top}}{n\alpha}(\mathbf{I}_{nd_1}-\acute{\W})\x_k \|^2\\\nonumber
&\leq\frac{2\alpha^2}{(1-\sigma)^2K}
\\\nonumber&\cdot\sum_{k=0}^{K-1}\left( \|\frac{1}{n} 1^{\top}\big(\frac{1}{\alpha}(\mathbf{I}_{nd_1}-\acute{\W})\x_k +\nabla \mathbf{f}(\x_k,{\y}^*(\x_k))\big)\|^2
\right.\\\nonumber&\left.+\|\frac{1}{n} 1^{\top}\nabla \mathbf{f}(\x_k,{\y}^*(\x_k))\|^2\right)\\
  &\leq \frac{2\alpha^2}{(1-\sigma)^2}\left(\frac{1}{K}\sum_{k=0}^{K-1} \|\frac{1}{n} 1^{\top}\nabla \mathbf{F}(\x_k,{\y}^*(\x_k))\|^2
+\frac{\hat{C}^2}{n}\right),
  \end{align}
  where the second inequality uses the fact that $\|a\|^2\leq 2\|a+b\|^2+2\|b\|^2$ for all $a,b\in \mathbb{R}^d$, and the last inequality follows from Eq. \eqref{eqn:approximate:prob} and \eqref{eqn:om}.

\end{proof}

\section{Additional Auxiliary Lemmas}

In this section, we present several technical lemmas used in the proofs.

\begin{lm}\label{vv}
({\cite[Lemma 3.11]{bubeck2015convex}})
If $F(\xx)$ is $\mu$-strongly convex and $L$-smooth, then
\begin{align*}
&\langle \nabla F(\xx)-\nabla F(\yy),\xx-\yy \rangle\geq\frac{\mu L}{\mu+L}\| \xx-\yy\|^2
\\&+\frac{1}{\mu+L}\| \nabla F(\xx)-\nabla F(\yy)\|^2.
\end{align*}
\end{lm}
\begin{lm}(\cite[Lemma 2]{cutkosky2018distributed})\label{lem:smooth}
If $F(\xx)$ is $L$-smooth convex function with minimizer $\xx^*$ s.t $\nabla F(\xx^*)=0$, then
$$\|\nabla F(\xx) \|^2=\|\nabla F(\xx)-\nabla F(\xx^*) \|^2\leq 2L\big(F(\xx)- F(\xx^*)\big).$$
\end{lm}
\begin{lm}\label{lem:sequ}(\cite[Lemma 1]{schmidt2011convergence})
Assume that the nonnegative sequence $\{ u_K\}$
satisfies the following recursion for all $k\geq 1$
\begin{equation*}
u_K^2 \leq S_K+\sum_{k=1}^{K} \lambda_k u_k,
\end{equation*}
with $\{S_K\}$ an increasing sequence, $S_0 \geq u_0^2$ and $\lambda_k\geq 0$ for all $k$. Then, for all $K\geq 1$,
\begin{equation*}
u_K \leq \frac{1}{2}\sum_{k=1}^{K} \lambda_k+\Big(S_K+(\frac{1}{2}\sum_{k=1}^{K} \lambda_k)^2 \Big)^{\frac{1}{2}}.
\end{equation*}
\end{lm}
\begin{lm}\label{ww}(\cite[Lemma 16]{koloskova2019decentralized})
Under Assumptions~\ref{assu:netw:item1} and \ref{assu:netw:item2}, we have
\begin{align*}
    \|{\mr W}^{k}-\frac{1}{n}1_n1_n^{\top} \|^2\leq \sigma^k,\quad \forall k\in[K].
\end{align*}
\end{lm}
\begin{lm}(\cite[Proposition 1]{mokhtari2016network})\label{eigenvalue_bounds}
Suppose Assumptions  \ref{assu:netw} and \ref{assu:g:strongly} hold. Then, for the matrices  $\mathbf H$, $\mathbf D$, and $\mathbf B$ defined in \eqref{hhh}, we have
\begin{align*}
 \beta \mu_{g} \mathbf I_{n d_2}\ \preceq \ & {\mathbf H} \ \preceq \ (2(1-\theta) + \beta C_{g_{\yy\yy}} \mr )\mathbf I_{n d_2} ,\\
 (2(1-\Theta)+\beta \mu_{g}) \mathbf I_{n d_2}\ \preceq\ & {\mathbf D}\ \preceq\ (2(1-\theta)+\beta C_{g_{\yy\yy}} \mr )\mathbf I_{n d_2} ,\\
 \mathbf 0\ \preceq\ & \ {\mathbf B}\ \preceq\ 2 (1-\theta) \mathbf I_{n d_2}.
\end{align*}
\end{lm}

\end{document}